\documentclass[5p,twocolumn,preprint,authoryear]{elsarticle}

\usepackage{lineno,hyperref}
\modulolinenumbers[5]

\usepackage{filecontents}
\usepackage{csvsimple}
\usepackage{subcaption}
\usepackage{tikz}
\usepackage{pgfplots}
\usepackage{pgfplotstable}
\usepackage{mathtools}
\usepackage{placeins}
\usepackage{multirow}
\usepackage{float}
\usepackage{ctable}
\usepackage{amssymb}
\usepackage{pifont}
\usepackage{textcomp}
\usepackage{gensymb}
\usepackage{todonotes}
\usepackage{comment}
\usepackage{pgfplots}

\usepackage[utf8]{inputenc} 
\usepackage[T1]{fontenc}
\usepgfplotslibrary{statistics}
\usepgfplotslibrary{groupplots}
\pgfplotsset{compat=1.13}

\hypersetup{%
	pdfborder = {0 0 0},
	colorlinks=true,
	urlcolor=cyan,
}

\makeatletter

\newcommand*{\@rowstyle}{}

\newcommand*{\rowstyle}[1]{
  \gdef\@rowstyle{#1}%
  \@rowstyle\ignorespaces%
}

\newcolumntype{=}{
  >{\gdef\@rowstyle{}}%
}

\newcolumntype{+}{
  >{\@rowstyle}%
}

\makeatother

\usepackage{array}
\newcolumntype{L}[1]{>{\raggedright\let\newline\\\arraybackslash\hspace{0pt}}m{#1}}
\newcolumntype{C}[1]{>{\centering\let\newline\\\arraybackslash\hspace{0pt}}m{#1}}
\newcolumntype{R}[1]{>{\raggedleft\let\newline\\\arraybackslash\hspace{0pt}}m{#1}}
\newcolumntype{?}{!{\vrule width 1pt}}

\journal{Expert Systems with Applications}

\biboptions{authoryear}

\begin{document}

\begin{frontmatter}

  \title{Low-latency Perception in Off-Road Dynamical Low Visibility Environments}

  \author{Nelson Alves\fnref{ufba-ppgee,senai,equally}}
  \cortext[Nelson Alves]{Corresponding author: Nelson Alves.}
  \ead{nelsonafn@gmail.com}
    \author{Marco Ruiz\fnref{senai,equally}}
  \ead{marco.rueda@fieb.org.br}
    \author{Marco Reis\fnref{senai,ufba-ppgm,senai-cmit,equally}}
  \ead{marcoreis@fieb.org.br}
      \author{Tiago Cajahyba\fnref{senai}}
  \ead{tiago.cajahyba@fieb.org.br}
    \author{Davi Oliveira\fnref{senai}}
  \ead{davi.oliveira@fieb.org.br}
     \author{Ana Barreto\fnref{senai,ufba-ppgee}}
  \ead{ana.barreto@fieb.org.br}
    \author{Eduardo F. Simas Filho\fnref{ufba-ppgee}}
  \ead{eduardo.simas@ufba.br}
     \author{Wagner L. A. de Oliveira\fnref{ufba-ppgee}}
  \ead{ oliveira.wagner@ufba.br}
       \author{Leizer Schnitman\fnref{ufba-ppgm}}
  \ead{leizer@ufba.br}
 	  \author{Roberto L. S. Monteiro\fnref{senai-cmit}}
  \ead{roberto.monteiro@fieb.org.br}

  \address[ufba-ppgee]{Electrical Engineering Program, Federal University of Bahia, Salvador, Brazil}
  \address[senai]{Brazilian Institute of Robotics, SENAI CIMATEC, Salvador, Brazil}
  \address[ufba-ppgm]{Federal University of Bahia, PPGM, Salvador, Brazil}
  \address[senai-cmit]{Computational Modeling and Industrial Technology Program, SENAI CIMATEC, Salvador, Brazil}
  
  \fntext[equally]{\textbf{NA \& MR contributed equally to this work.}}

\begin{abstract}
	\textit{This work proposes a perception system for autonomous vehicles and advanced driver assistance specialized on unpaved roads and off-road environments. In this research, the authors have investigated the behavior of Deep Learning algorithms applied to semantic segmentation of off-road environments and unpaved roads under differents adverse conditions of visibility. Almost 12,000 images of different unpaved and off-road environments were collected and labeled. It was assembled an off-road proving ground exclusively for its development. The proposed dataset also contains many adverse situations such as rain, dust, and low light. To develop the system, we have used convolutional neural networks trained to segment obstacles and areas where the car can pass through. We developed a Configurable Modular Segmentation Network (CMSNet) framework to help create different architectures arrangements and test them on the proposed dataset.  Besides, we also have ported some CMSNet configurations by removing and fusing many layers using TensorRT, C++, and CUDA to achieve embedded real-time inference and allow field tests. 
	The main contributions of this work are: a new dataset for unpaved roads and off-roads environments containing many adverse conditions such as night, rain, and dust; a CMSNet framework; an investigation regarding the feasibility of applying deep learning to detect region where the vehicle can pass through when there is no clear boundary of the track; a study of how our proposed segmentation algorithms behave in different severity levels of visibility impairment; and an evaluation of field tests carried out with semantic segmentation architectures ported for real-time inference.
    The proposed dataset (named Kamino) is available at \url{https://github.com/Brazilian-Institute-of-Robotics/offroad\_dataset}, and the experiments at \url{https://github.com/Brazilian-Institute-of-Robotics/autonomous\_perception}.
}

\end{abstract}


\end{frontmatter}


\captionsetup[subfigure]{font=scriptsize}
\section{Introduction}
In autonomous vehicles or general robotic systems, perception is the subsystem responsible for perceiving the environment, i.e., for carrying out the recognition process from data of different sensors. The perception subsystem is one of the most critical tasks in the development of an autonomous car \citep{brummelen:2018:autonomous-vehicle-perception}.  It receives raw data from several sensors, such as RGB cameras, infrared cameras, LiDARs, radars, and must be able to extract understanding scene information. 

There are three paradigms of perception: mediated perception, direct perception, and behavioral cloning (behavior reflex perception or end-to-end driving) \citep{brummelen:2018:autonomous-vehicle-perception}. The approach chosen by this work was the construction of a perception system following the classic paradigm (mediated perception), which is the most used in the development of autonomous cars nowadays \citep{brummelen:2018:autonomous-vehicle-perception}. This approach uses algorithms to recognize relevant elements within the scene then combine them into a unified representation (world model) that is used as an input source for the planning and control module to decide the behavior of the vehicle \citep{chen:2015:deepdriving}. 

Despite using the classic paradigm, visual perception is still a challenge for machines. In this work, it was employed computer vision algorithms that use data-guided modeling with Deep Learning and Convolutional Neural Networks (CNNs) \citep{lecun:1998:ieee-gradient-based-learning} to perform the visual scene perception.

CNNs have become almost ubiquitous, being used in several types of problems, such as: image classification \citep{simonyan:2015:vgg,He:2016:IEEECVPR-resnet}; object detection \citep{Redmon:2016:YOLO:ieeecvpr}; instance segmentation \citep{He:2017:mask-rcnn:ieeeiccv}; and semantic segmentation \citep{Long:2015:FCN:ieeecvpr,Chen:2018:deeplab:ieeeTPAMI,Zhao:2017:PSPNet:ieeecvpr}.
However, in the search for more accurate algorithms, there has been a trend towards more complex and deeper network architectures, reaching up to hundreds of millions of parameters and tens of billions of multiply-accumulate operations (MACs).  So there is no guarantee that such algorithms are fast and computationally efficient to be embedded in real-time applications with limited computational capacity and power restrictions such as autonomous cars and robots.

On the other hand, the search for fast and computationally efficient inference has also been a concern of several other recent works \citep{Zhang:2018:ShuffleNet:IEEE-CVF,Zoph:2018:NasNet:IEEE-CVF,Sandler:2018:MobileNetV2:IEEE-CVF}, which propose network architectures capable of keeping a reduced size and performing well in the benchmarks. These networks manage to keep the computational cost in units of millions of parameters and hundreds of millions of MACs. There are also other works aiming to implement or facilitate the reimplementation of inference algorithms in real-time \citep{Jacob:2018:Quantization:IEEE-CVF}.

\begin{figure}[!t]
    \begin{subfigure}[b]{0.48\linewidth}
            \includegraphics[width=\linewidth]{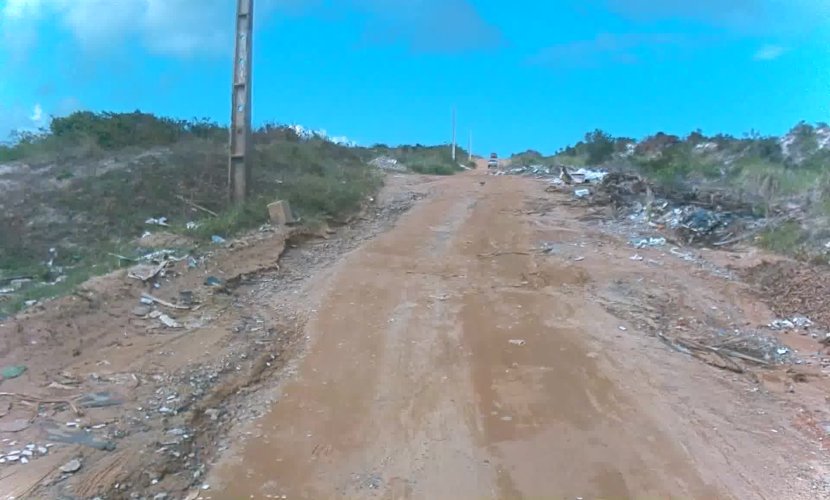}
            \caption{Unpaved road in Brazil}
        \end{subfigure}%
    \hfill%
    \begin{subfigure}[b]{0.48\linewidth}
            \includegraphics[width=\linewidth]{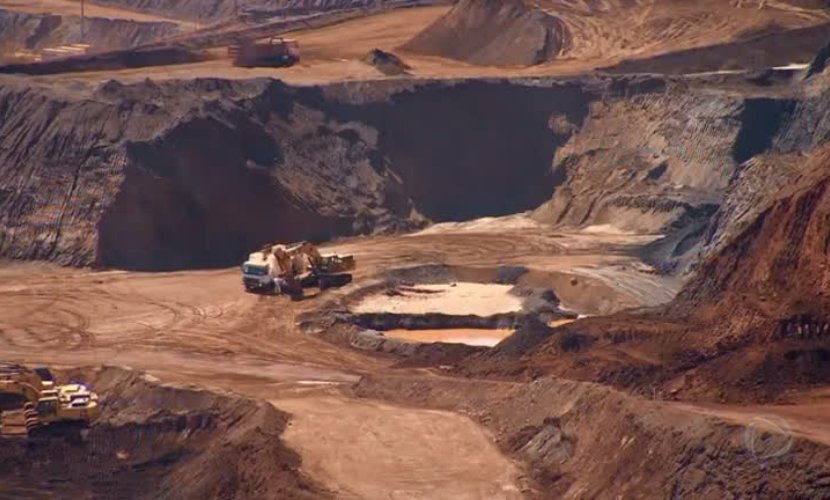}
            \caption{Mining environment}
        \end{subfigure}
    \begin{subfigure}[b]{0.48\linewidth}
            \includegraphics[width=\linewidth]{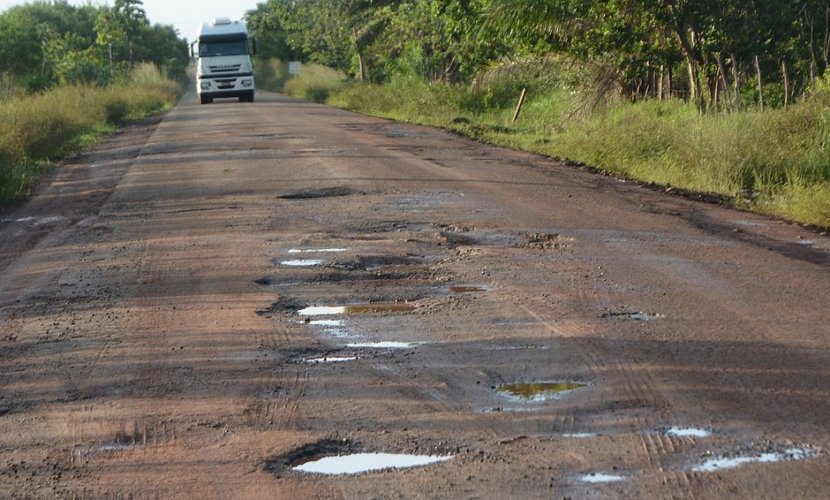}
            \caption{Bumpy intercity road}
        \end{subfigure}%
    \hfill%
    \begin{subfigure}[b]{0.48\linewidth}
            \includegraphics[width=\linewidth]{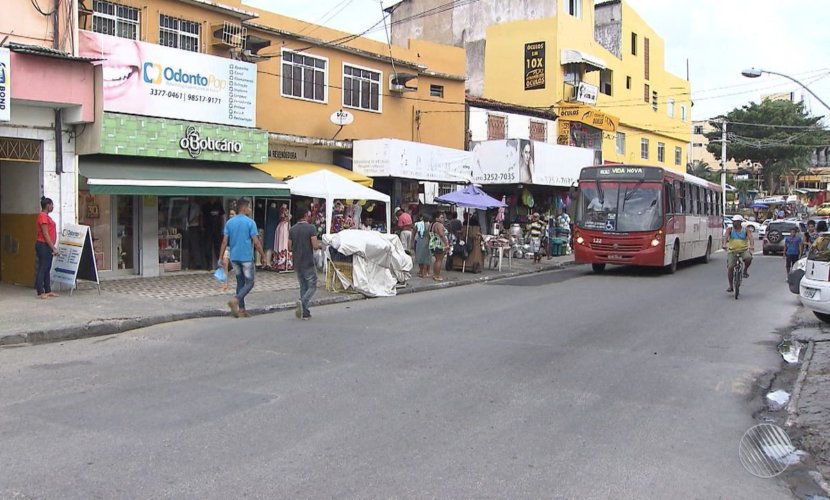}
            \caption{Road poorly signposted}
        \end{subfigure}
    \caption{Common Road and Highway Situations in Brazil.}
    \label{fig:fig1}
\end{figure}

Datasets are another challenge in the development of visual perception systems suitable for off-road environments and unpaved roads. Although autonomous cars research is advancing fast, most of the datasets available for perception module training are focused on urban environments \citep{kitti-fritsch:2013:itsc,complex-urban-dataset-jjeong-2019-ijrr,cordts:2016:cityscapes}. However, in developing countries, there are still numerous urban and rural roads without paving (Fig. \ref{fig:fig1}). In Brazil, only 12.4 \% of the road network is paved, according to National Transport Confederation \citep{cnt:2018:noticias,dnit:2018:relatorio-de-gestao-tematica}.

\begin{figure}[!t]
	\captionsetup[subfigure]{font=scriptsize,labelformat=empty}
	\begin{subfigure}[b]{0.245\linewidth}
		\caption{Image}
		\includegraphics[width=\linewidth]{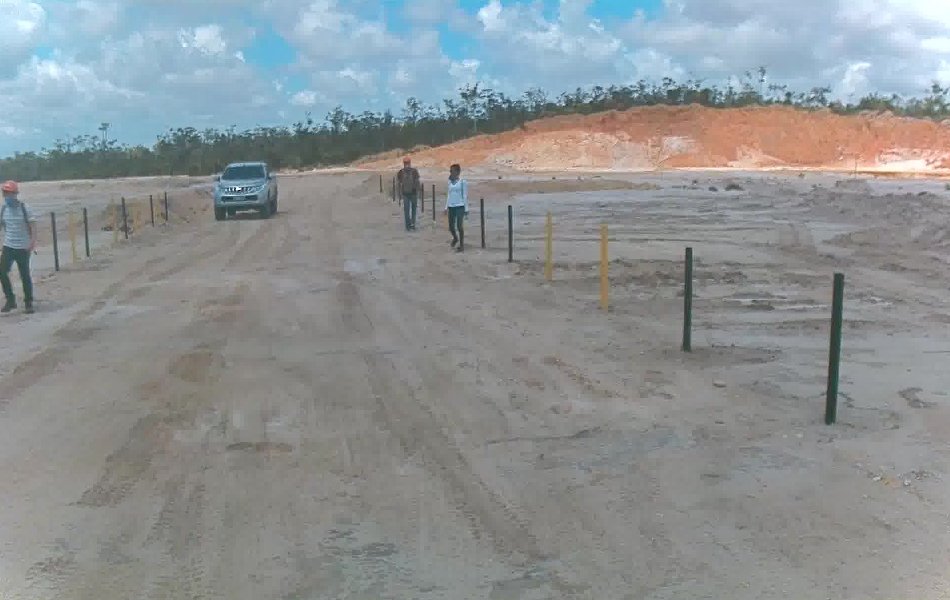}
	\end{subfigure}%
	\hfill
	\begin{subfigure}[b]{0.245\linewidth}
		\caption{Expected}
		\includegraphics[width=\linewidth]{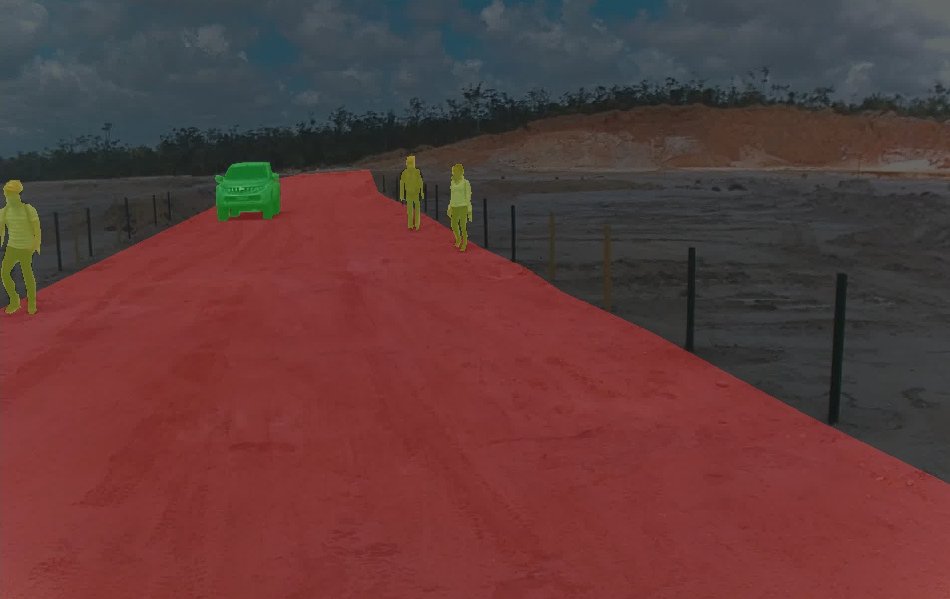}
	\end{subfigure}%
	\hfill
	\begin{subfigure}[b]{0.245\linewidth}
		\caption{PSPNet}
		\includegraphics[width=\linewidth]{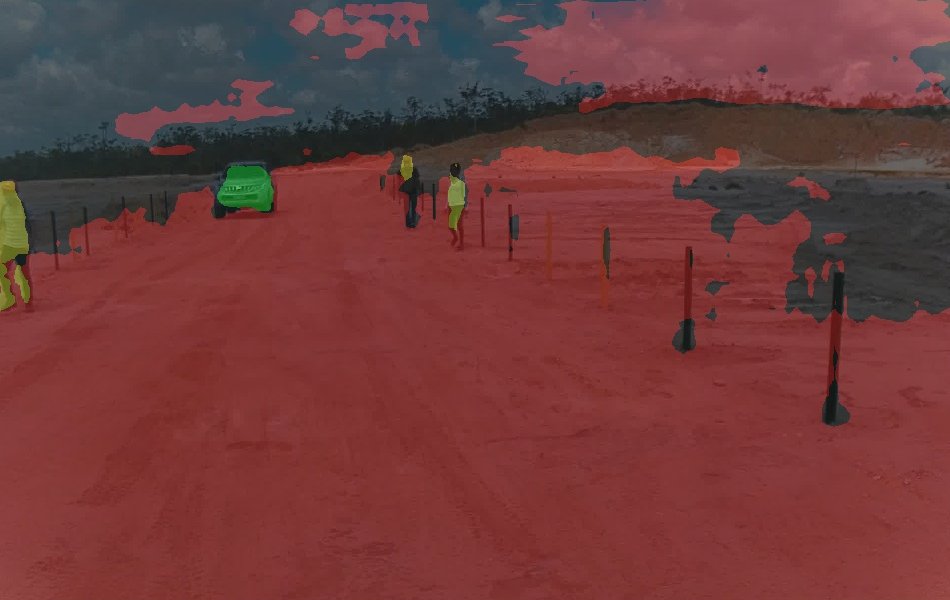}
	\end{subfigure}%
	\hfill
	\begin{subfigure}[b]{0.245\linewidth}
		\caption{DeepLab}
		\includegraphics[width=\linewidth]{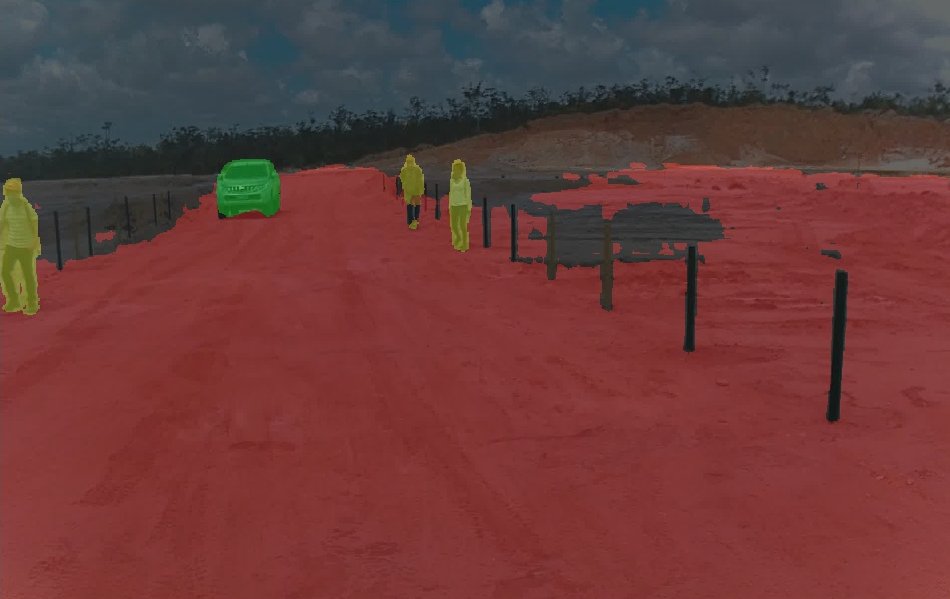}
	\end{subfigure}%
	\vspace{0.003\linewidth}
	
	\begin{subfigure}[b]{0.245\linewidth}
		\includegraphics[width=\linewidth]{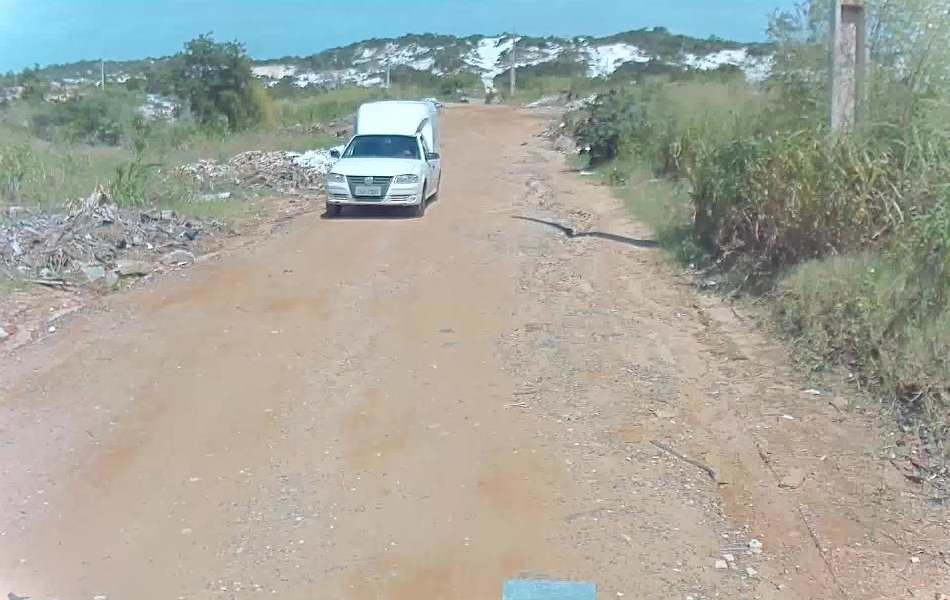}
	\end{subfigure}%
	\hfill
	\begin{subfigure}[b]{0.245\linewidth}
		\includegraphics[width=\linewidth]{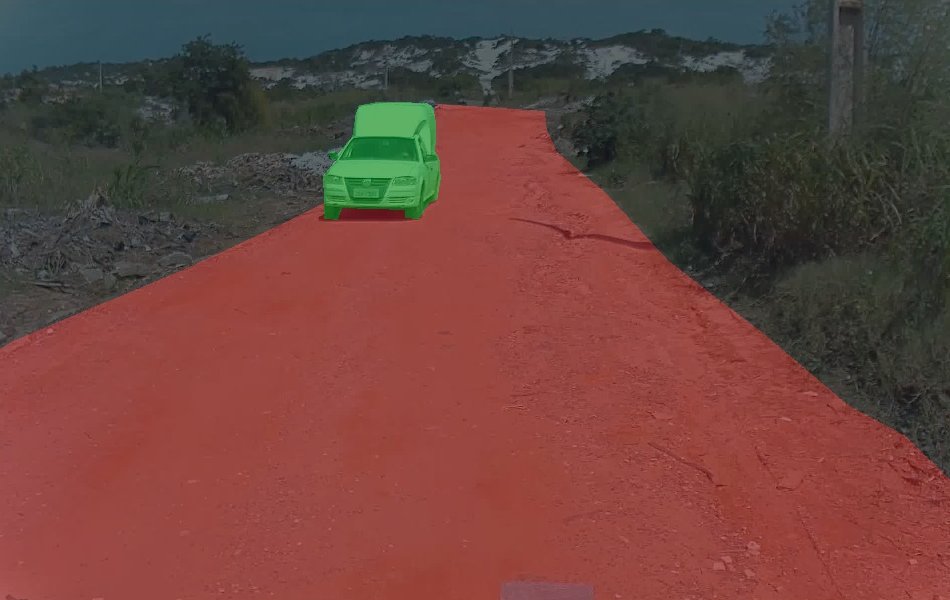}
	\end{subfigure}%
	\hfill
	\begin{subfigure}[b]{0.245\linewidth}
		\includegraphics[width=\linewidth]{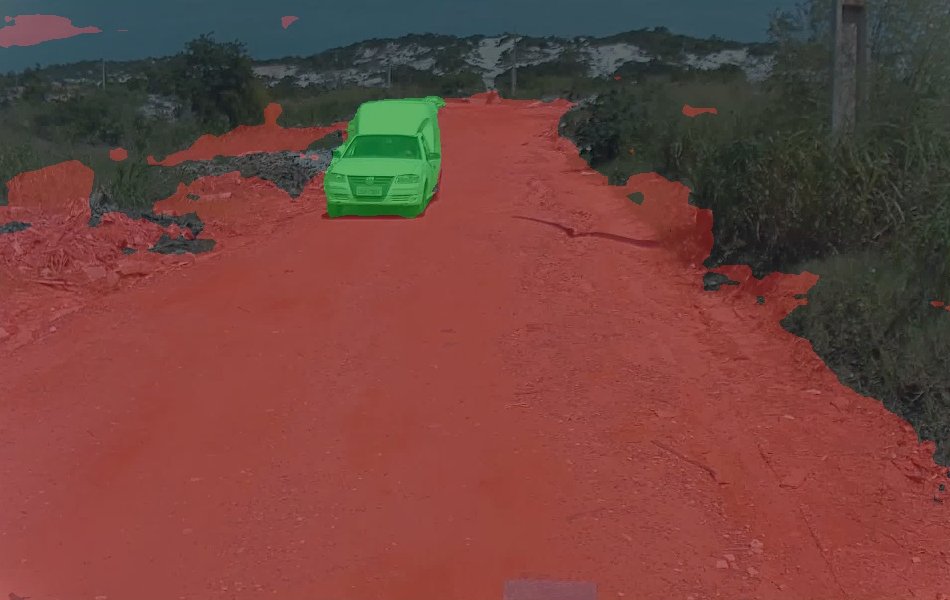}
	\end{subfigure}%
	\hfill
	\begin{subfigure}[b]{0.245\linewidth}
		\includegraphics[width=\linewidth]{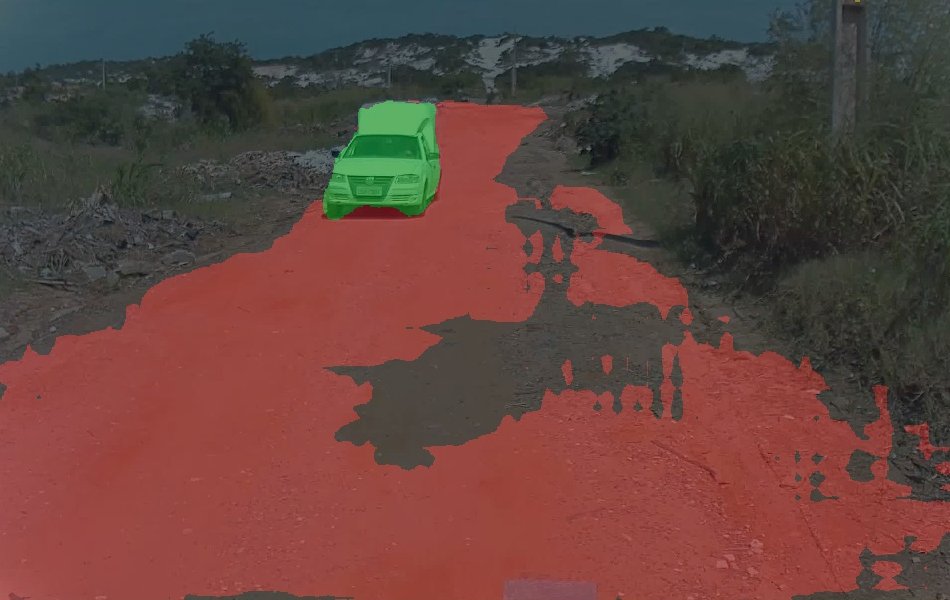}
	\end{subfigure}%
	\vspace{0.003\linewidth}
	
	\begin{subfigure}[b]{0.245\linewidth}
		\includegraphics[width=\linewidth]{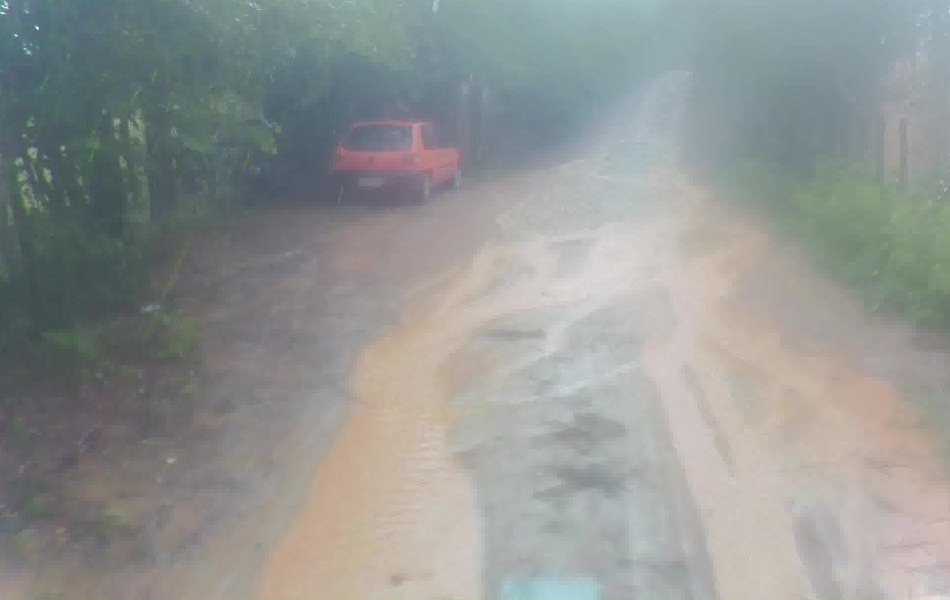}
	\end{subfigure}%
	\hfill
	\begin{subfigure}[b]{0.245\linewidth}
		\includegraphics[width=\linewidth]{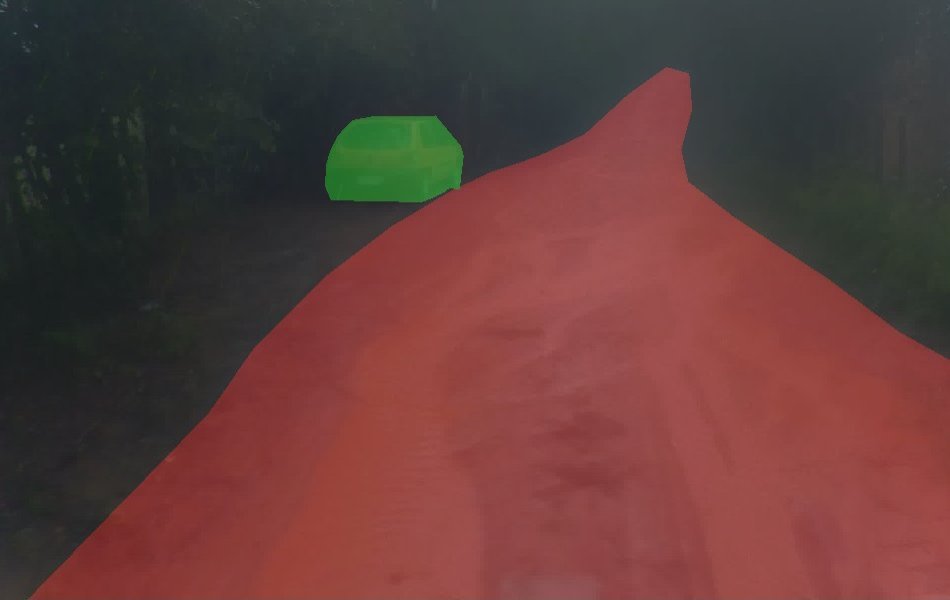}
	\end{subfigure}%
	\hfill
	\begin{subfigure}[b]{0.245\linewidth}
		\includegraphics[width=\linewidth]{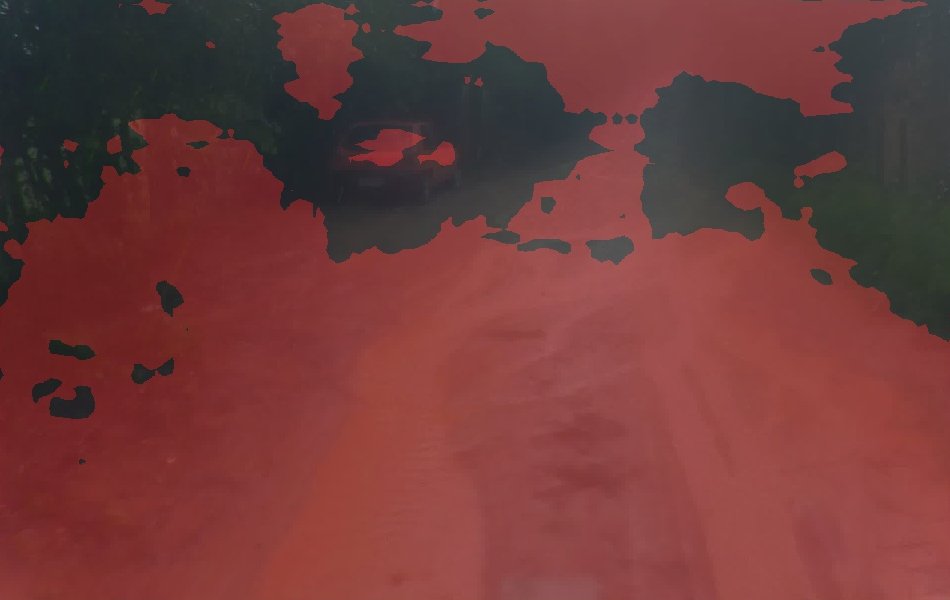}
	\end{subfigure}%
	\hfill
	\begin{subfigure}[b]{0.245\linewidth}
		\includegraphics[width=\linewidth]{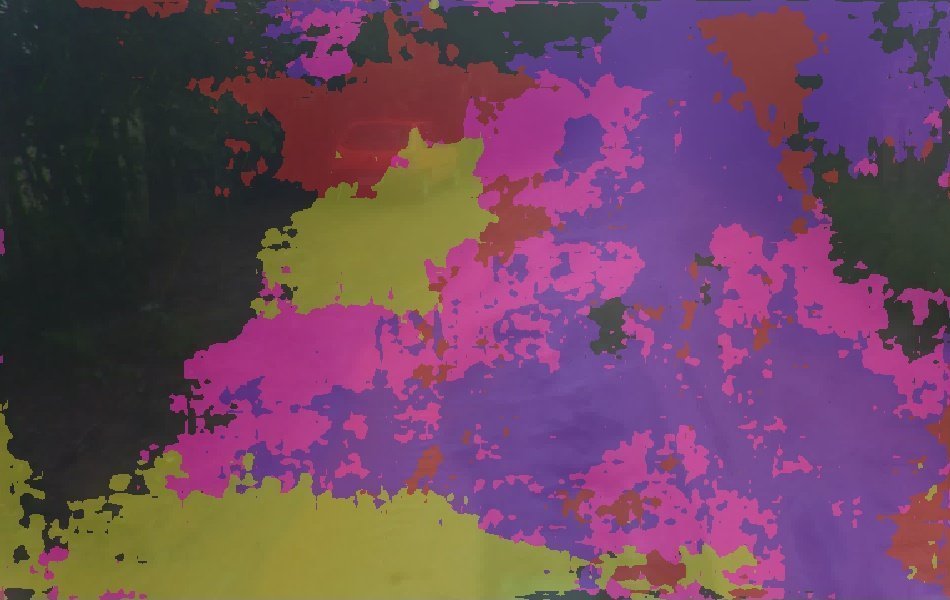}
	\end{subfigure}%
	
	\caption{Test of PSPNet and DeepLab with Cityscape on unpaved roads.}
	\label{fig:test_pspnet_deeplab_cityscape_unpaved_roads}
\end{figure}

It was carried out some tests with PSPNet \citep{Zhao:2017:PSPNet:ieeecvpr} and DeepLabV3 \citep{Chen:2018:deeplab:ieeeTPAMI} networks trained with the Cityspace \citep{cordts:2016:cityscapes} urban dataset to check the possibility of using these pre-trained networks in visual perception in off-road environments and unpaved roads (Fig. \ref{fig:test_pspnet_deeplab_cityscape_unpaved_roads}). It was possible to see that those systems currently being developed for autonomous vehicles may not be suitable for developing countries remaining restricted to a small set of roads in urban centers. This restriction limits even the implementation of autonomous systems in cargo vehicles, such as buses and trucks.

Given such a scenario, this work proposes a perception subsystem for autonomous vehicles and Advanced Driver Assistance Systems (ADAS) specialized in unpaved roads and off-road environments. Such a system uses Deep Learning and CNN to carry out the semantic segmentation of obstacles and the region boundaries where the vehicle can pass through. For this research, we have built an off-road track test exclusively for the development of the project and collected almost 12,000 images on several unpaved roads in urban and rural environments and annotated them. Furthermore, this work proposes a Configurable Modular Segmentation Network (CMSNet) framework, encompassing several innovations of modern architectures \citep{Chen:2018:EncoderDecoderWA:ECCV:deeplabv3Plus, Zhao:2017:PSPNet:ieeecvpr, Long:2015:FCN:ieeecvpr}, that allows choosing between different modules and backbones for feature extraction\citep{Sandler:2018:MobileNetV2:IEEE-CVF, He:2016:IEEECVPR-resnet, simonyan:2015:vgg}.  

It was investigated the behavior of the Deep Learning algorithm applied for semantic segmentation of off-road tracks in adverse visibility conditions, including rain and night. We have carried out the training of several CSMNet configurations to selecting an appropriate arrangement to segment obstacles and traffic areas. This research also verified how the accuracy and the inference time are affected, as well as made comparisons with other works trained with preexisting datasets to evaluate the impact of the proposed perception system.

The main contributions of this work are:
\begin{itemize}
	\item The proposal for a new dataset for unpaved roads and off-road environments containing several adverse visibility situations, such as rain, dust and poor lighting (night condition); 
	\item An investigation of the feasibility of applying deep learning to detect track limits where there is no clear delimitation between what is a road and what is not a road, as is the case with sandy off-road environments; 
	\item The proposition of a Configurable Modular Semantic Segmentation Neural Network (CMSNet) framework; 
	\item A study of how our proposed segmentation algorithms behave in differents level of visibility impairments severity; and 
	\item The evaluation of semantic segmentation architectures ported to embedded field application and having capability for real-time inference. 
\end{itemize}

This paper is organized as follows. Section \ref{sec:related-works} presents some works that are related to this research. Section \ref{sec:perception} describes the proposed CMSNet framework, summarizes the Kamino project and the hardware composition, and describes the construction of the proposed dataset. Section \ref{sec:perception} presents the experimental setups and the evaluations, including methodology, metrics of performance used to computerate the results, the results achieved in different conditions, and comparative with other works. Finally, section \ref{sec:conclusion} presents the conclusions.

\section{Related works}
\label{sec:related-works}
This section presents the main works related to this research grouped by themes and analyzes the relationships between those works and our proposal.

\begin{figure}[!t]
	\includegraphics[width=\linewidth]{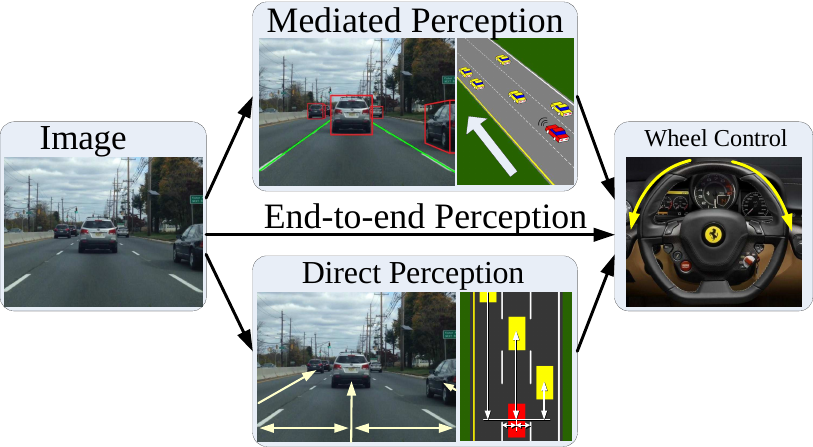}
	\caption{Perception paradigms. Adapted from \citep{chen:2015:deepdriving}}.
	\label{fig:paradigmas-de-percepcao}
\end{figure}

\textit{Perception for autonomous vehicles}.
\cite{brummelen:2018:autonomous-vehicle-perception} presented a review of the state-of-the-art concerning perception in autonomous vehicles. Among the types of perception shown (Fig. \ref{fig:paradigmas-de-percepcao}), mediated perception is the most used. The mediated perception only interprets sensor data to understand the scene while the planning and control module performs the remaining system functionalities. On the other hand, the end-to-end perception generates the control information to the vehicle straight from the data provided by the sensors, and the direct perception \citep{chen:2015:deepdriving}  maps the information received by the sensors into a set of key indicators related to the driving possibilities, given the current state of the track or traffic at that moment.  We decided to use mediated perception because it allows us to observe all processing steps instead of delegating the planning and control to a black-box algorithm.

\begin{figure}[!t]
	\includegraphics[width=\linewidth]{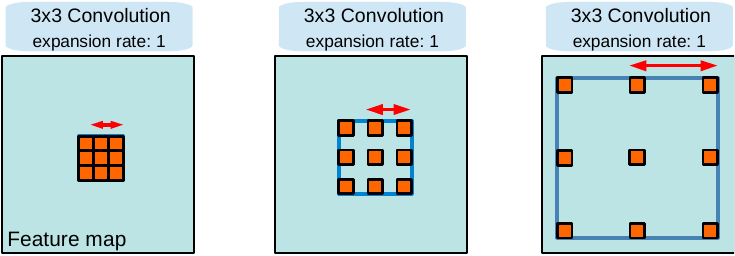}
	\caption{Dilated filter with different expansion rate. Adapted from \citep{DeepLabV3:DBLP:journals/corr/ChenPSA17}}
	\label{fig:atrous}
\end{figure}

\textit{Semantic and instance segmentation}.
The core of our research was the investigation of Deep Learning algorithms for semantic segmentation focused on the segmentation of unpaved roads and off-road environments. One of the architectures used as a basis for this investigation was the FCN \citep{Long:2015:FCN:ieeecvpr}. Such a work showed how to convert classification networks \citep{Krizhevsky:2012:ICD:alexnet,simonyan:2015:vgg,Szegedy:2015:inception-win-imagenet2014} into segmentation networks, and at the time of its publication had achieved 20\% improvement over previous work in the PASCAL VOC 2012 benchmark. There is also the work proposed by \cite{Zhao:2017:PSPNet:ieeecvpr}, that was responsible for applying the spatial pyramid pooling module in semantic segmentation to explore the global and regional context of the information contained in the images. This work was responsible for reaching the state-of-the-art accuracy of 85.4 \% in the PASCAL benchmark. Furthermore, the work proposed by \cite{DeepLavV1:CRFs:2015:ChenPKMY14} and \cite{Chen:2018:deeplab:ieeeTPAMI} applied atrous convolution on pixel-level classification allowing to enlarge the feature processing resolution (the field of view) and to keep the size of the filters stable. The \autoref{fig:atrous} shows expanded filters at different rates. Such work was also responsible for proposing Atrous Spatial Pyramid Pooling (ASPP) to perceive the context in images at different scales. This architecture managed to reach the mark of 79.7 \% 'mIoU' in the PASCAL dataset for semantic segmentation, and was updated to improve its accuracy in \cite{DeepLabV3:DBLP:journals/corr/ChenPSA17,Chen:2018:EncoderDecoderWA:ECCV:deeplabv3Plus}. Spatial Pyramid Pooling structures and Atrous Spatial Pyramid Pooling are modules available on our CMSNet.

\textit{Datasets}. 
In this work, we also have proposed a new dataset covering unpaved roads and off-road environments in adverse visibility conditions to enable the investigation of Deep Learning algorithms' behavior on the task of semantic segmentation in such circumstances. In another hand, most of the datasets published aim for urban situations. One of the first one was CamVid \citep{BROSTOW200988,brostowSFC:2008:ECCV08-camvid}.  This dataset has 32 classes and images captured from the driver's perspective, with more than 10 minutes of video collected at 30Hz and annotated at 1 Hz. It has 700 high-quality images manually labeled. Another one is the Kitti dataset \citep{kitti-geiger:2012:cvpr, kitti-geiger:2013:ijrr, kitti-fritsch:2013:itsc, kitti-menze:2015:cvpr} that contains several benchmarks, including semantic segmentation for roads. It also has stereo vision, 3D object detection, and tracking benchmarks. Also regard to paved urban environments, one of the most important datasets for semantic segmentation is Cityscapes \citep{cordts:2015:cvprw-cityscapes}. It contains stereo video sequences captured in 50 different cities with pixel-level labeling. Altogether there are 5,000 images precisely labeled and 20,000 images with coarse annotations. All of them, Kitti, CamVid, and Cityscapes, are used as a reference to analyze our proposed dataset.

\textit{Segmentation on off-roads environment}.
The work proposed by \cite{Real-Time-Semantic-Off-Road:Maturana:2018} is the one that most resembles the proposal of this research. In the same way as this research, they used RGB cameras and convolutional networks to distinguish what is the road limits and obstacles. They also built a dataset, but at the moment of our research, it was not available yet for comparison and evaluation. Another similar aspect was the concern with the inference time for allowing to embed the application with the segmentation occurring in real-time. There is also the work proposed in \cite{Semantic-Forested:Valada:2016} that has proposed an architecture using the VGG-16 \citep{simonyan:2015:vgg} as a backbone to extract features. Besides, it also presents a dataset for this type of environment. Both \cite{Semantic-Forested:Valada:2016}, and \cite{Real-Time-Semantic-Off-Road:Maturana:2018} evaluated their systems only in environments where there is a relative difference in texture/color on the track limits. Off-road test track such used in our proposed dataset adds a degree of complexity as they do not present a notable difference between what is or is not the region where the car may pass through, with both parts being made up of sand and having the same color. The \autoref{subsection:results-on-deepscene-dataset} shows comparison results of our proposed architecture and those works.

\section{Perception in off-road environments}
\label{sec:perception}
Within the scope of this research was used the semantic segmentation technique to carry out the task of finding the track limits. Semantic segmentation is the task that assigns classification at the pixel level by grouping them as belonging to the same object. The advantage of this approach is that in addition to segmenting the road limits, it can also discover and segment obstacles on the road, eliminating in some cases the use of a second network for object detection. 

\subsection{CMSNet}

The CMSNet is the framework proposed in this research that allows configuring several architectures with modules commonly used in state-of-the-art deep neural networks for semantic segmentation. Also, it is capable of operating with different backbones for feature extraction. 

\textit{Backbone}.
Choosing the backbone suitable for the target application is an important step. There are variations of architectures capable of achieving accuracy above 98\% in the Top-5 and 88\% in the Top-1 in the ImageNet benchmark \citep{russakovsky:2015:imagenet, EfficientNet:pmlr:2019:tan19a}. However, when building a perception system based on Deep Learning for real-time inference, in addition to accuracy, it is necessary to take into account the latency of the backbone. Thus, the choice of network for extracting features must consider architectural aspects that offer a cost-benefit ratio between accuracy and latency. The authors of this work have chosen the MobileNetv2 architecture \citep{Sandler:2018:MobileNetV2:IEEE-CVF} as the main backbone for feature extraction in such research because it demands low computational power compared to other architectures in the same level of accuracy. Besides, the CMSNet also supports ResNet and VGG as backbone  \citep{simonyan:2015:vgg, He:2016:IEEECVPR-resnet}.

The \cite{Sandler:2018:MobileNetV2:IEEE-CVF} architecture, in its standard version, has 3.5 million parameters and has a computational cost of 300 million of Multiply–accumulate (MAC) operation. It uses Depthwise Separable Convolutions and a residual block structure with a bottleneck.  We have slightly modified it by removing the latest convolution and pooling layers. Such change decreased the total number of parameters from 3.5 million to 1.84 million — approximately 48\% fewer parameters. The \autoref{tab:mobilenet_architecture_os16} shows the final configuration for output strides 16 and 8 (OS16 and OS8), where \textit{{h}} is the height, \textit{{w}} is the width, \textit{{c}} is the number of channels, \textit{{e}} is the expansion factor for each block, \textit{{d}} is the input dimension, \textit{{n}} indicates the block repetition, and \textit{{s}} defines the stride.

\begin{table}[!t]
    \centering
    \scriptsize
	\caption{Adapted MobilenetV2 architecture for OS16 and OS8, where \textit{{h}} is the height, \textit{{w}} is the width, \textit{{c}} is the number of channels, \textit{{e}} is the expansion factor for each block, \textit{{d}} is the input dimension, \textit{{n}} indicates the block repetition, and \textit{{s}} defines the stride.}
	\label{tab:mobilenet_architecture_os16}
	\begin{tabular}{r+r+r+r+r+c+c+c+c+c} 
	    \specialrule{.1em}{.05em}{.05em}
		\multicolumn{2}{c}{\bfseries OS16} & \multicolumn{2}{c}{\bfseries OS8} & \multicolumn{1}{c}{\multirow{2}*{\bfseries \textit{c}}} &  \multirow{2}*{\bfseries Operador} & \multirow{2}*{\bfseries \textit{e}} & \multirow{2}*{\bfseries \textit{d}} & \multirow{2}*{\textit{n}} & \multirow{2}*{\bfseries \textit{s}} \\ 
		\cline{0-3} 
		\multicolumn{1}{c}{\bfseries \textit{h}} & \multicolumn{1}{c}{\bfseries \textit{w}} &\multicolumn{1}{c}{\bfseries \textit{h}} & \multicolumn{1}{c}{\bfseries \textit{w}} &  &  &  &  &  &  \\
		
		\specialrule{.1em}{.05em}{.05em} 
		 483&769& 483&769&   4   & conv2d              & - & 32   & 1 & 2 \\ 
		 242&385& 242&385&  32  & \textit{bootleneck} & 1 & 16   & 1 & 1 \\ 
		 242&385& 242&385&  16  & \textit{bootleneck} & 6 & 24   & 2 & 2 \\ 
		 121&192& 121&192&  24  & \textit{bootleneck} & 6 & 32   & 3 & 2 \\ 
		  61& 97&  61& 97&  32  & \textit{bootleneck} & 6 & 64   & 4 & 2 \\ 
		  31& 49&  61& 97&  64  & \textit{bootleneck} & 6 & 96   & 3 & 1 \\ 
		  31& 49&  61& 97&  96  & \textit{bootleneck} & 6 & 160  & 3 & 1 \\ 
		  31& 49&  61& 97& 160  & \textit{bootleneck} & 6 & 320  & 1 & 1 \\ 
		\specialrule{.1em}{.05em}{.05em}
	\end{tabular}
\end{table}

\textit{Semantic segmentation architecture}.
In addition to the backbone for extracting features, it is necessary to build structures responsible for performing the core activity — e.i., carrying out the pixel-level classification. There are several network architecture proposals for semantic segmentation. However, we have considered only a few ones in the scope of this work to have their characteristics and innovations analyzed \citep{Long:2015:FCN:ieeecvpr, Zhao:2017:PSPNet:ieeecvpr, Chen:2018:EncoderDecoderWA:ECCV:deeplabv3Plus}. These architectures presented significant and complementary contributions in the field of semantic segmentation so that different solutions can be proposed and tested based on them. These arrangements were the basis for the construction of the configurable modular framework (CMSNet) proposed and developed in this research. The CMSNet allows several configurations by enabling or removing some structures, as described in the following.

\textit{Shortcut}.
\label{subsection:shotcut-strategy}
In the architecture proposed by \cite{Long:2015:FCN:ieeecvpr}, the latest step is responsible for generating the segmentation mask in an appropriate size. Such a result is achieved by the upsampling of the activation map on the last layer of the network. It uses a transposed convolution (deconvolution) to perform interpolation and generate the output image. Instead of using linear interpolation with fixed parameters, this layer can learn the best way to interpolate the output producing the most suitable segmentation mask.

\begin{figure}[t!] 
	\begin{center} 
		\includegraphics[width=0.75\linewidth]{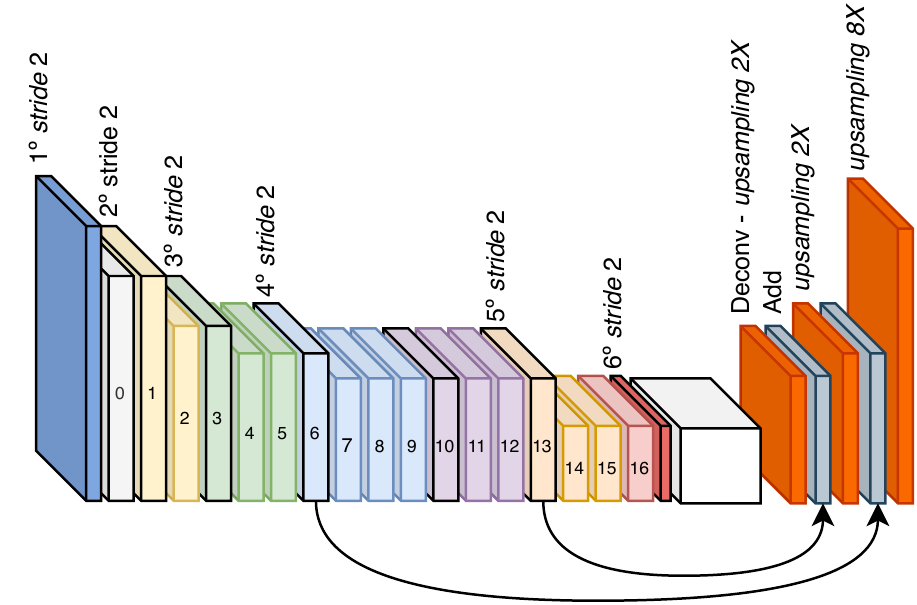} 
	\end{center} 
	\caption{Shortcut strategy.} 
	\label{fig:shortcut} 
\end{figure}

The upsampling can be done in a single step or by multiple ones to improve detailing.  When performed in more than one stage, after each resizing, a shortcut is used to add the most external features to the result before the next resizing. This shortcut helps to improve the detail of the segmentation (Fig. \ref{fig:shortcut}).

Shortcuts are one of the options available in the configurable modular architecture proposed in our research. It can be enabled or disabled on its configurations.

\textit{Scene analysis by Spatial Pyramid Pooling (SPP)}.
Although fully convolutional networks \citep{Long:2015:FCN:ieeecvpr} performed well in semantic segmentation, they have difficulty to take into account the global context during the analysis of each pixel \citep{Zhao:2017:PSPNet:ieeecvpr}. This difficulty can lead to incorrect classification, as it does not consider the appropriate relationships between classes, e.i., confusing pixels of the track with the background since both contain sand of the same color.

\begin{figure}[t!] 
	\begin{center} 
		\includegraphics[width=0.75\linewidth]{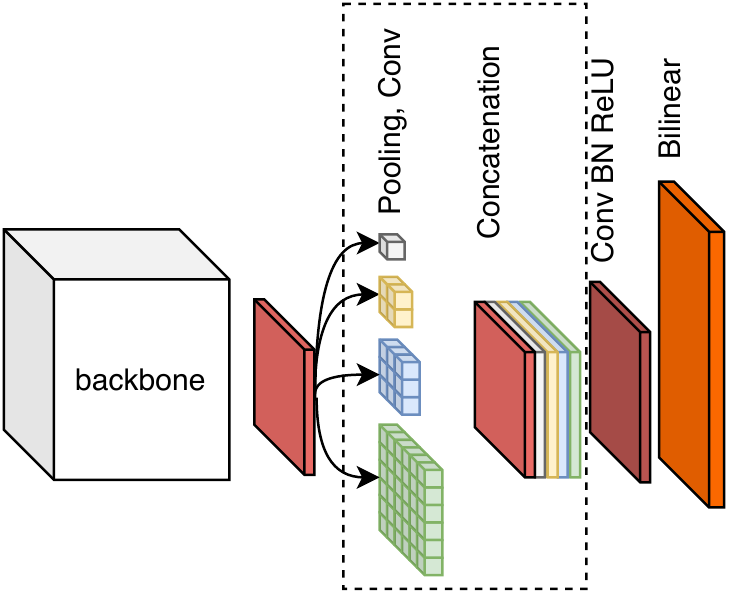} 
	\end{center} 
	\caption{Spatial Pyramid Pooling Module} 
	\label{fig:pyramid-pooling-module} 
\end{figure}

The network architecture proposed in \cite{Zhao:2017:PSPNet:ieeecvpr} adds a module formed by a pyramid of pooling layers, followed by convolution and concatenation (Fig. \ref{fig:pyramid-pooling-module}). This structure is capable of providing scene analysis at different scales, allowing to infer the contribution of global or local context in the classification of each pixel, and mitigating the consequences of the lack of context analysis found in the \cite{Long:2015:FCN:ieeecvpr}. In this module, each pooling of the pyramid comes with a pointwise convolution having $d/N$ filters, where N represents the pooling size, and $d$ represents the input channels at the convolution.

Like the shortcuts, the spatial pyramid pooling is present in the CMSNet framework proposed in this work.  It uses four average pooling with different compressions rate: the first is global pooling, the second is 1/2 of the resolution of the features, the third is 1/3 of the height and width of the features block, and the fourth is 1/6 of the resolution block.  All of these values are concatenated with the original data to go through another convolution, \autoref{fig:pyramid-pooling-module}.

\textit{Dilated convolution}.
\label{subsection:atrous-convolution}
The standard convolution followed by pooling, as used in FCN, increases the output stride and reduces the size of the feature map on the output of the networks' deepest layers. That is interesting to allow extending the field of view of the filter and improve the ability to observe the context without need larger filters that increase the computational cost. However, narrowing the feature map through consecutive strides is harmful to semantic segmentation as it causes the loss of spatial information in the deeper layers of the backbone \citep{Chen:2018:deeplab:ieeeTPAMI,DeepLavV1:CRFs:2015:ChenPKMY14}. A solution to this problem may be the use of atrous convolution, which allows keeping the size of the feature map (stride) constant and arbitrarily control the field of view without increasing the number of network parameters or computational cost \citep{Chen:2018:deeplab:ieeeTPAMI,DeepLavV1:CRFs:2015:ChenPKMY14}. That makes it possible to achieve larger maps of features in the output and supports semantic segmentation.
 
CMSNet always uses extended convolution. However, it is possible to configure whether it will start from the place where there would be the 4th or 5th stride pooling, generating features outputs with a stride of 16 (1/16 of the size of the input image) or 8 (1 / 8 of the input resolution) respectively.

\begin{figure}[t!] 
	\begin{center} 
		\includegraphics[width=0.75\linewidth]{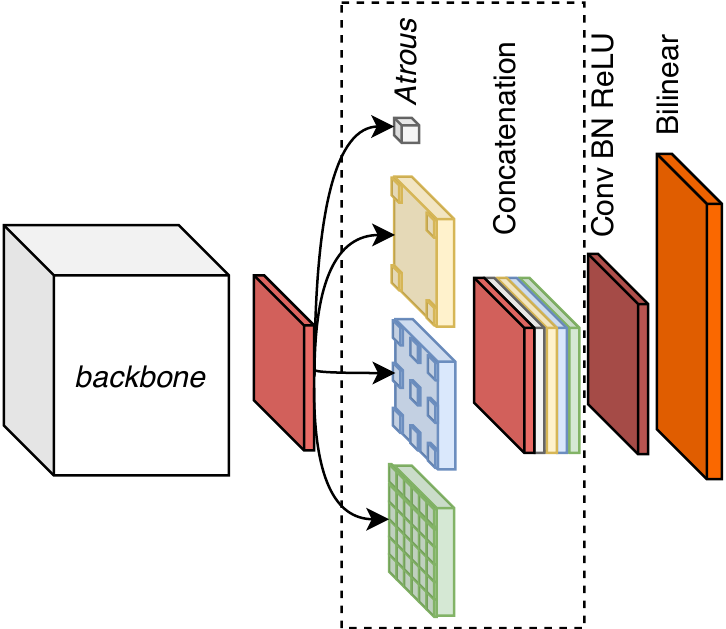} 
	\end{center} 
	\caption{Atrous Spatial Pyramid Pooling Module} 
	\label{fig:atrous-pyramid-pooling-module} 
\end{figure}

\textit{Atrous Spatial Pyramid Pooling (ASPP)}.
\label{subsection:atrous-pooling}
Just like the SPP module (Fig. \ref{fig:pyramid-pooling-module}), it is also possible to improve the understanding of the global and local context of the scene through the application of a pyramid module formed by dilated convolution -- atrous spatial pyramid pooling (Fig. \ref{fig:atrous-pyramid-pooling-module}). This structure helps to segment objects considering the context at different scales by applying filters with various sample rates \citep{Chen:2018:deeplab:ieeeTPAMI, DeepLabV3:DBLP:journals/corr/ChenPSA17, Chen:2018:EncoderDecoderWA:ECCV:deeplabv3Plus}. In the CMSNet framework presented in this work, we have used the ASPP module with expansion rates of 1, 6, 12, or 18  for output stride 16, end and expansion rates of 1, 12, 24, or 36 for output stride 8.

\begin{figure}[h!] 
	\begin{center} 
		\includegraphics[width=0.75\linewidth]{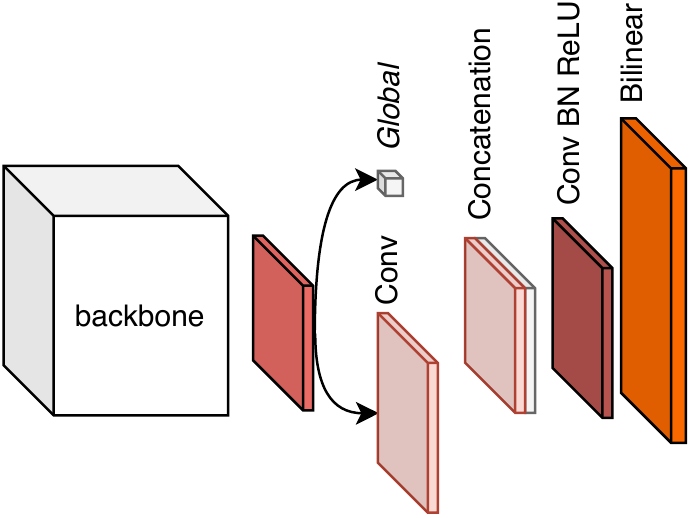} 
	\end{center} 
	\caption{Global Pyramid Pooling Module.} 
	\label{fig:global-pooling-module} 
\end{figure}

\textit{Global Pyramid Pooling (GPP)}.
\label{subsection:global-pooling}
Even using separable convolution, pyramid pooling modules introduce a computational overhead. To deal with this, \cite{Sandler:2018:MobileNetV2:IEEE-CVF} and \cite{DeepLabV3:DBLP:journals/corr/ChenPSA17} proposed a global pyramid pooling to provide a cost-effective global context analysis for semantic segmentation. This solution uses just one global pooling concatenated with a pointwise convolution (Fig. \ref{fig:global-pooling-module}).

\textit{Bilinear Interpolation}.
\label{subsection:bilinear-interpolation} 
In the pyramids methods, SPP, ASPP, and GPP, convolution followed by bilinear interpolation is used instead of transposed convolution. The transposed convolution is computationally less efficient, and its results are equivalent to these two operations together. Both functions have the purpose of learning the best way to interpolate the low-resolution segmentation maps and resize them to the image size. The CMSNet supports only the convolution followed by bilinear interpolation. We opted not to use the transposed convolution to keep the computational cost consistent.

\begin{figure}[h!] 
	\begin{center} 
		\includegraphics[width=\linewidth]{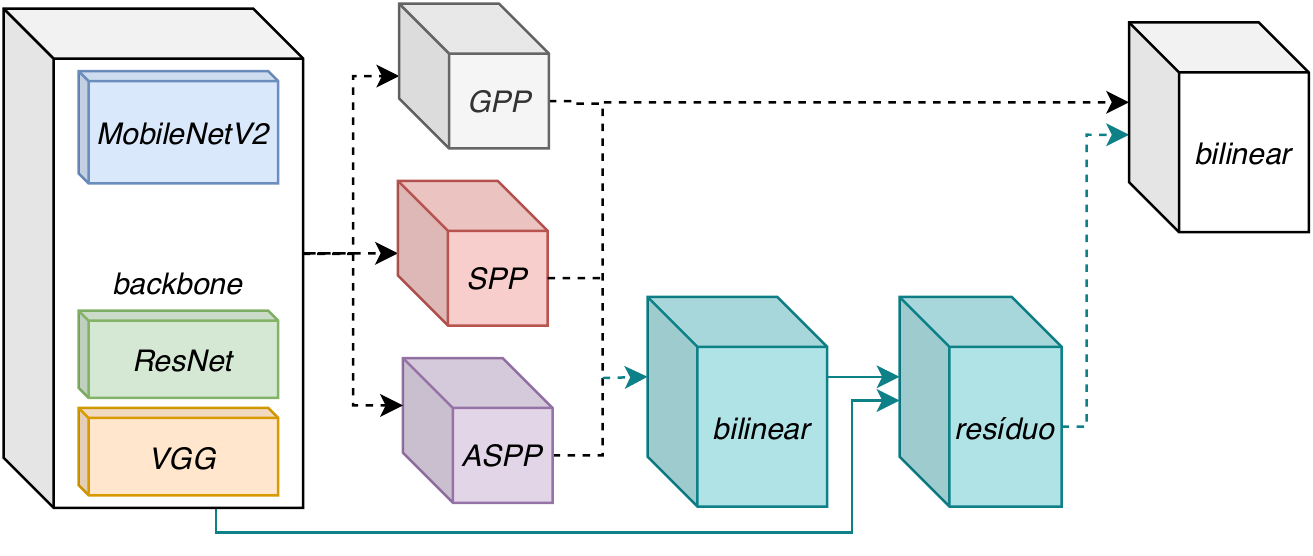} 
	\end{center} 
	\caption{CMSNet framework.} 
	\label{fig:modular-configurable-architecture} 
\end{figure}

\textit{CMSNet framework}.
\label{subsection:nossa-arquitetura} 
The framework proposed in our research allows compounding different variations of architectures to carry out tests on various innovation arrangements and to compare the latency and accuracy results achieved for the target application. In this case, the goal is to segment areas where the car can pass through and obstacles on unpaved roads and off-road environments in various visibility conditions such as day, night, dust, and rain. The \autoref{fig:modular-configurable-architecture} shows the components of proposed CMSNet framework. It can be configured by parameter to operate with the backbones MobileNetV2 \citep{Sandler:2018:MobileNetV2:IEEE-CVF}, ResNet \citep {He:2016:IEEECVPR-resnet} or VGG \citep{simonyan:2015:vgg} supporting output stride 8 or 16. It is also possible to choose between the GPP, SPP, or ASPP pyramid modules, as well as enabling the shortcut with the output stride is set to 16.

\subsection{Kamino project}
Even using a backbone optimized for computational efficiency, CNNs for dense pixel classification demands high parallel processing power and memory bandwidth. These requirements create problems in the moment of embedding the perception subsystem for field tests with real-time inference. One possible way to do this would be building dedicated hardware using FPGA or ASIC, but such solutions are highly complex to implement and may not be flexible concerning changes. The solution used in this research was porting the subsystem to the NVIDIA DrivePX 2 Autochauffeur. Nevertheless, once the network was developed on an x86\_64 platform, it was necessary to reimplement it with C++/CUDA merging several layers to be able to run it in real-time on the ARMv8-A.

\begin{figure}[!t]
	\includegraphics[width=\linewidth,trim={0cm 2cm 5cm 8cm},clip]{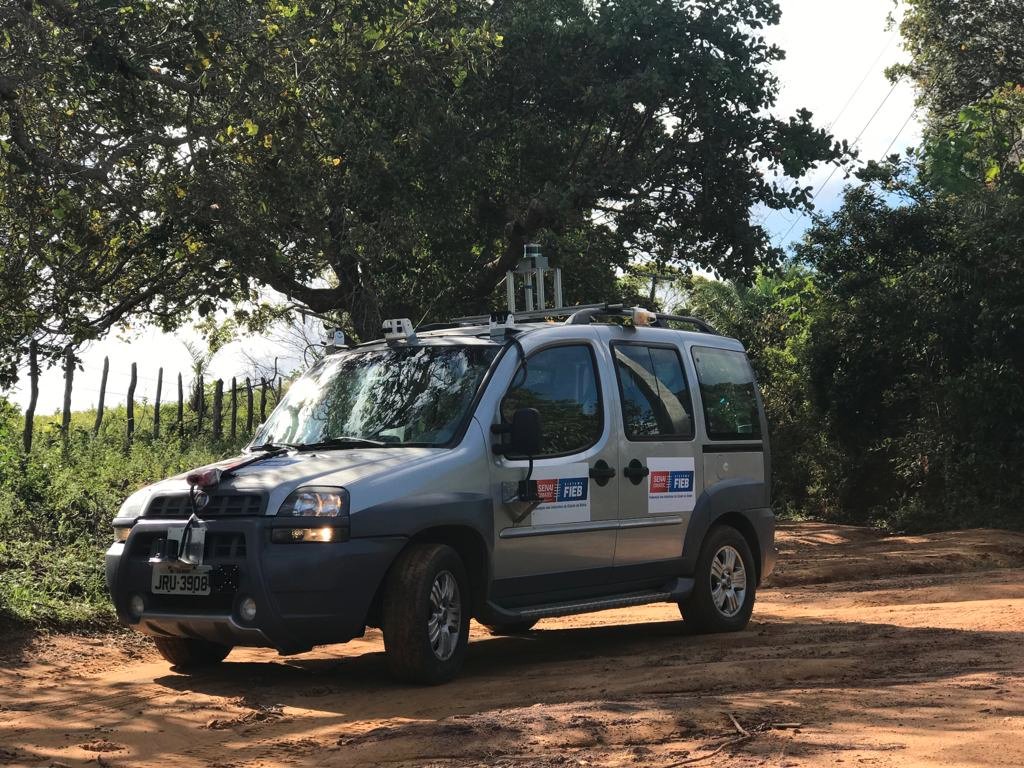}
	\caption{The vehicle used for data acquisition and validation of the proposed system.}
	\label{fig:carro-plataforma}
\end{figure}

\begin{figure}[!t]
	\includegraphics[width=\linewidth]{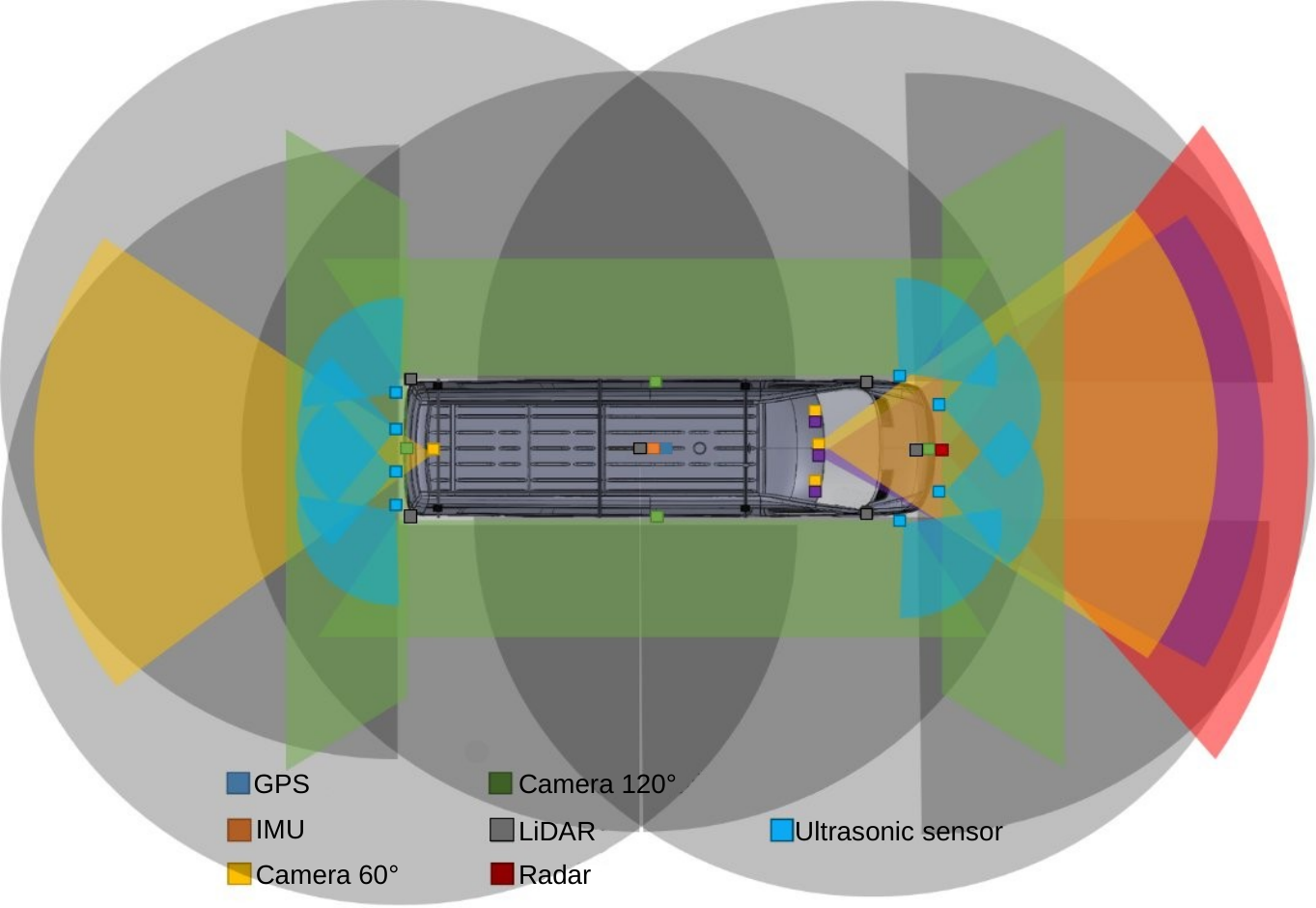}
	\caption{Sensor layout and operating region.}
	\label{fig:montagem-de-sensores}
\end{figure}

\textit{The hardware}.
A utility van (Fig. \ref{fig:carro-plataforma}) was used to mount the hardware for data acquisition and system validation. The system was composed of four RGB cameras with 60º Field of View (FOV), four RGB cameras with 120º FOV, one 16-beam LiDAR, four 8-beam LiDARs, eight ultrasonic sensors, one Inertial Measurement Unit (IMU), one GPS, one Radar, and one DrivePX 2 (Fig. \ref{fig:montagem-de-sensores}).

\subsection{Kamino dataset}
\label{subsection:kamino_dataset}
Unpaved roads represent a scenario relatively unexplored for the insertion of autonomous vehicle technology. The authors of this research created a dataset for off-road and unpaved roads to overcome this situation. The dataset developed has images collected in different environments, including a test track built to emulate off-road environments and adverse conditions such as at night, rainy, and dusty environments.

\begin{figure}[!t]
	\begin{subfigure}[b]{0.45\linewidth}
		\includegraphics[width=\linewidth,trim={0cm 0cm 0cm 10cm},clip]{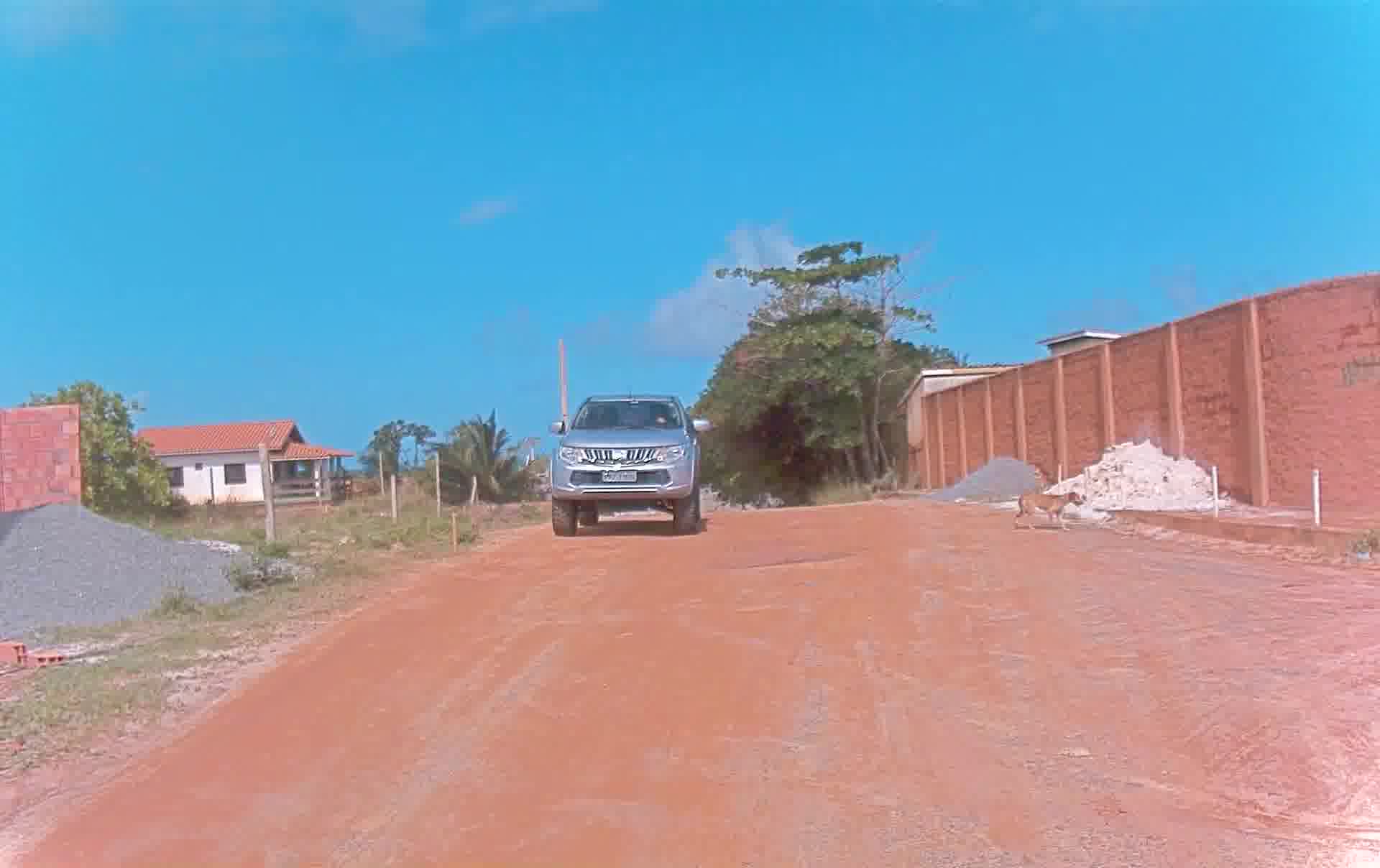}
		\caption{Jauá.}
	\end{subfigure}%
	\hfill%
	\begin{subfigure}[b]{0.45\linewidth}
		\includegraphics[width=\linewidth,trim={0cm 5cm 0cm 5cm},clip]{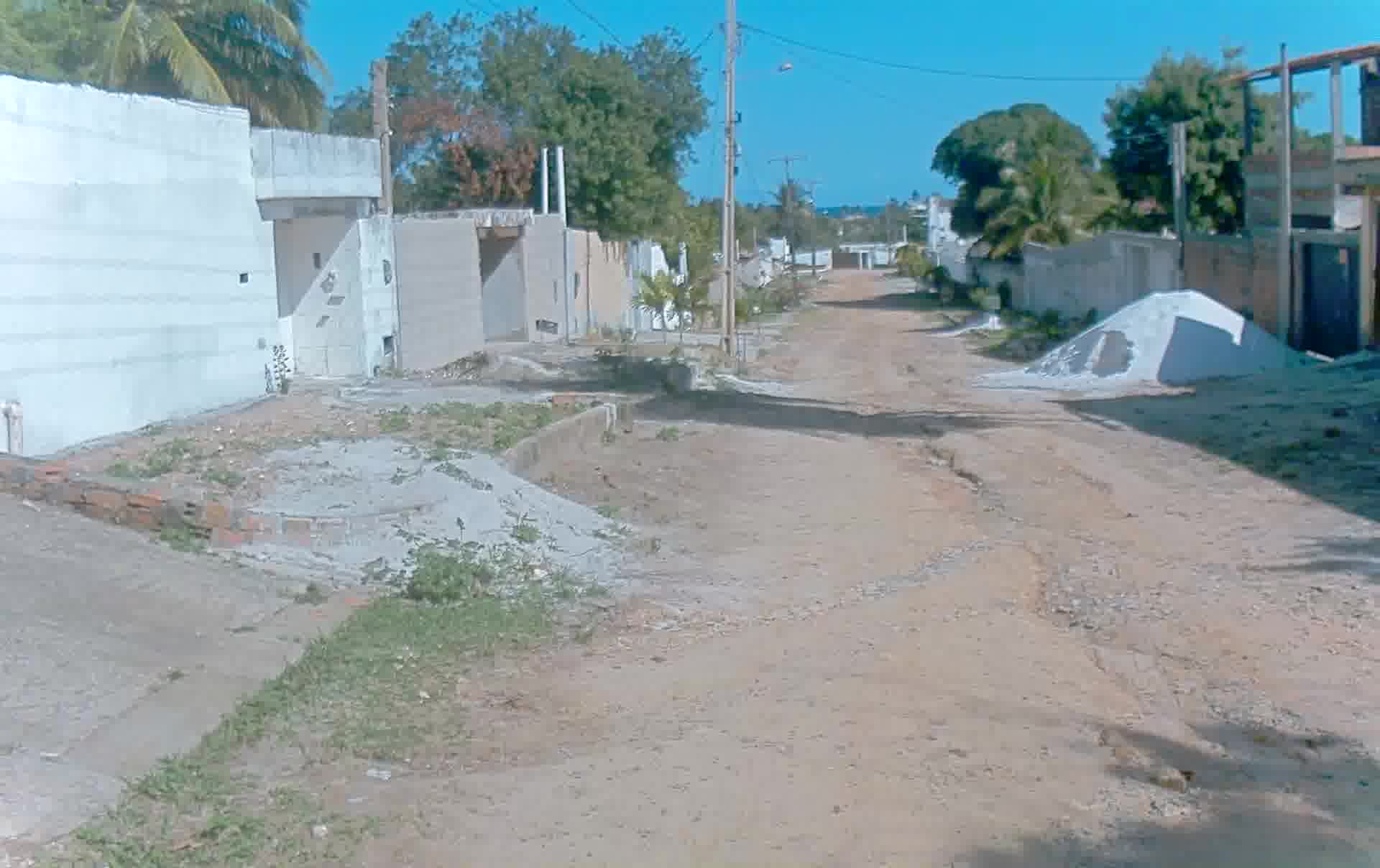}
		\caption{Jauá.}
	\end{subfigure}
	\begin{subfigure}[b]{0.45\linewidth}
		\includegraphics[width=\linewidth,trim={0cm 5cm 0cm 5cm},clip]{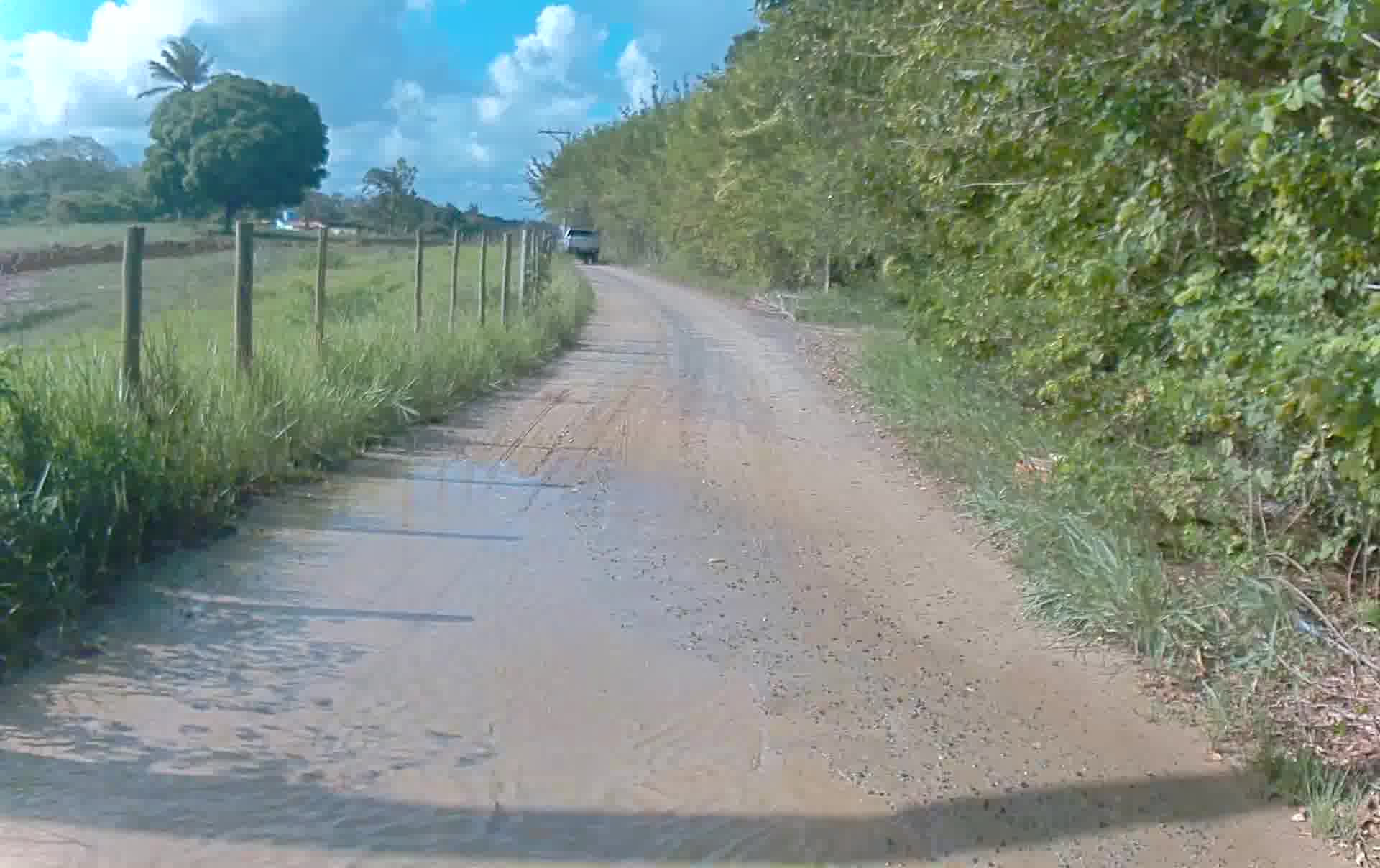}
		\caption{Estrada dos Tropeiros.}
	\end{subfigure}%
	\hfill%
	\begin{subfigure}[b]{0.45\linewidth}
		\includegraphics[width=\linewidth,trim={0cm 5cm 0cm 5cm},clip]{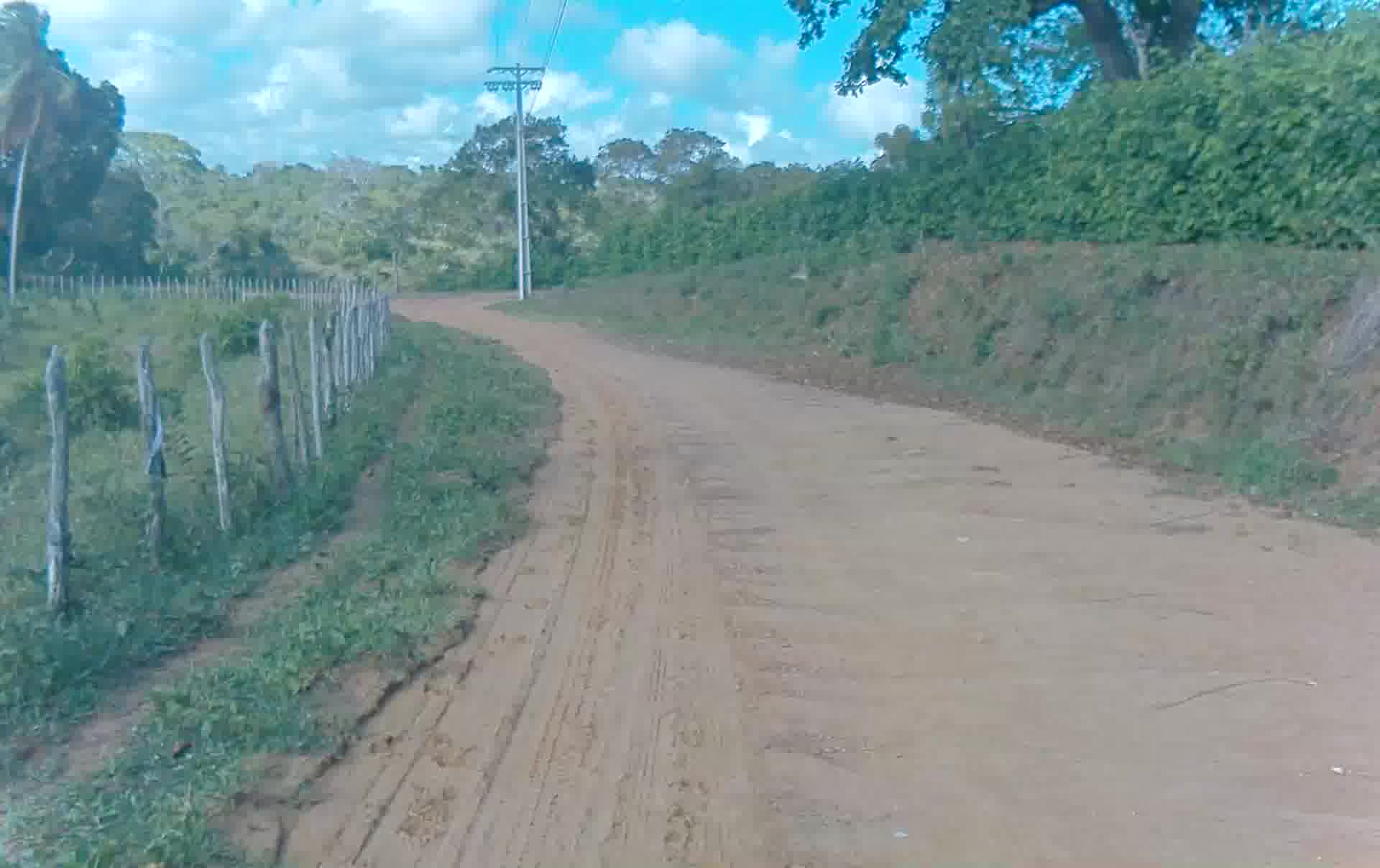}
		\caption{Estrada dos Tropeiros.}
	\end{subfigure}
	\begin{subfigure}[b]{0.45\linewidth}
		\includegraphics[width=\linewidth,trim={0cm 5cm 0cm 5cm},clip]{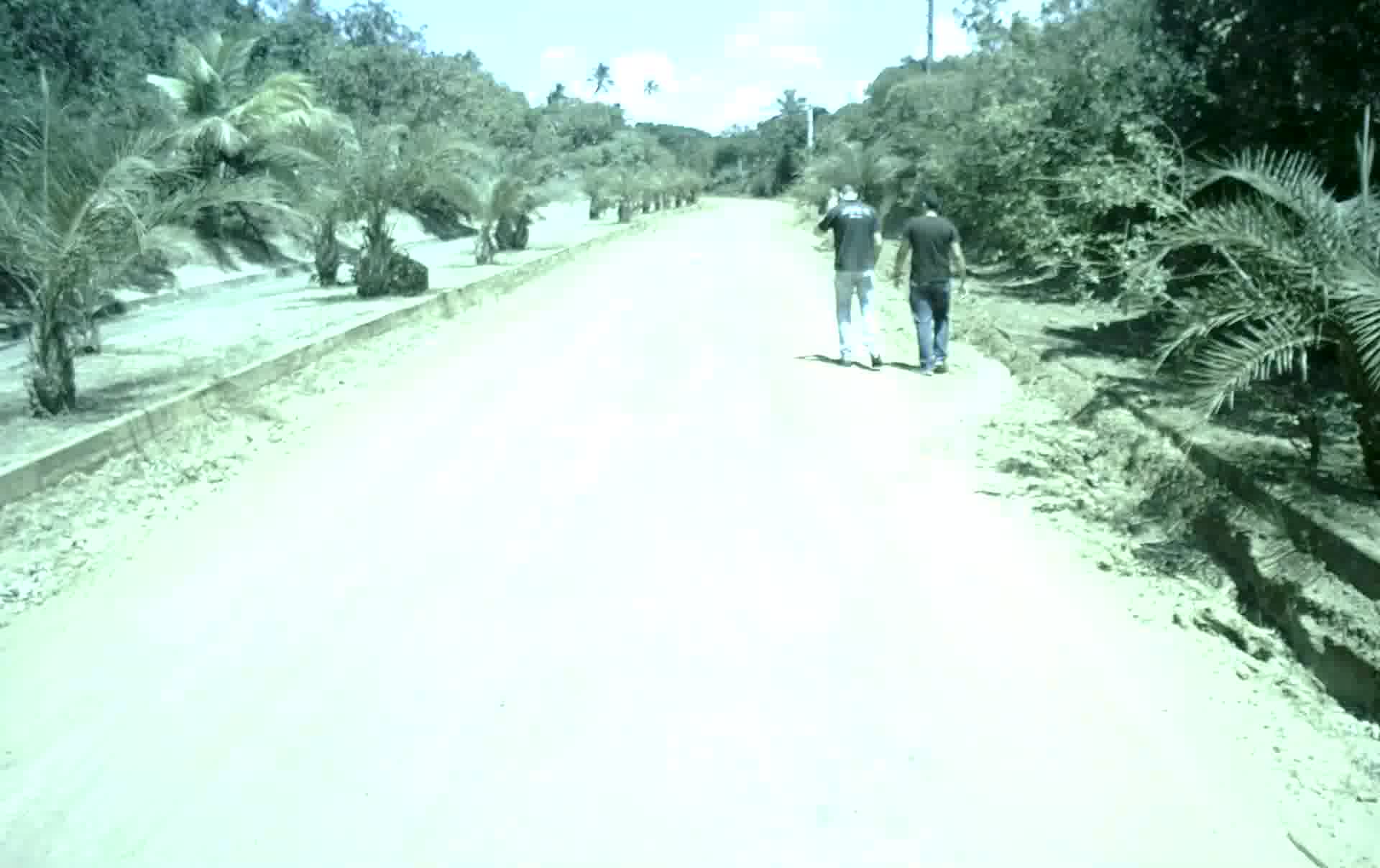}
		\caption{Praia do Forte.}
	\end{subfigure}%
	\hfill%
	\begin{subfigure}[b]{0.45\linewidth}
		\includegraphics[width=\linewidth,trim={0cm 7cm 0cm 3cm},clip]{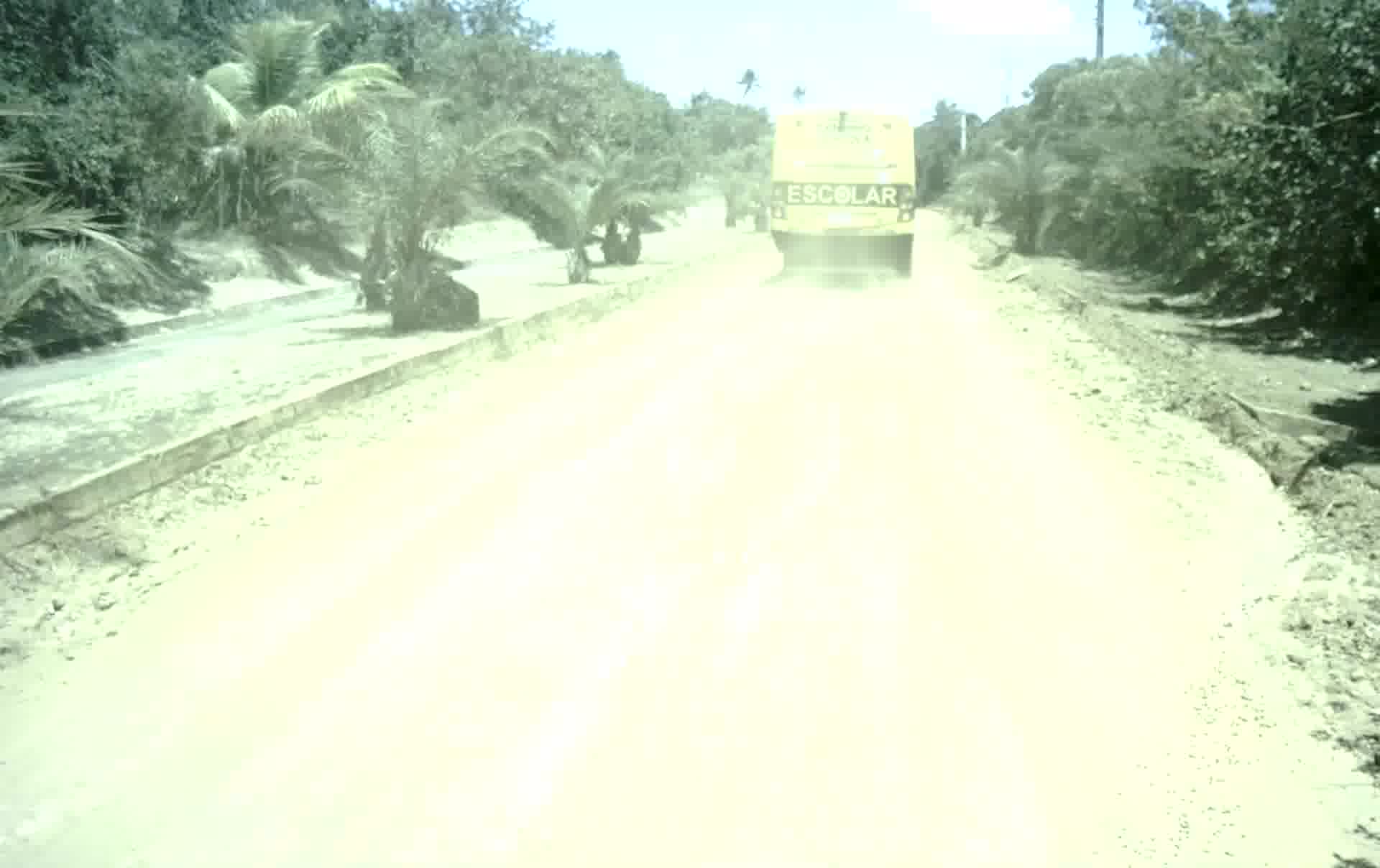}
		\caption{Praia do Forte.}
	\end{subfigure}
	\caption{Images collected in the metropolitan region of Salvador.}
	\label{fig:imagens-coletados-na-regiao-metropolitana-de-salvados-ba}
\end{figure}

\textit{Setup}.
We mounted a hardware platform with various sensors for collecting many hours of data. Subsequently, the most relevant pieces of information were selected and converted into frames at 1 or 5 FPS. Further, we accurately labeled the images resulting from that process. Several unpaved roads in the metropolitan region of Salvador-BA were used as the scenario for data capturing (Fig. \ref{fig:imagens-coletados-na-regiao-metropolitana-de-salvados-ba}). That includes the north coast of Bahia state and the track built to simulate off-road environments.

\begin{figure}[!t]
	\begin{subfigure}[b]{0.32\linewidth}
		\includegraphics[width=\linewidth,trim={2cm 0cm 0cm 2cm},clip]{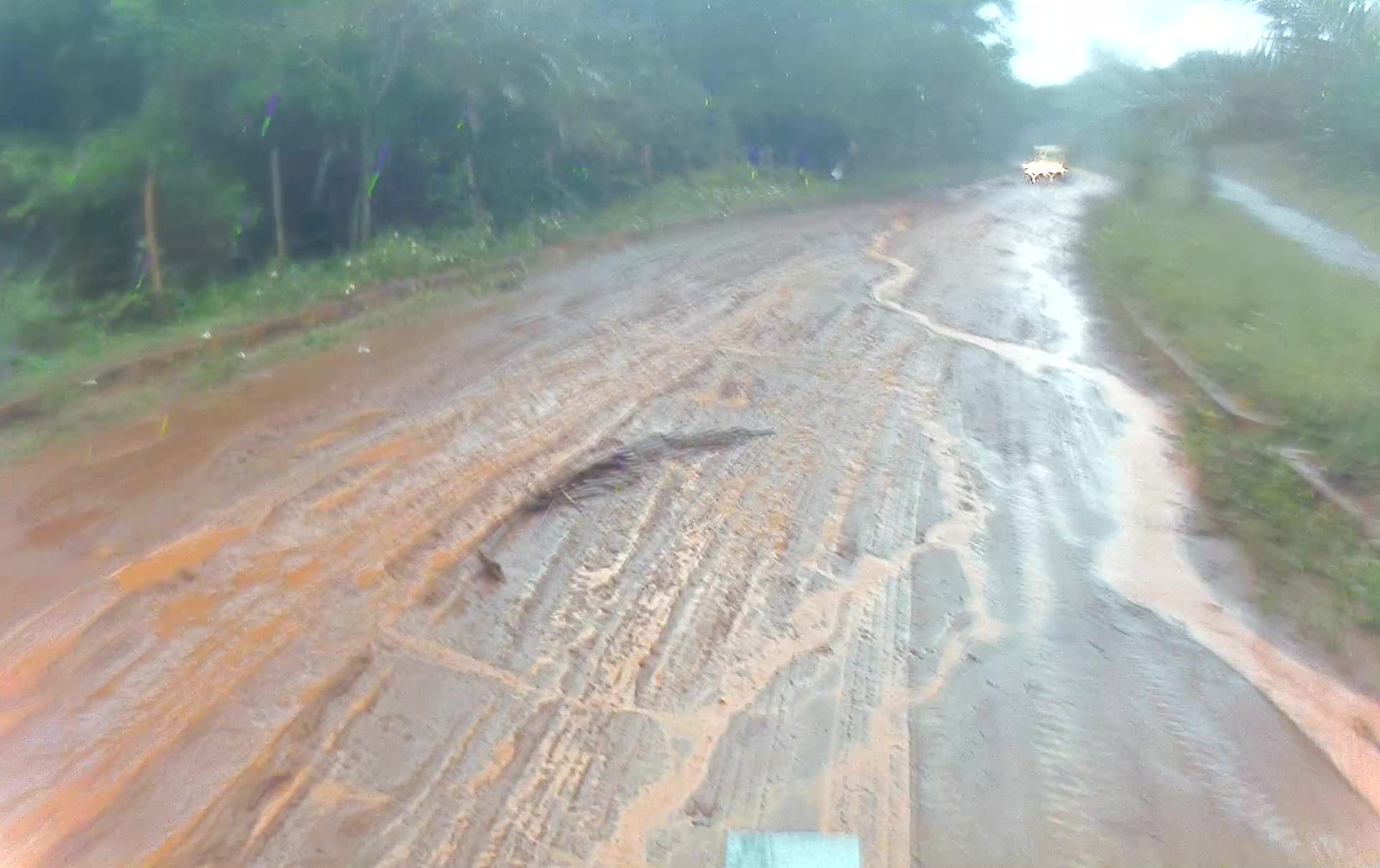}
		\caption{P. do Forte raining.}
	\end{subfigure}%
	\hfill
	\begin{subfigure}[b]{0.32\linewidth}
		\includegraphics[width=\linewidth,trim={2cm 0cm 0cm 2cm},clip]{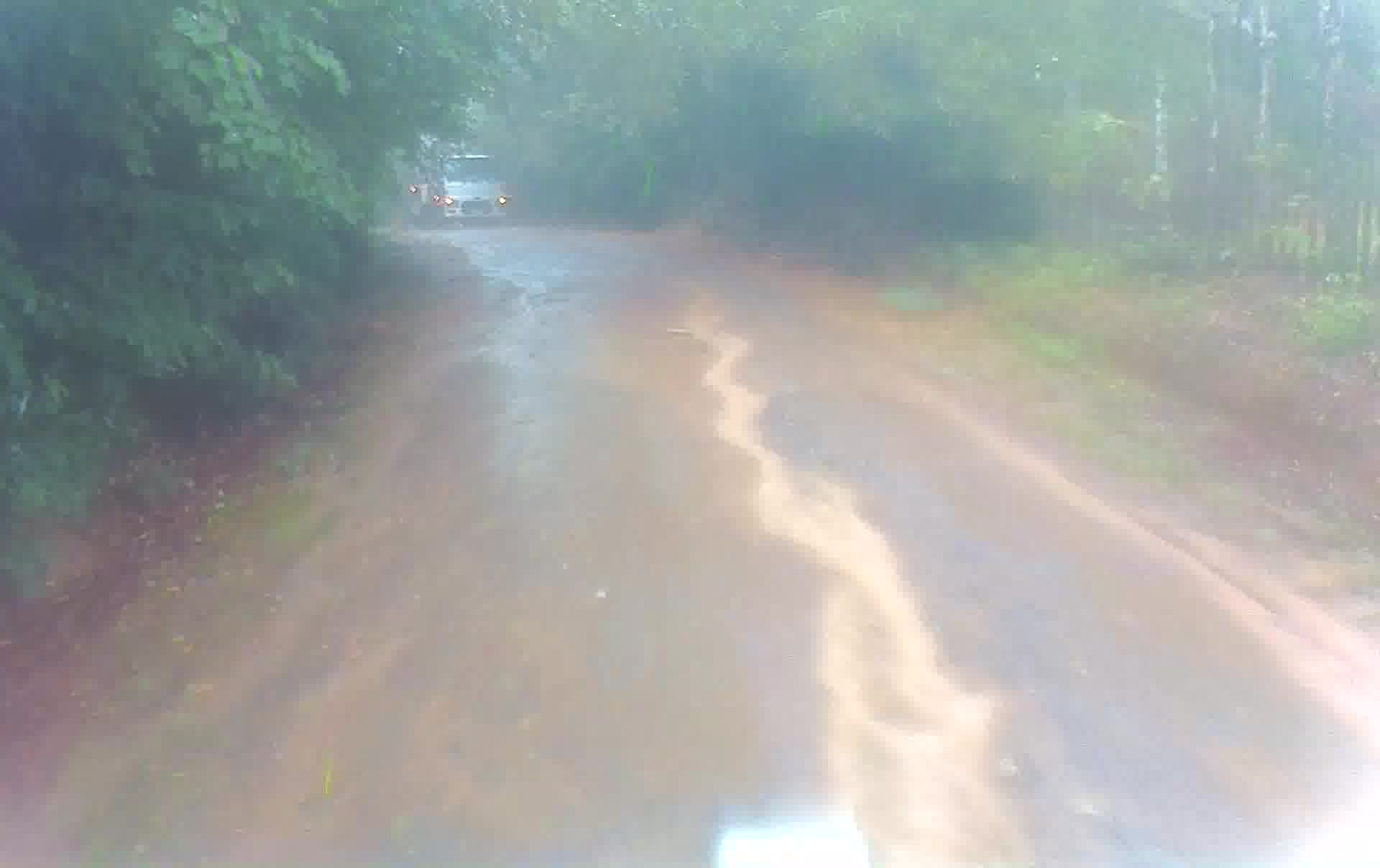}
		\caption{Tropeiros raining.}
	\end{subfigure}
	\hfill
	\begin{subfigure}[b]{0.32\linewidth}
		\includegraphics[width=\linewidth,trim={2cm 0cm 0cm 2cm},clip]{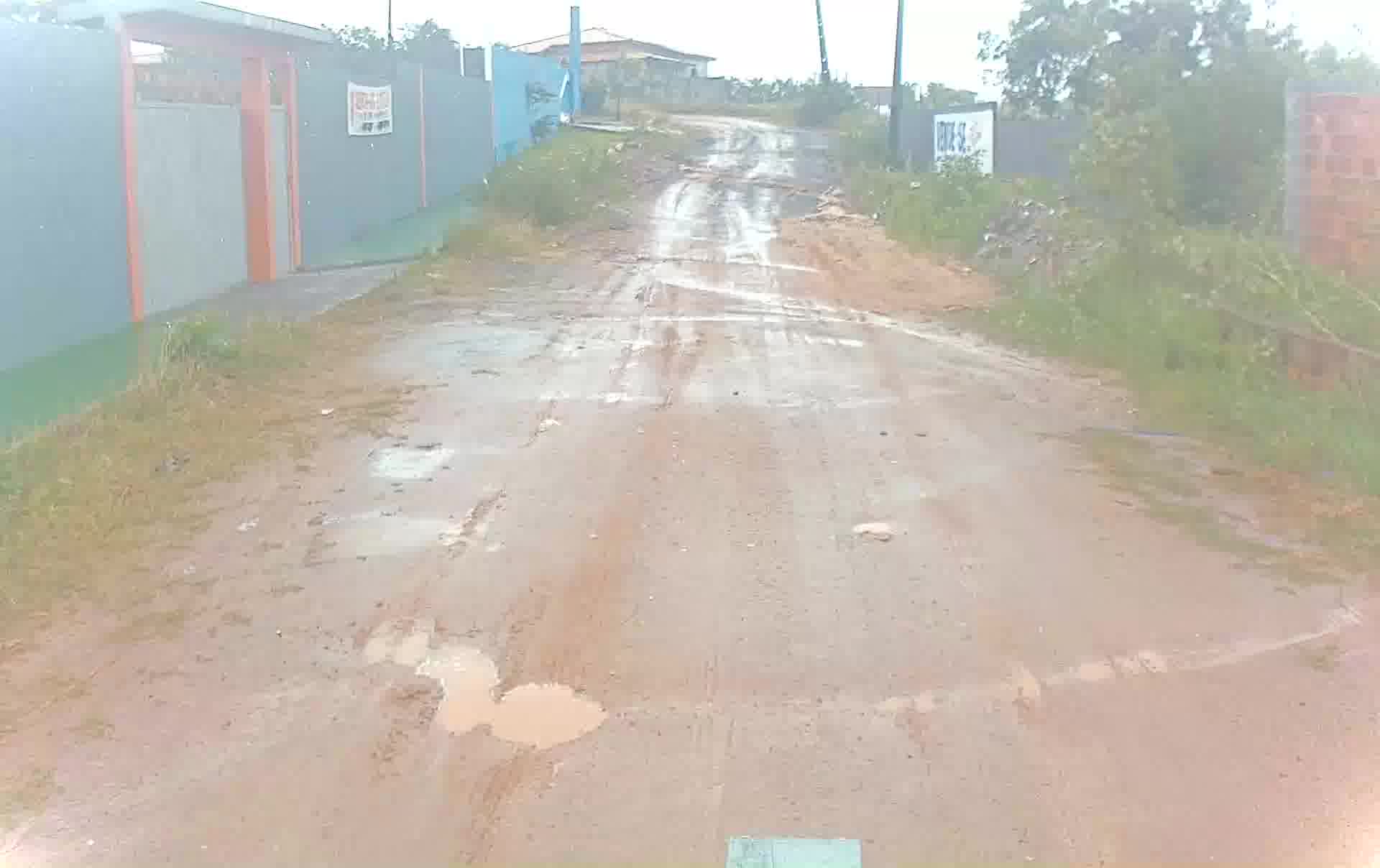}
		\caption{Jauá raining.}
	\end{subfigure}%
	
	\begin{subfigure}[b]{0.32\linewidth}
		\includegraphics[width=\linewidth,trim={2cm 0cm 0cm 2cm},clip]{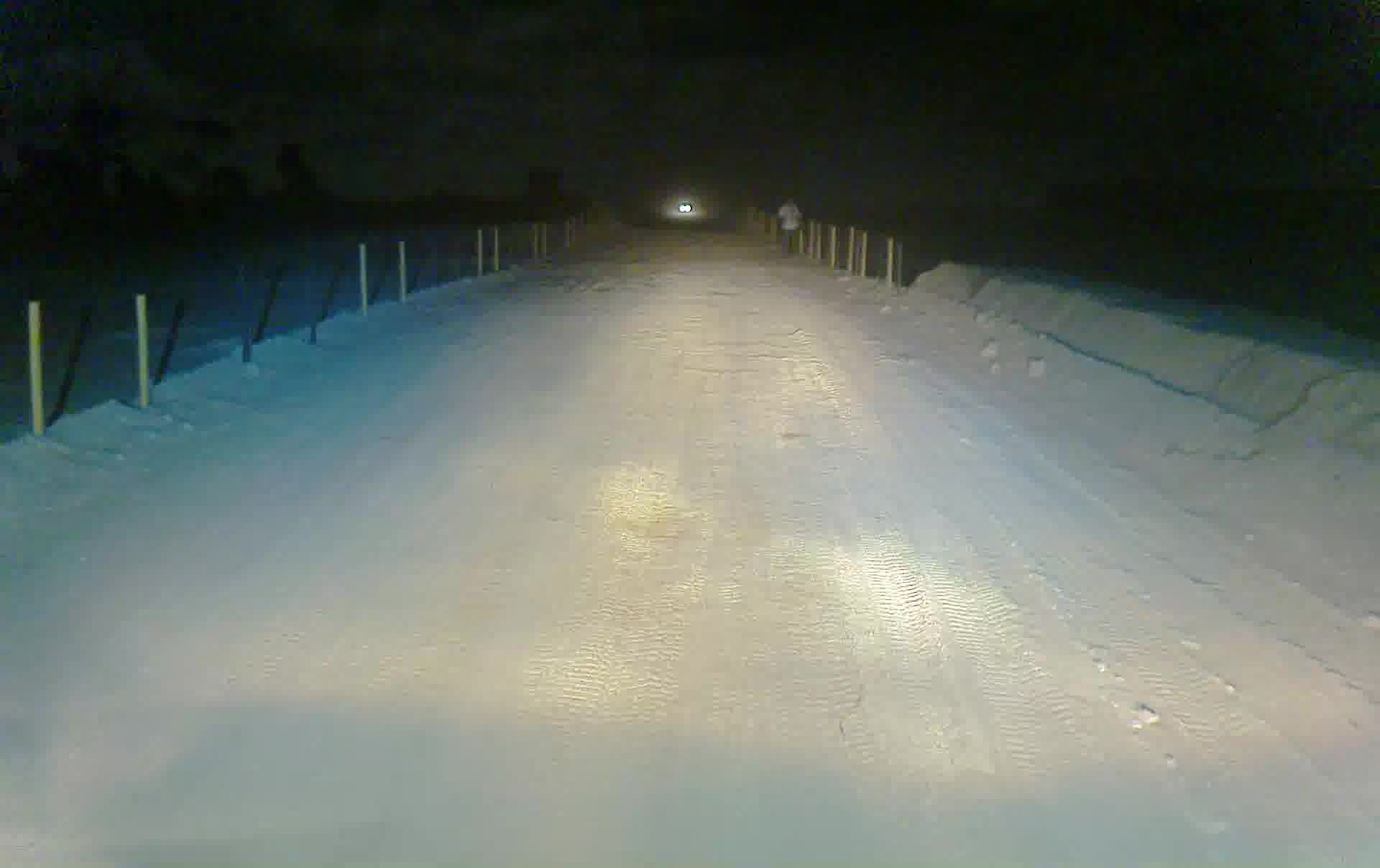}
		\caption{Test track at night.}
	\end{subfigure}%
	\hfill
	\begin{subfigure}[b]{0.32\linewidth}
		\includegraphics[width=\linewidth,trim={2cm 0cm 0cm 2cm},clip]{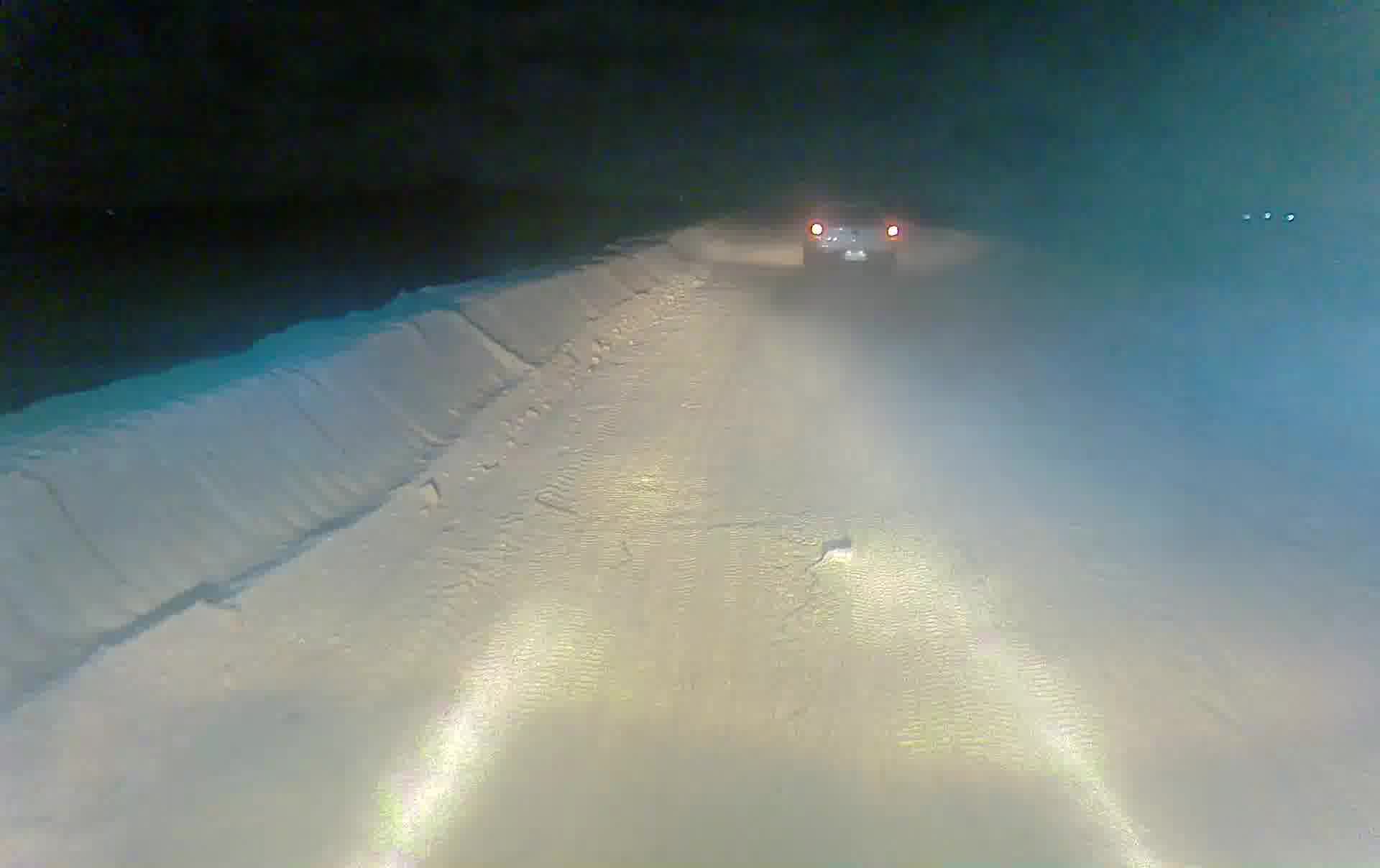}
		\caption{Track dusty/night.}
	\end{subfigure}
	\hfill
	\begin{subfigure}[b]{0.32\linewidth}
		\includegraphics[width=\linewidth,trim={2cm 0cm 0cm 2cm},clip]{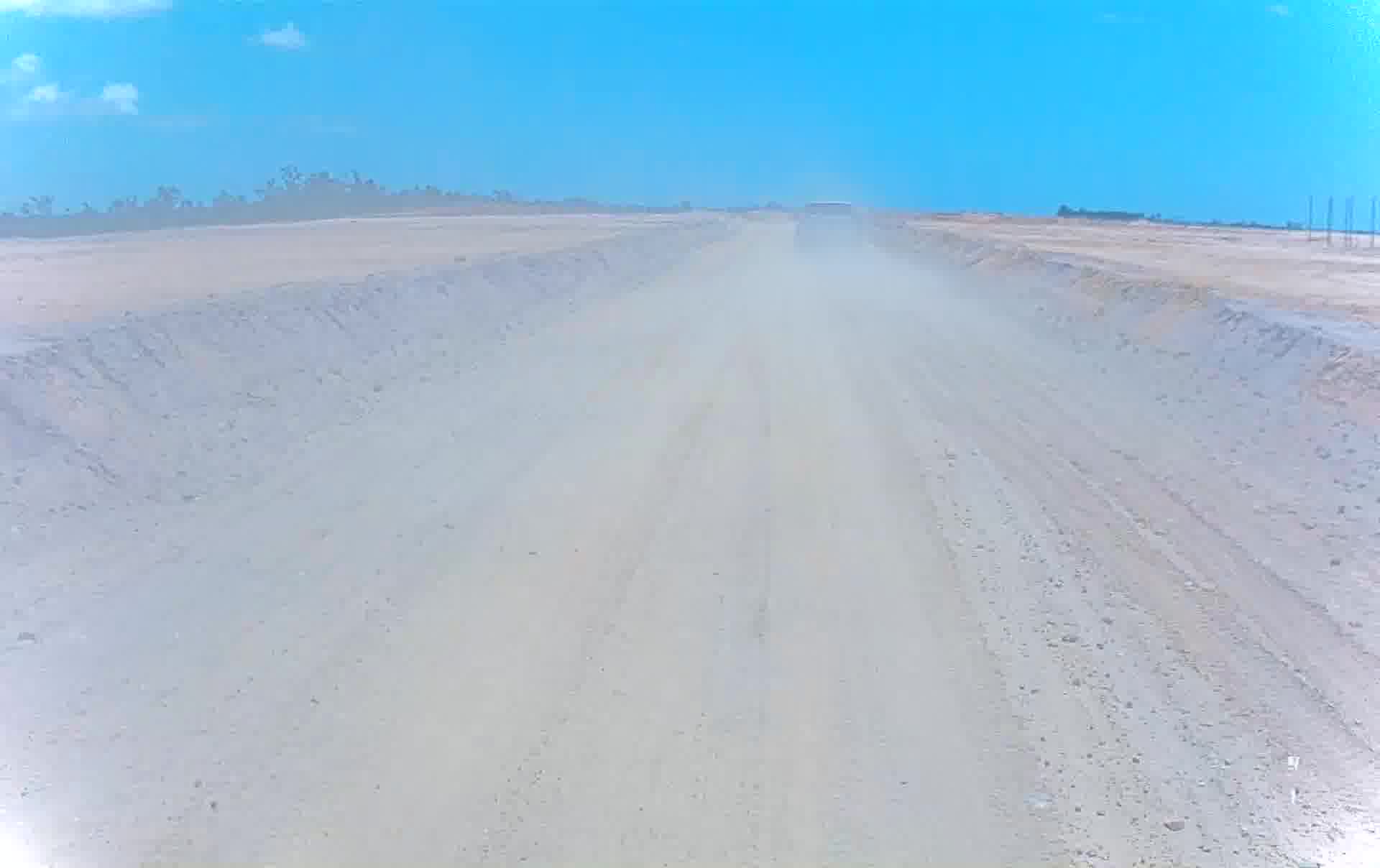}
		\caption{Track with dusty.}
	\end{subfigure}%
	\caption{Condições adversas.}
	\label{fig:situacoes-adversas}
\end{figure}

For technical reasons, it was not possible to have all the adverse conditions in all places where the collection took place. The \autoref{tab:adverse-situation} and the \autoref{fig:situacoes-adversas} show the list of the locations and adverse situations where we have collected data.

\newcommand{\spi}[1]{\textsuperscript{\textit{#1}}}

\begin{table}[t!]
	\centering
    \scriptsize
	\caption{Adverse condition.}
	\label{tab:adverse-situation}
	\begin{tabular}{l|l|l}
	\specialrule{.1em}{.05em}{.05em}
		\bfseries Type 	& \bfseries Place & \bfseries Condition \\ 
		\specialrule{.1em}{.05em}{.05em}
		Off-road  & CIMATEC test track & Daytime, night, dirty  \\
		\hline
		\multirow{3}{*}{Unpaved roads}   	& Jauá            		   	& \multirow{3}{*}{Daytime, Raining}   \\
		& Praia do Forte     	   	&    \\
		& Estrada dos Tropeiros	   	&    \\
		\specialrule{.1em}{.05em}{.05em}
	\end{tabular}
\end{table}

\begin{figure}[!t]
	\begin{subfigure}[b]{0.45\linewidth}
		\includegraphics[width=\linewidth,trim={0cm 0cm 0cm 2cm},clip]{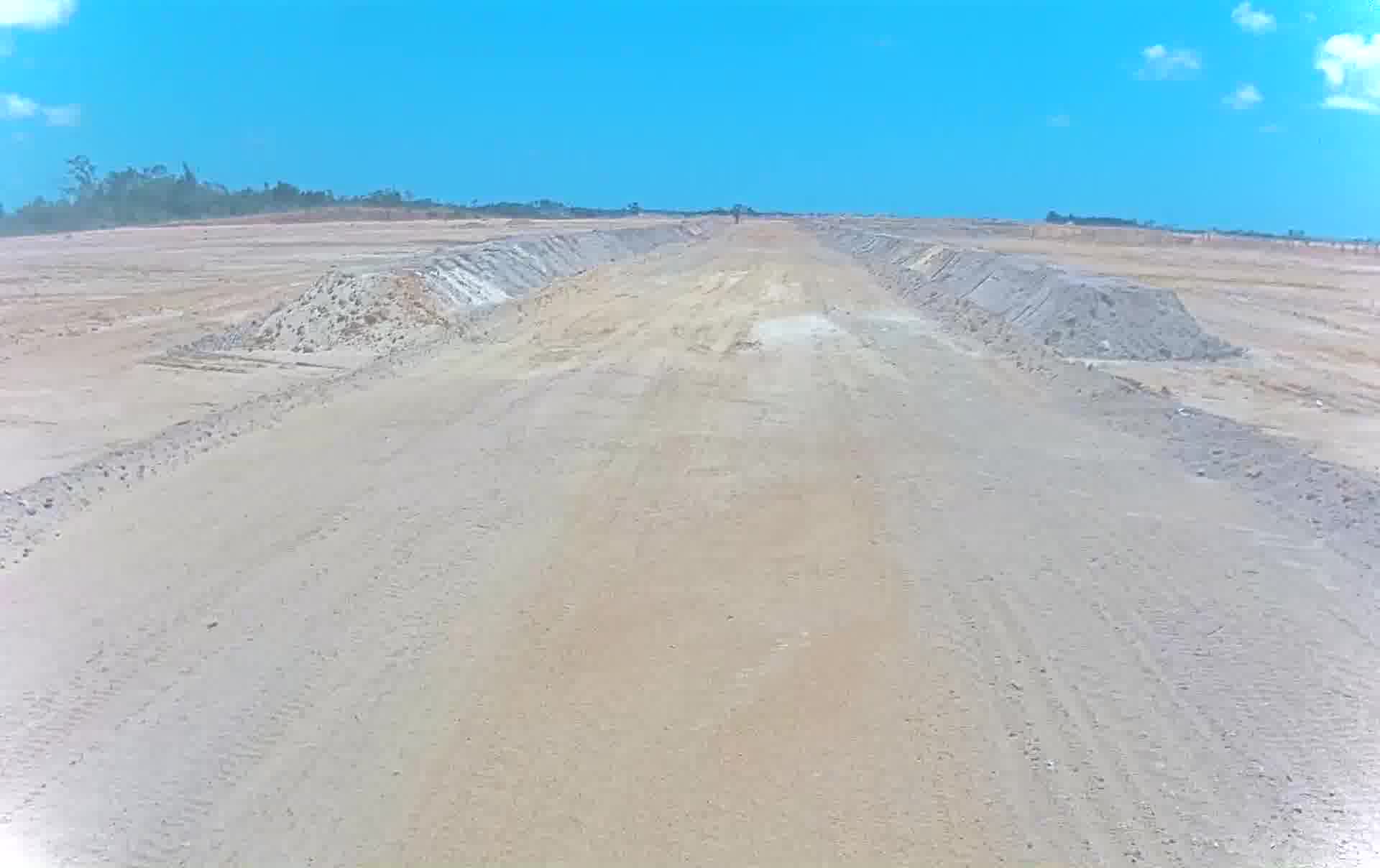}
		\caption{Slopes and open space.}
	\end{subfigure}%
	\hfill%
	\begin{subfigure}[b]{0.45\linewidth}
		\includegraphics[width=\linewidth,trim={0cm 0cm 0cm 2cm},clip]{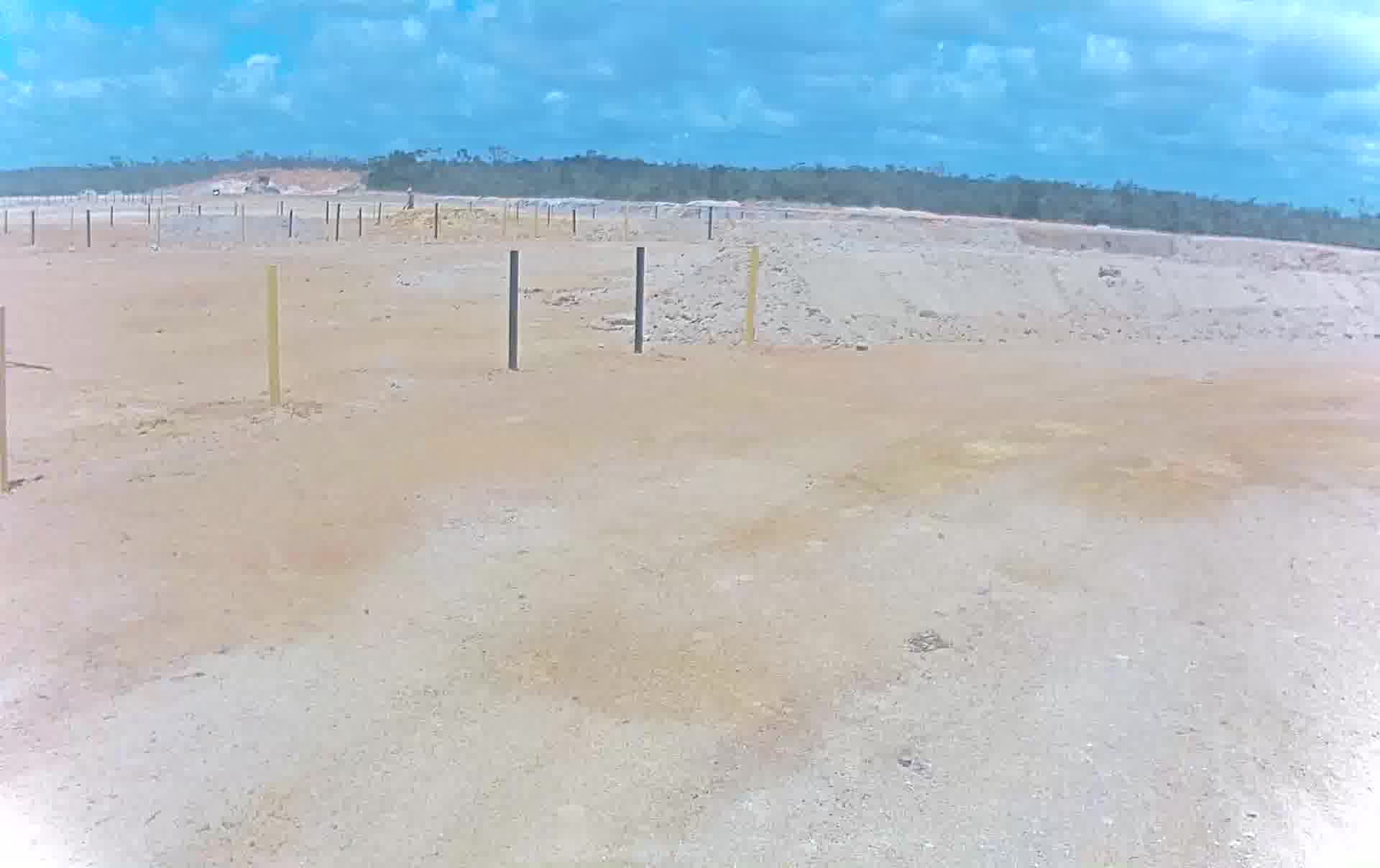}
		\caption{Pickets and slopes.}
	\end{subfigure}
	
	\begin{subfigure}[b]{0.45\linewidth}
		\includegraphics[width=\linewidth,trim={0cm 0cm 0cm 2cm},clip]{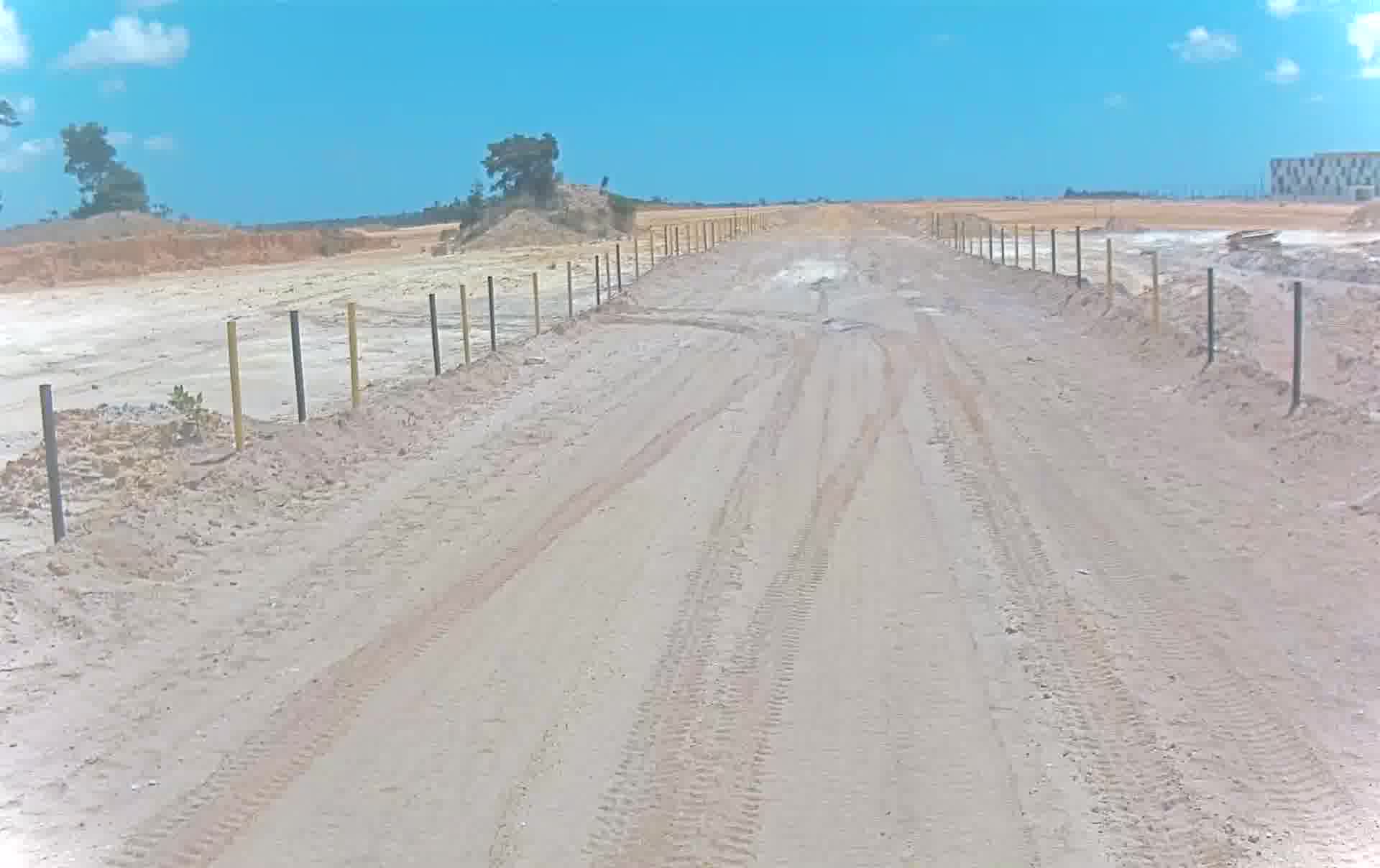}
		\caption{Pickets.}
	\end{subfigure}%
	\hfill%
	\begin{subfigure}[b]{0.45\linewidth}
		\includegraphics[width=\linewidth,trim={0cm 0cm 0cm 2cm},clip]{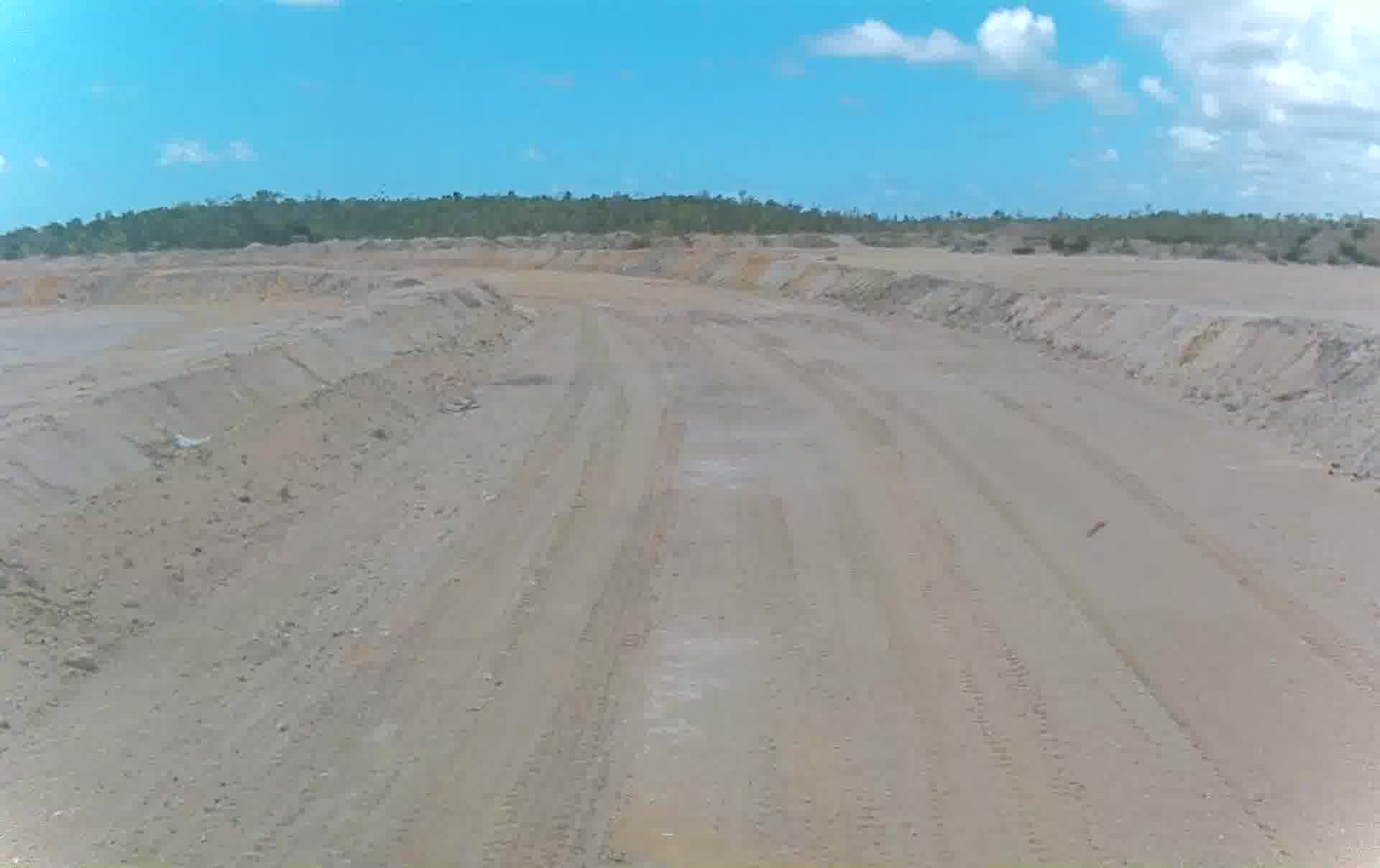}
		\caption{Slopes.}
	\end{subfigure}
	
	\caption{Different limits of the test track.}
	\label{fig:diferentes-limites-de-pista}
\end{figure}

\textit{Off-road test track}.
Taking into account the application of vehicles for transporting cargo and passengers in industrial operation, we have developed a test track simulating off-road environments such as the mining where the difference in colors and textures are slight, making it difficult to segment the track area. The \autoref{fig:diferentes-limites-de-pista} shows parts of the track and their different kinds of limiters.

\begin{figure*}[!t]
	\includegraphics[width=1\textwidth]{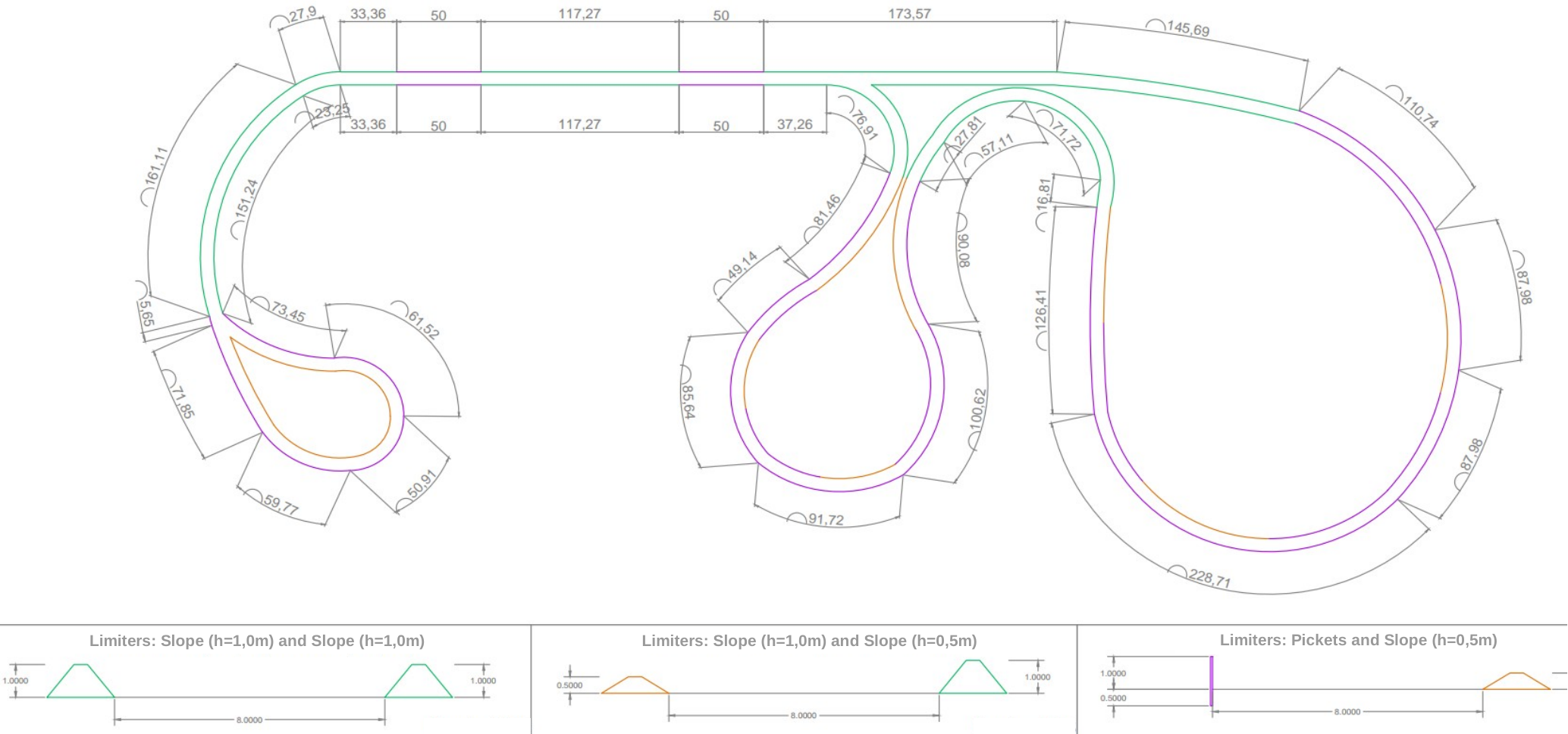}
	\caption{Map of the off-road test track.}
	\label{fig:projeto-da-pista-off-road}
\end{figure*}

The test track is approximately 3,000 meters long. It is a closed circuit with straight sectors and open and closed curves to the right and the left. 
We marked the test track limits with pickets and embankment slopes of different sizes. \autoref{fig:projeto-da-pista-off-road} shows the track design. It is possible to see the lines in green indicating slopes of 1 meter, yellow lines indicating slopes of 50 cm, and purple lines indicating pickets and empty spaces interspersed.

\textit{Data collection}.
In those scenarios, the lack of paving on the roads leads to the absence of well-defined edges delimiting correctly where the region of traffic ends or begins. Besides, the weak variation in textures and colors on the off-road track test makes the segmentation task even more difficult. To validate the effectiveness of the proposed perception in that environment and to allow training and testing of the system, we collected data in different situations and locations. We also decided to record the videos with the car zigzagging to increase the capture perspectives.

We have collected data on unpaved roads with good and bad visibility conditions in different places of the metropolitan region of Salvador, such as Jauá, Estrada dos Tropeiros, and Praia do Forte.  We recorded images in a mix of dirt roads, urban environments with houses and buildings, and rural areas with farms, narrow and unpaved roads partially delimited by a curb surrounded by some palm trees. The data were collected during the morning and the afternoon, with sun and rain. 

\begin{figure}[!t]
	\begin{subfigure}[b]{0.45\linewidth}
		\includegraphics[width=\linewidth,trim={0cm 0cm 0cm 2cm},clip]{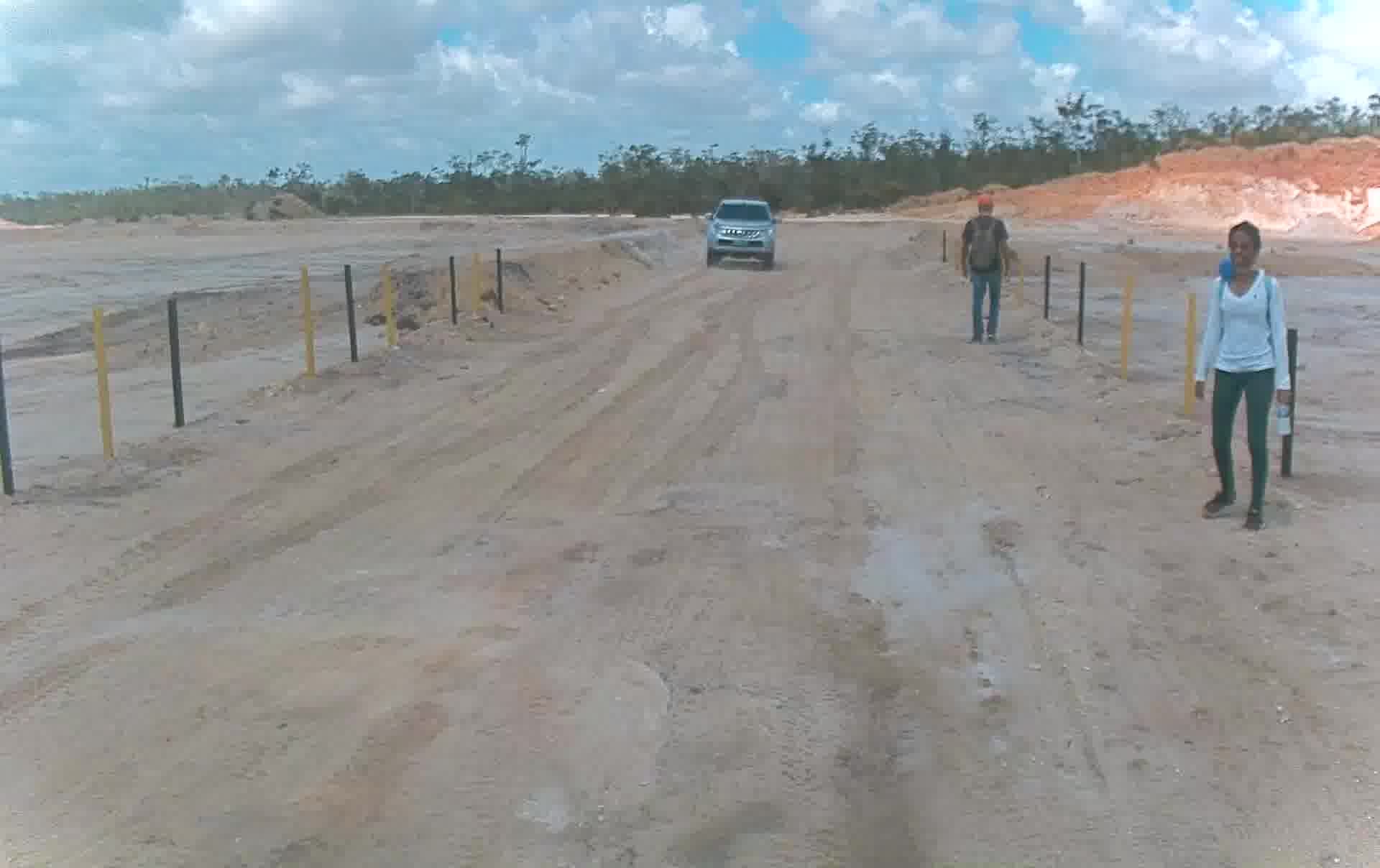}
		\caption{Original image.}
	\end{subfigure}%
	\hfill%
	\begin{subfigure}[b]{0.45\linewidth}
		\includegraphics[width=\linewidth,trim={0cm 0cm 0cm 2cm},clip]{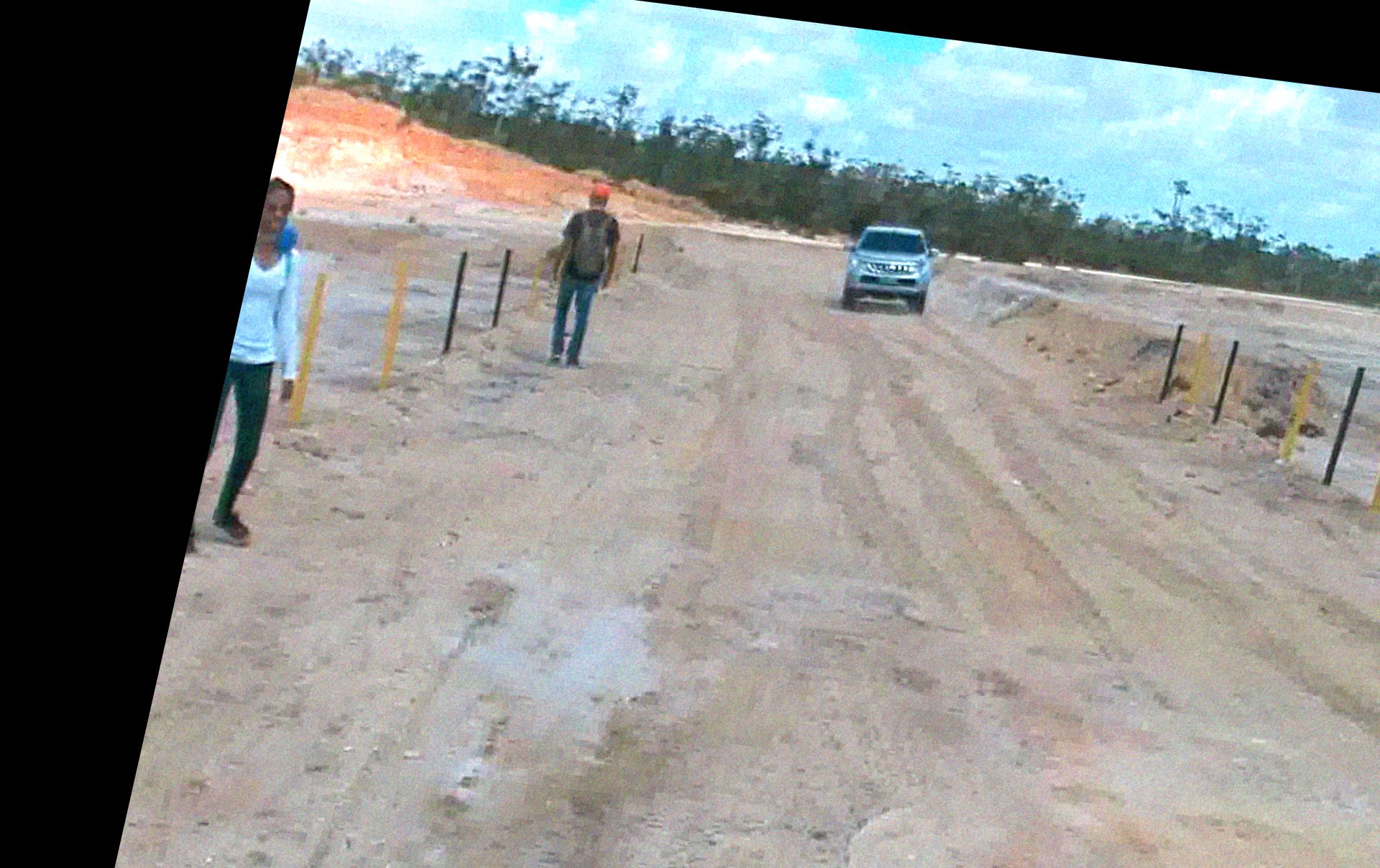}
		\caption{Noise, rotating and crop.}
	\end{subfigure}
	\begin{subfigure}[b]{0.45\linewidth}
		\includegraphics[width=\linewidth,trim={0cm 0cm 0cm 2cm},clip]{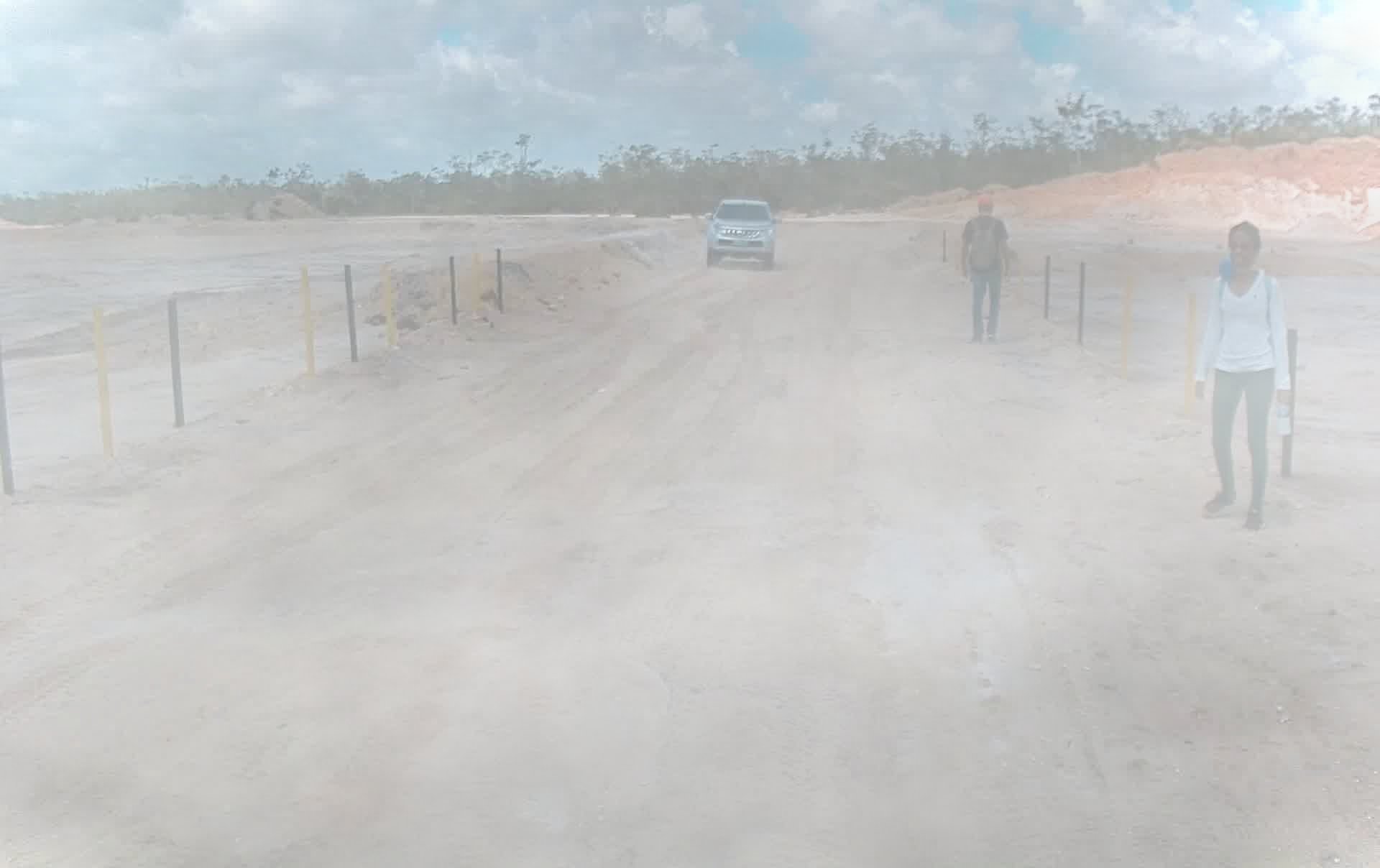}
		\caption{Fog.}
	\end{subfigure}%
	\hfill%
	\begin{subfigure}[b]{0.45\linewidth}
		\includegraphics[width=\linewidth,trim={0cm 0cm 0cm 2cm},clip]{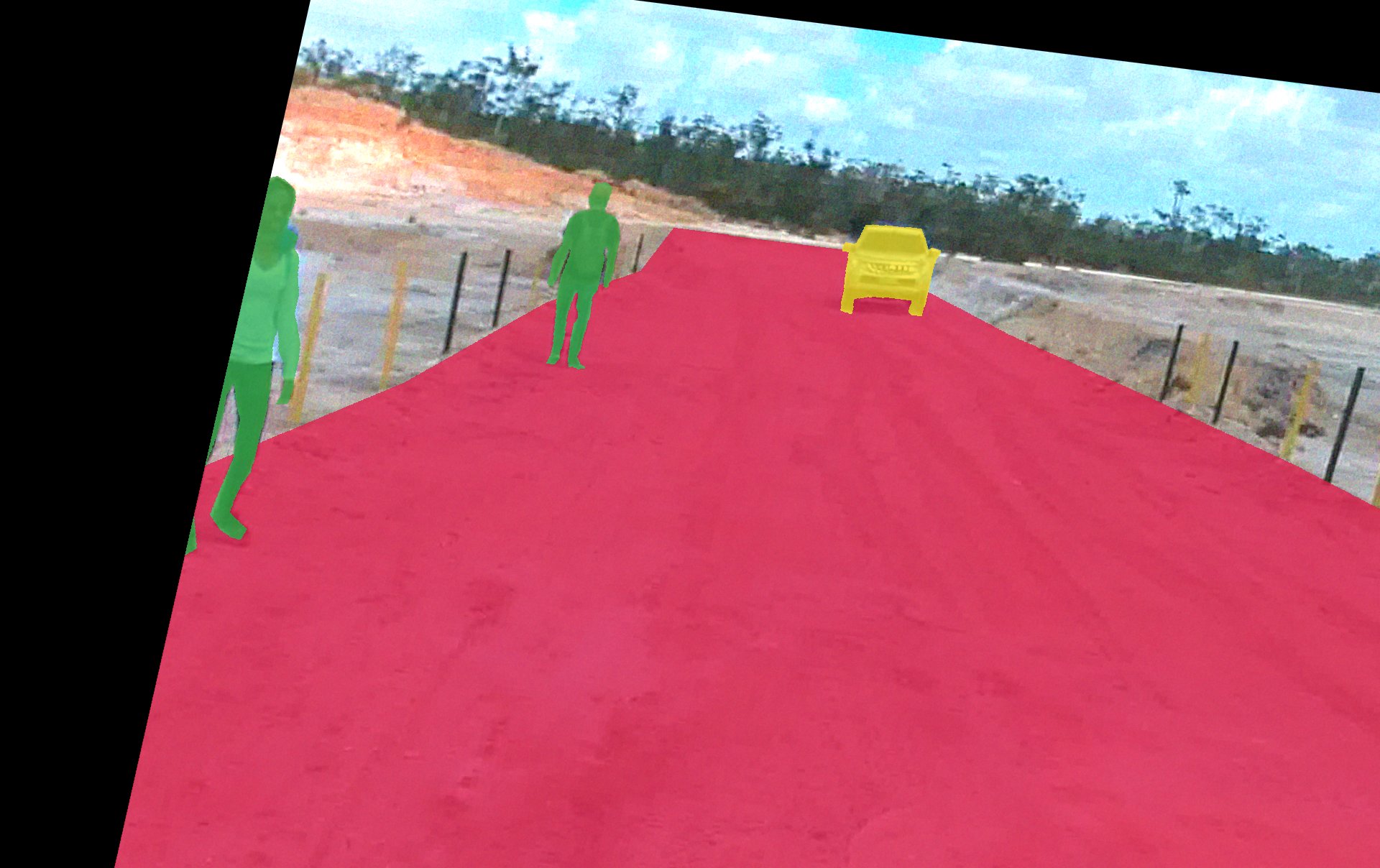}
		\caption{Image, segmentation mask.}
	\end{subfigure}
	\caption{Artificial data generation.}
	\label{fig:geracao-artificial-de-dados}
\end{figure}

In addition to the acquisitions on unpaved roads, we have also recorded some data in the controlled environment — the test track built for the research (Fig. \ref{fig:projeto-da-pista-off-road}). The data acquisition was carried out around noon, in the evening, and at night. We recorded images in adverse conditions such as low light and dust to increase the diversity of the dataset. Besides recording images at night, with dust and rain, we also create a script to allows synthetically increasing the dataset diversity by rendering fog, snow, and other impairments (Fig. \ref{fig:geracao-artificial-de-dados}). Such scripts were developed with the help of the Imgaug library. \citep{imgaug}.

\begin{figure}[!t]
	\begin{subfigure}[b]{0.45\linewidth}
		\includegraphics[width=\linewidth,trim={2cm 0cm 0cm 2cm},clip]{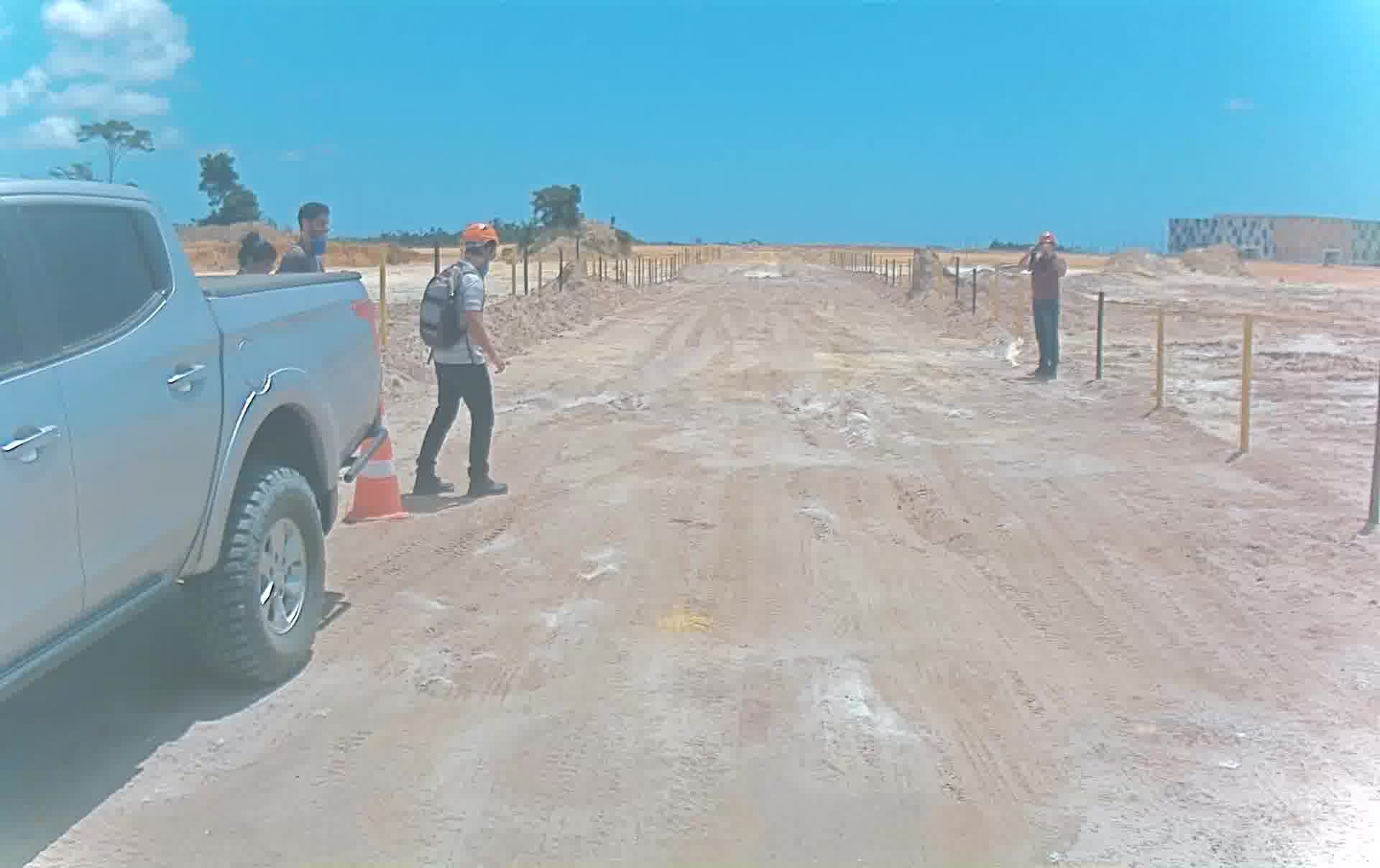}
		\caption{Original image.}
	\end{subfigure}%
	\hfill%
	\begin{subfigure}[b]{0.45\linewidth}
		\includegraphics[width=\linewidth,trim={2cm 0cm 0cm 2cm},clip]{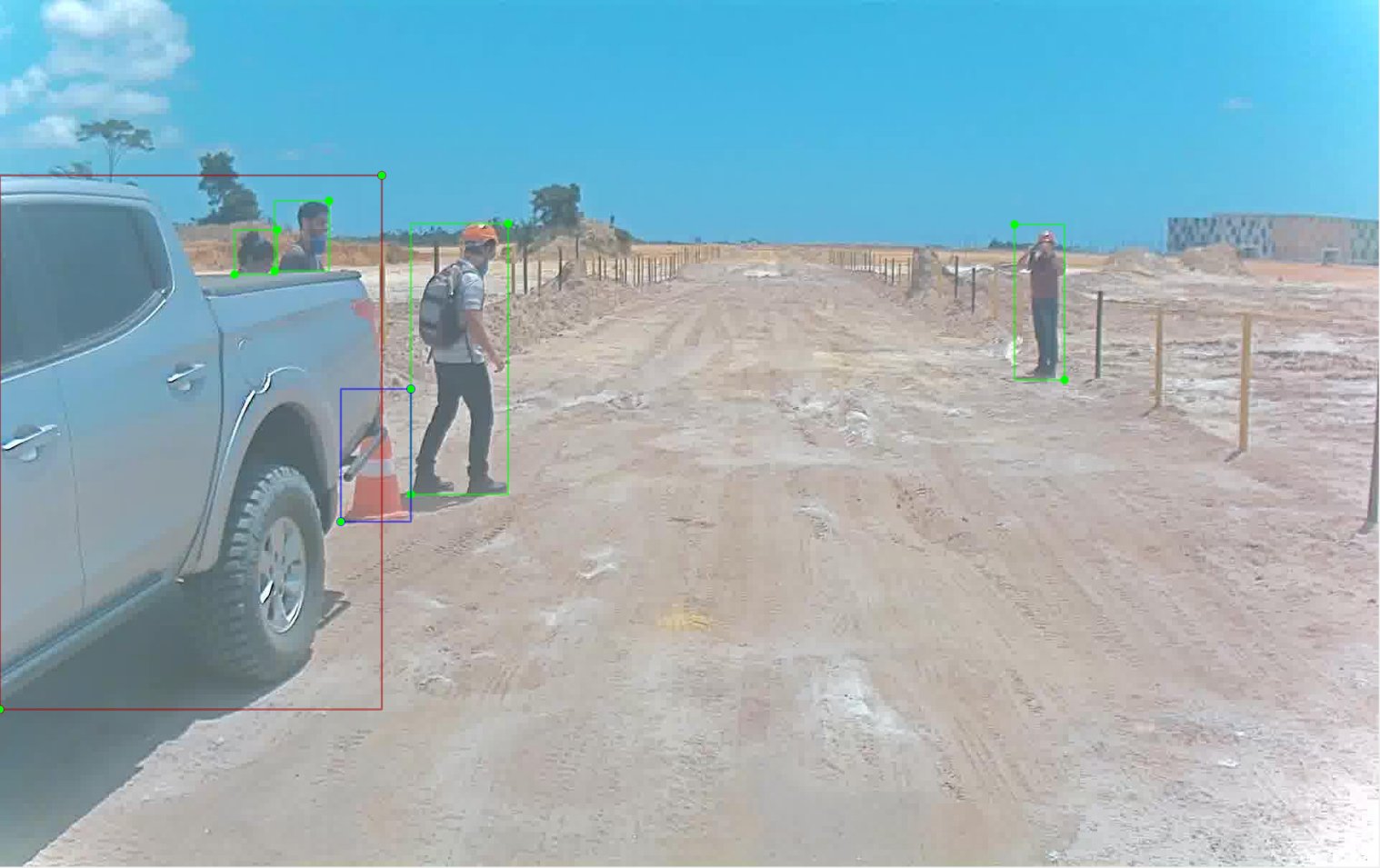}
		\caption{Detection and bounding box.}
	\end{subfigure}
	\hfill%
	\begin{subfigure}[b]{0.45\linewidth}
		\includegraphics[width=\linewidth,trim={2cm 0cm 0cm 2cm},clip]{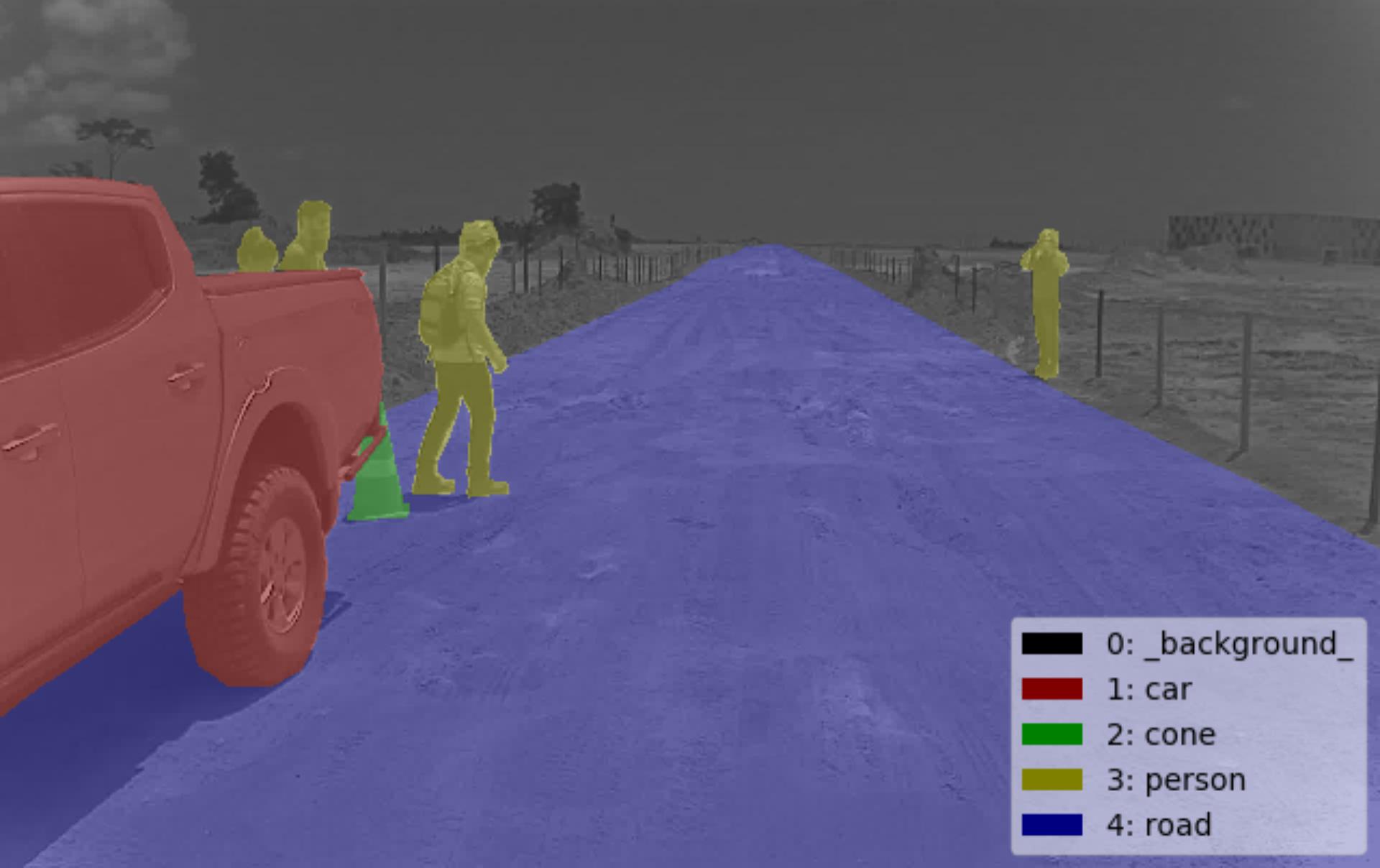}
		\caption{Semantic segmentation.}
	\end{subfigure}%
	\hfill%
	\begin{subfigure}[b]{0.45\linewidth}
		\includegraphics[width=\linewidth,trim={2cm 0cm 0cm 2cm},clip]{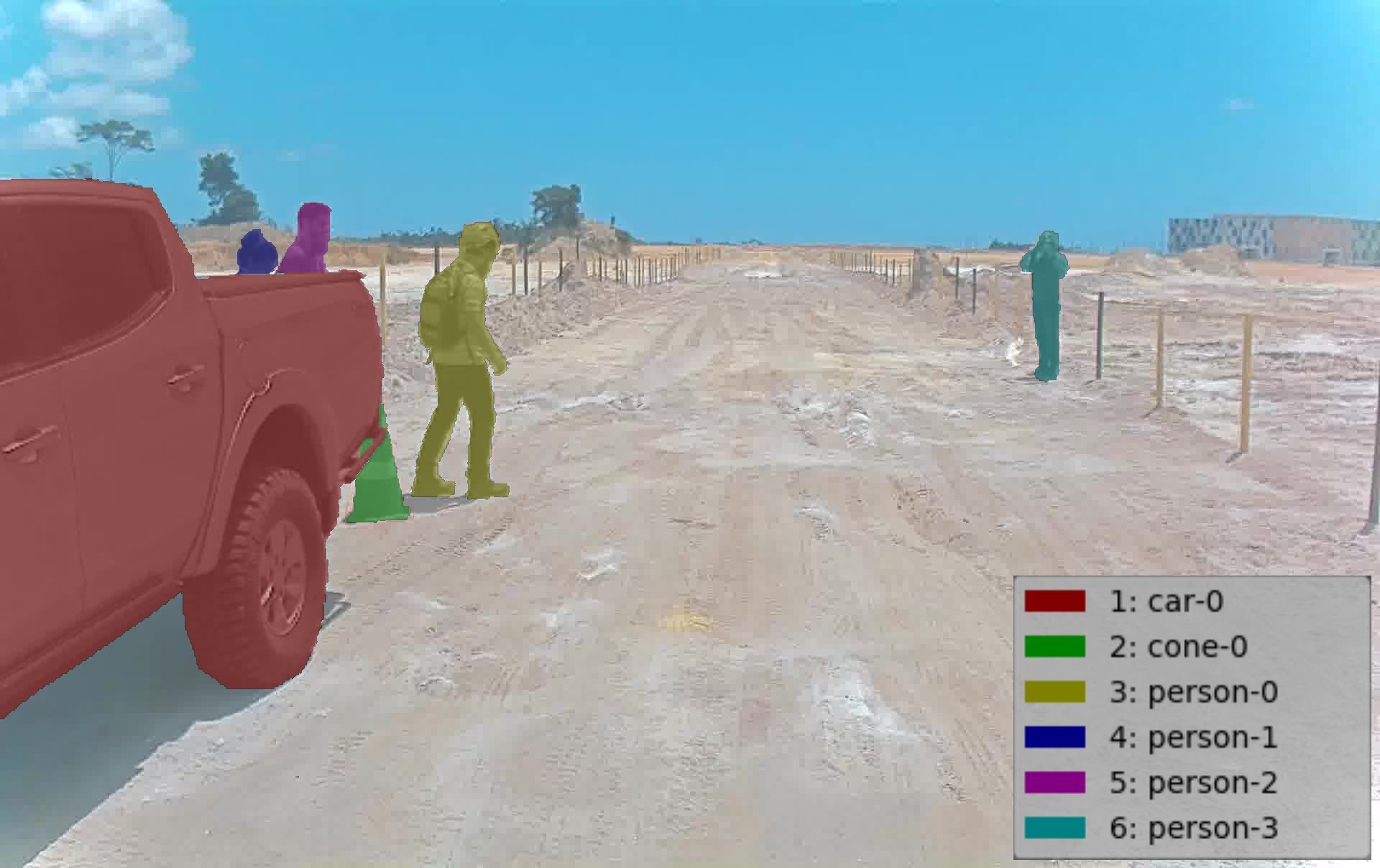}
		\caption{Instance segmentation.}
	\end{subfigure}
	\begin{subfigure}[b]{0.45\linewidth}
		\includegraphics[width=\linewidth,trim={2cm 0cm 0cm 2cm},clip]{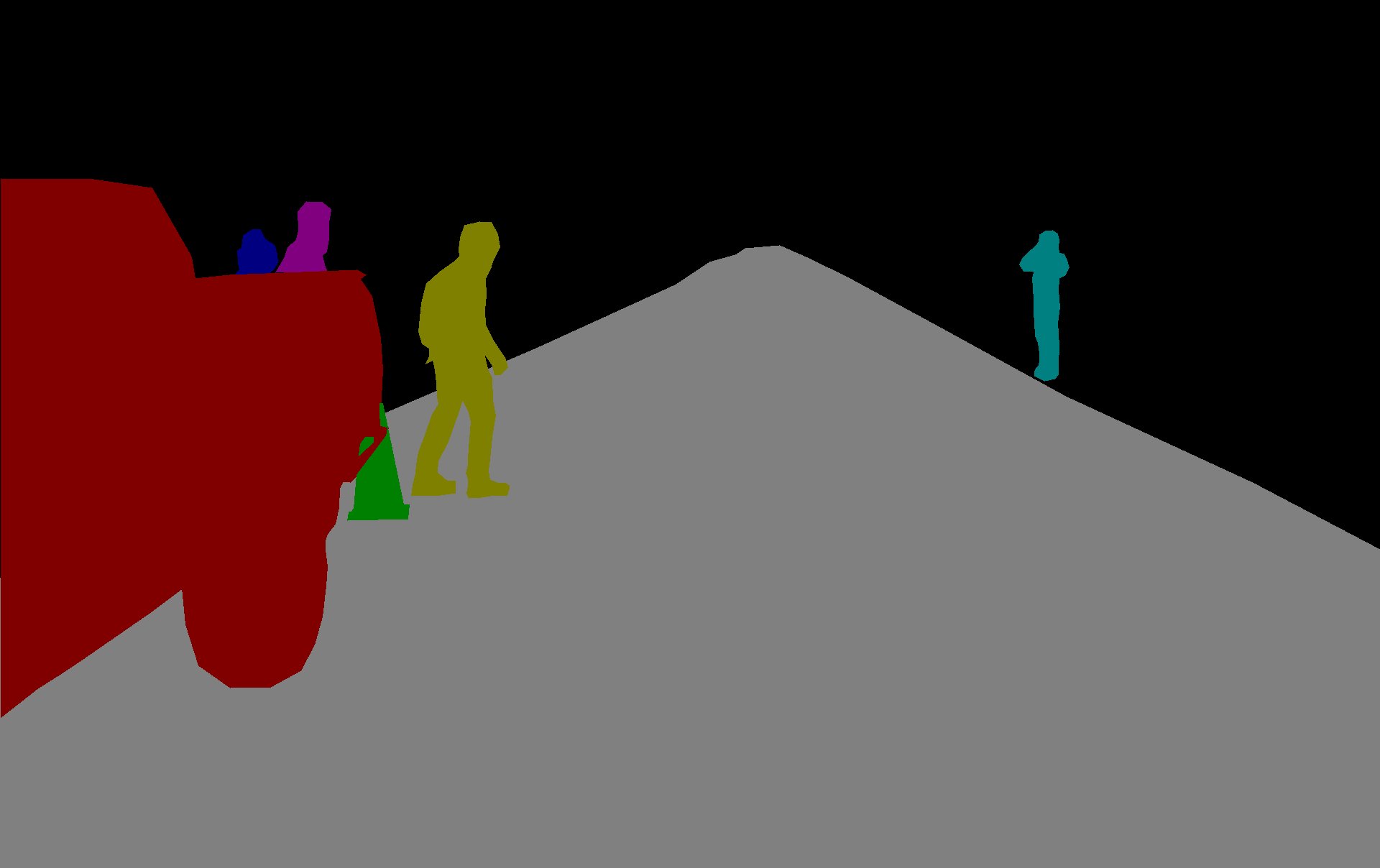}
		\caption{Panoptic Ground-truth.}
	\end{subfigure}%
	\hfill%
	\begin{subfigure}[b]{0.45\linewidth}
		\includegraphics[width=\linewidth,trim={2cm 0cm 0cm 2cm},clip]{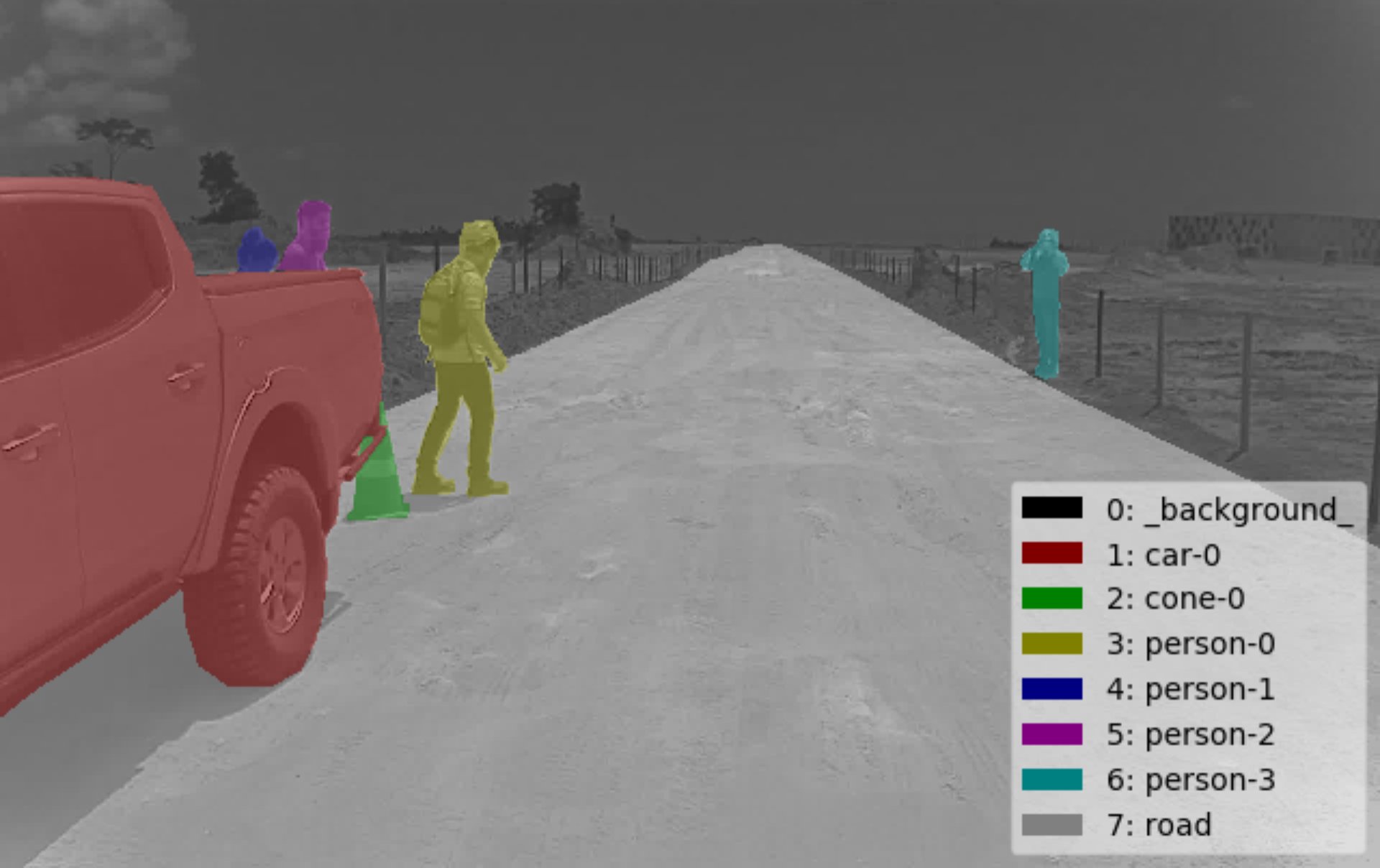}
		\caption{Panoptic Segmentation.}
	\end{subfigure}
	\caption{Types of annotations.}
	\label{fig:diferentes-tipos-de-anotacoes}
\end{figure}

\textit{Annotation}.
We labeled the data suitable for panoptic segmentation \footnote{ Panoptic is what allows showing or seeing the whole at one view}. Panoptic segmentation treats countable things like people and cars simultaneously with non-countable stuff such as road and vegetation. This task unifies the semantic and instance segmentation (Fig. \ref{fig:diferentes-tipos-de-anotacoes}) \citep{Alexander:2018:CoRR:PanopticSegmentation}. We adopt this strategy because it allows generating ground-truth to the instance and semantic segmentation as well as to object detection. Even though the focus of this work is the semantic segmentation of unpaved roads, this choice seemed to be prudent because it allows future research using the same dataset.

We have used the LabelMe \citep{Russell:2008:LabelMe} annotation style applying polygons to outline the object. The results of each image annotation — groups of polygons and the respective classes associated with it — were written in a .json file. An identifier was attached to the label to ensure the correct annotation of different instances of the same class (e.g., \textit{person-0, person-1, ..., car-0, car-1, ..., car-n}).  On the other hand, in the labeling of non-countable stuff, such as the road, we have used only the label (e.g., \textit{road}).

\begin{figure}[!t]
	\begin{subfigure}[b]{0.32\linewidth}
		\includegraphics[width=\linewidth,trim={2cm 0cm 0cm 2cm},clip]{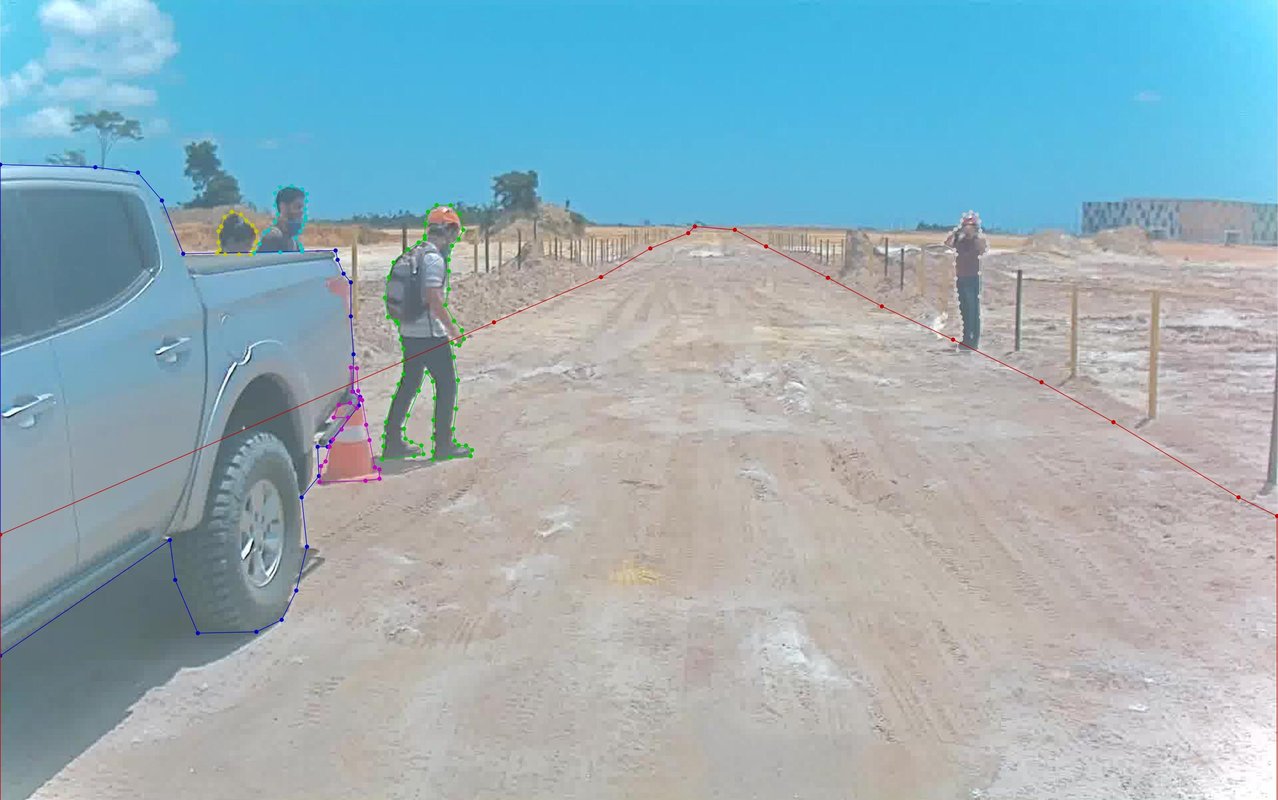}
	\end{subfigure}%
	\hfill%
	\begin{subfigure}[b]{0.32\linewidth}
		\includegraphics[width=\linewidth,trim={2cm 0cm 0cm 2cm},clip]{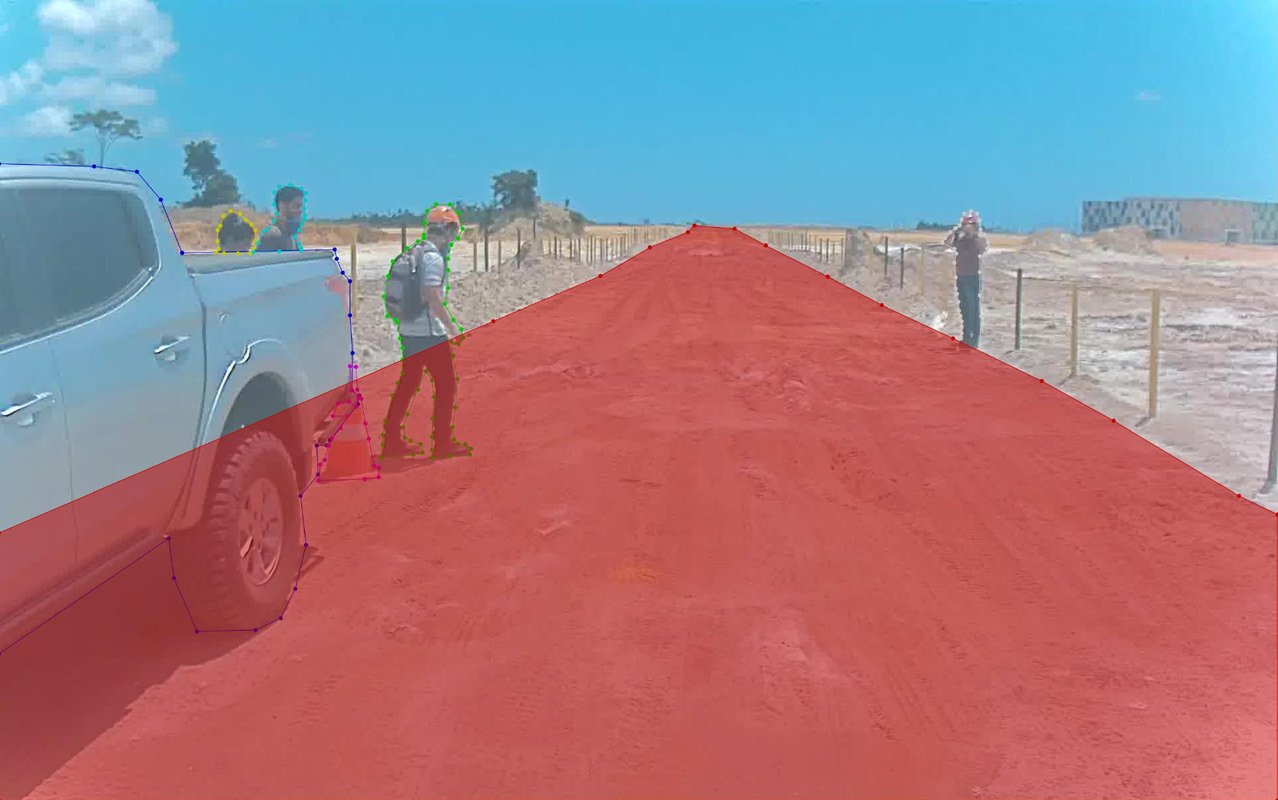}
	\end{subfigure}
	\hfill%
	\begin{subfigure}[b]{0.32\linewidth}
		\includegraphics[width=\linewidth,trim={2cm 0cm 0cm 2cm},clip]{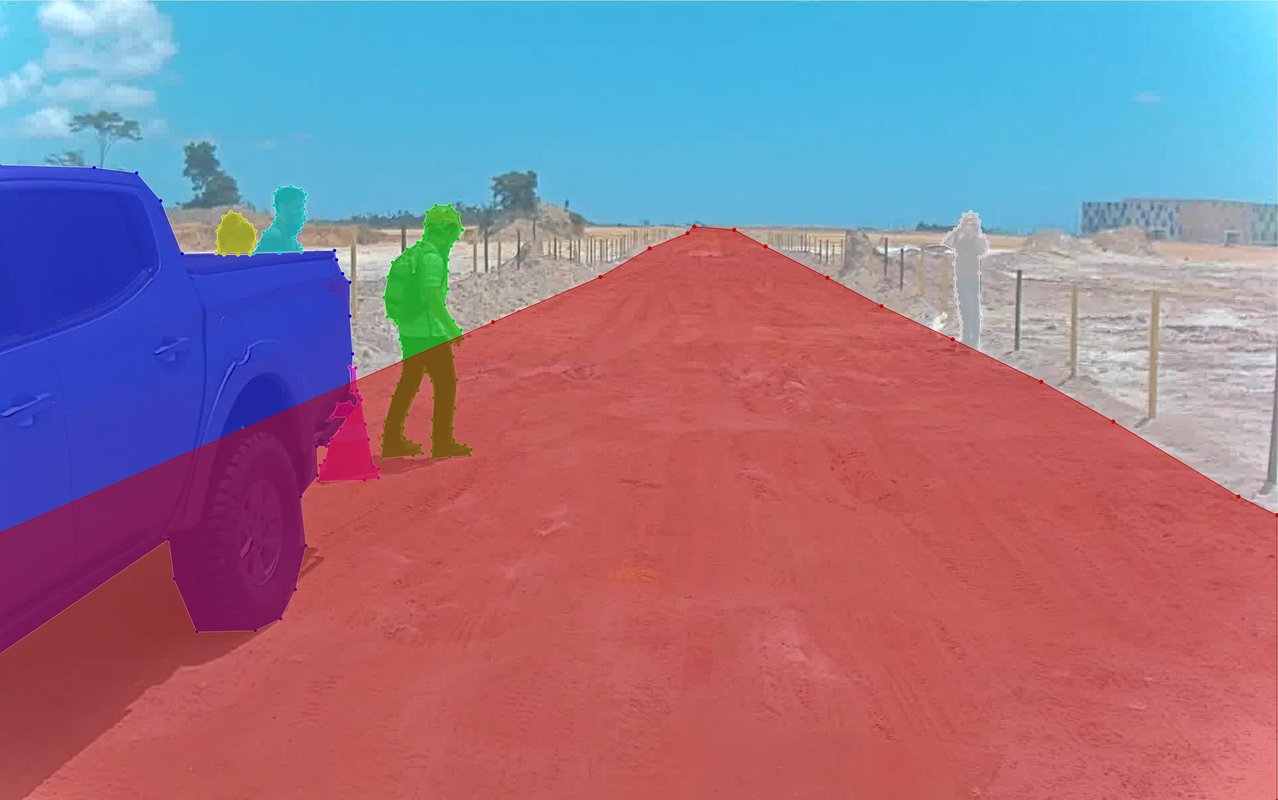}
	\end{subfigure}%
	\caption{Image annotation process.}
	\label{fig:processo-de-anotação-das-imagens}
\end{figure}

In the annotation process, we annotated the road first and after all the elements over it, so that the result was an annotation of layers over layers (Fig. \ref{fig:processo-de-anotação-das-imagens}). We have used this strategy to speed up the creation of the dataset. To avoid overlapping the road over other class like person or car, the script developed to convert the .json files into .png masks uses a pre-established order to render the information.  

In this research, we are considering only the segmentation of traffic areas and obstacles in real-time as being relevant to the perception subsystem. For this task, the segmentation of sky, buildings, and other elements not directly involved in the decision-making process to drive a vehicle are not required. Besides, we found a limited number of relevant classes in those less dense off-road environments. So, we have opted for a reduced number of annotated classes to decrease the effort and speed up the development of the research.

The strategy of focusing on a few groups has proved to be adequate to validate the concept. In total, eight classes were recorded, grouped into six distinct categories, prioritizing the traffic area (road) and obstacles encountered during several hours of data acquisition. The \autoref{tab:lista-declasses-e-categorias} shows the classes grouped in categories where only the road and background classes do not have multiple instances. In the ground category, there is only the class road, the human group has only the person class, and the animal group has the animal class. On the other hand, the vehicle group has the classes car, motorcycle, truck, and bike. These elements are relevant to the research as it involves imminent obstacles and risks to driving. There is also the cone class in the infrastructure group and the background class, including the elements considered as not relevant by this work.

\begin{table}[!t]
	\centering
	\scriptsize
	\caption{List of classes and categories, average pixels occupied in all images, and the total number of occurrences.}
	\label{tab:lista-declasses-e-categorias}
	\begin{tabular}{l+l+c+c} 
	   \specialrule{.1em}{.05em}{.05em}
		\bfseries Group &
		\bfseries Class &
		\bfseries Pixels Avg. &
		\bfseries Total instances \\ 
		\specialrule{.1em}{.05em}{.05em}
		Ground  & \textit{road}  & 47.20\% & 11,508   \\
		Human  & \textit{person}  & 0.08\% & 1,896  \\
		Vehicle  & \textit{car}   & 0.29\% & 4,186   \\
		& \textit{moto}   & 0.006\% & 114   \\
		& \textit{truck}  & 0.03\% & 154  \\
		& \textit{bus}    & 0.03\% & 101   \\
		& \textit{bike}   & 0.001\% & 41   \\
		Animal & \textit{animal} & 0.001\% & 27  \\
		Infrastructure   & cone  & 0.002\% & 129   \\ 
		Void   & \textit{background} & 52.34\% & 11,512   \\
		\specialrule{.1em}{.05em}{.05em}
	\end{tabular}
\end{table}

\textit{Data description}.
In total, this dataset has 11,479 annotated images. We have recorded some data on the off-road test track created to support the research and others on unpaved roads in the metropolitan region of Salvador-BA, Brazil (Tab. \ref{tab:adverse-situation}). The off-road data was captured during day and night, having or not dust. On the other hand, the unpaved data were recorded during the day with clean and rainy weather. On the unpaved roads of Jauá, Praia do Forte, and Estrada dos Tropeiros, we collected 823 images in rainy conditions and 5,135 in the daytime. We also recorded and annotated 1,556 on the off-road test track during the day, 1,546 in the late afternoon, and 1,953 at night (Fig. \ref{tab:dataset}). Besides that, there is also possible to generate additional synthetic data through a script to increase the dataset.  When activated on the training, this script produces dynamic images applying filters and random cuts from real images annotated.

\begin{table}[!t]
	\centering
	\scriptsize
	\caption{Annotated images.}
	\label{tab:dataset}
	\begin{tabular}{l+l+c+c+c+c}
		\specialrule{.1em}{.05em}{.05em}
		\bfseries Type    &
		\bfseries Places          &
		\bfseries Day   &
		\bfseries Evening&
		\bfseries Night&
		\bfseries Rain \\ 
		\specialrule{.1em}{.05em}{.05em}
		Paved   & E. dos Tropeiros& 202   & --     & --     & 209 \\ 
		        & Linha Verde     &       &        &      &  \\
		        & Jauá            &       &        &      &  \\
		Unpaved & E. dos Tropeiros& 5,135 & --     & --     &  823 \\
				& Jauá            &       &        &      &   \\
				& Praia do Forte  &       &        &    &   \\
		Off-road& Test Track      & 1,556 & 1,546 & 1,953  &  55 \\
		\specialrule{.1em}{.05em}{.05em}
		Total	&                 & 6,893 & 1,546 &  1,953  & 1,087  \\
		\specialrule{.1em}{.05em}{.05em}
	\end{tabular}
\end{table}

The authors of this work have decided to label only the classes considered relevant for the validation of the perception subsystem for an ADAS or an autonomous vehicle in an off-road and unpaved environment. Unlike other datasets such as Cityscapes~\citep{cordts:2016:cityscapes} and KITTI~\citep{kitti-menze:2015:cvpr} that annotate wall, buildings, sky, tree, sidewalk, we have only annotated the traffic and non-traffic area, in addition to dynamic obstacles such as cars, people, and animals. This approach facilitates the annotation task in addition to keeping the algorithm focused on segmenting what is relevant for the vehicle on the road. \autoref{tab:lista-declasses-e-categorias} shows the classes annotated in the scene.

As can be seen in \autoref{tab:lista-declasses-e-categorias}, our dataset has an imbalance. That happens due to the lack of some classes in unpaved environments that is distant from downtown, with several cars and pedestrians. Also, there is a perceived rarity regarding animals crossing the track.

Like the unpaved roads, the test track used to develop and validate the system also has a limited number of people and cars. Nevertheless, several datasets such as COCO~\citep{Lin:2014:ECCV:COCO}, Pascal VOC, and Cityscapes~\citep{cordts:2016:cityscapes} have already covered the segmentation of people and animals. However, the segmentation of unpaved roads and traffic area in an off-road environment such as the test track, where the traffic area has the same color and texture as the non-traffic area, is a contribution of our research and dataset.

Until the moment that paper was written, we hadn't found datasets for unpaved roads, rural areas, and off-road like this. Our dataset and the Mapillary \citep{Neuhold:2017:Mapillary} are the only ones that cover paved, non-paved off-road, and adverse condition altogether, as shown in \autoref{tab:comparison_datasets}. However, the Mapillary has a few samples of off-road and unpaved images as compared with ours. Besides, our dataset is the one with the highest number of pixels labeled. We have an annotated pixel density of 47.66\%, even when we don't take into account the background label.

\begin{table*}[!ht]
	\centering
	\scriptsize
\caption{Comparison between ours Kamino datasets and other ones.}
\label{tab:comparison_datasets}
\begin{tabular}{l+c+c+c+c+c+c+c+c}
\specialrule{.1em}{.05em}{.05em}
\multicolumn{1}{c}{\bfseries Dataset} &
  \multicolumn{1}{c}{\bfseries \begin{tabular}[c]{@{}c@{}}\# \\ images\end{tabular}} &
  \multicolumn{1}{c}{\bfseries \begin{tabular}[c]{@{}c@{}}\# \\ classes\end{tabular}} &
  \multicolumn{1}{c}{\bfseries Paved} &
  \multicolumn{1}{c}{\bfseries \begin{tabular}[c]{@{}c@{}}Non-\\ paved\end{tabular}} &
  \multicolumn{1}{c}{\bfseries \begin{tabular}[c]{@{}c@{}}Off-\\ road\end{tabular}} &
  \multicolumn{1}{c}{\bfseries \begin{tabular}[c]{@{}c@{}}Adv. \\ cond.\end{tabular}} &
  \multicolumn{1}{c}{\bfseries Semantic} &
  \multicolumn{1}{c}{\bfseries Instance}
   \\
\specialrule{.1em}{.05em}{.05em}

A2D2             & 41k  & 38  & \checkmark & \ding{55}                        & \ding{55}                        & \checkmark & \checkmark & \checkmark \\
Mapillary & 25k  & 152 & \checkmark & {\color[HTML]{FE0000} \checkmark} & {\color[HTML]{FE0000} \checkmark} & \checkmark & \checkmark & \checkmark\\
Cityscapes       & 5k   & 30  & \checkmark & \ding{55}                        & \ding{55}                        & \ding{55} & \checkmark & \checkmark \\
KITTI            & 5k   & 30  & \checkmark & \ding{55}                        & \ding{55}                        & \ding{55} & \checkmark & \checkmark  \\
CamVid           & 700  & 32  & \checkmark & \ding{55}                        & \ding{55}                        & \ding{55} & \checkmark & \ding{55} \\

DeepScene        & 372  & 6   & \ding{55} & \ding{55}                        & \checkmark                        & \ding{55} & \checkmark & \ding{55} \\
YCOR             & 1k   & 8   & \ding{55} & \ding{55}                        & \checkmark                        & \checkmark & \checkmark & \ding{55}  \\
\bfseries{Kamino}           & 11.5k & 10  & \checkmark & \checkmark                        & \checkmark                        & \checkmark & \checkmark & \checkmark \\

\specialrule{.1em}{.05em}{.05em}
\end{tabular}
\end{table*}

In \autoref{tab:valores-absolutos-e-media-de-pessoas-veiculos-e-animais-por-imagem}, we have a comparison regarding the number of vehicles, animals, and people between our dataset and Cityscapes~\cite{cordts:2016:cityscapes} or KITTI~\cite{kitti-menze:2015:cvpr}. The total number of occurrences of dynamic entities in the scenes of our dataset is smaller, as expected, due to differences among the data acquisition environments.

\begin{table}[!t]
	\centering
	\scriptsize
	\caption{Absolute and average values of instances per image.}
	\label{tab:valores-absolutos-e-media-de-pessoas-veiculos-e-animais-por-imagem}
	\begin{tabular}{l+c+c+c+c+c+c}
	\specialrule{.1em}{.05em}{.05em}
		\bfseries Dataset & 
		\bfseries Person &
		\bfseries Vehicle &
		\bfseries Animal &
		\bfseries P\%  &
		\bfseries V\%  &
		\bfseries A\%\\ 
		\specialrule{.1em}{.05em}{.05em}
		kamino               & 1.9k      & 4.56k    & 27     & 0.08      & 0.37    & 0.001 \\ 
		DeepScene        & 0k      & 0k    & -- & 0.0  &  0.0  & -- \\
		YCOR        & --      & -- & --  & --  &  --  & --  \\
		Cityscapes   & 24.4k     & 41.0k    & 0      & 7.0     & 11.8      & 0.0     \\ 
		KITTI        & 6.1k      & 30.3k    & 0      & 0.8     &  4.1      & 0.0     \\ 
		CamVid        & --      & --    & 0  & --  &  --   & 0.0  \\
		\specialrule{.1em}{.05em}{.05em}
	\end{tabular}
\end{table}

In addition to the annotated data, this dataset has several videos and LiDARs point cloud collected during the development. Altogether there were four LiDARs, two Velodyne VLP-16, and two Quanergy M8. We also recorded data from 4 SEKONIX cameras with 120º FOV and 3 SEKONIX cameras with 60º FOV.

\section{Experimental setup and evaluation}
\label{sec:experimental}
\subsection{Methodology}

\textit{Datset}.
The dataset developed in this work has 11,479 labeled images. However, in the experiments carried out in this research, we used just a data subset to speed up the training process for the CMSNet arrangements. We named such a subset of Kamino-Small. He has a total of 5,523 images, in several situations, as shown in the \autoref{tab:subset_composition}. Altogether, it has 4,026 samples for training, 449 for validation, and 1,048 for testing. These data are distributed between daytime, raining, night, and evening. Furthermore, in the off-road test track, some images have dust also.

In addition to a reduced set of images, we also chose not to include some classes in training and testing. We have not considered groups such as bus, motorcycle, animal, and bike. These classes are rare in the proposed dataset and, in some cases, are not sufficient for the test stage. That approach, of merging or ignoring some groups in tests, is also used in other datasets such as Cityscapes \citep{cordts:2015:cvprw-cityscapes} and \cite{Semantic-Forested:Valada:2016}.

\begin{table}[!ht]
	\centering
	\scriptsize
	\caption{Distribution of data in training, validation and testing sets.}
	\label{tab:subset_composition}
	\begin{tabular}{l +c +c +c +c}
		\specialrule{.1em}{.05em}{.05em}
		\bfseries Condition &
		\bfseries Training &
		\bfseries Validation &
		\bfseries Testing &
		\bfseries All \\
		\specialrule{.1em}{.05em}{.05em}
		Daytime        & 1,471 ($73.5\%$) & 164 ($8.2\%$) &   367 ($18.3\%$) & 2,002 \\
		Daytime\spi{1} &   666 ($73.1\%$) &  74 ($8.1\%$) &   171 ($18.8\%$) &   911 \\
		Raining        &   539 ($72.2\%$) &  60 ($8.0\%$) &   148 ($19.8\%$) &   747 \\
		Night  \spi{1} &   751 ($71.9\%$) &  84 ($8.0\%$) &   209 ($20.0\%$) & 1,044 \\
		Evening\spi{1} &   599 ($73.1\%$) &  67 ($8.2\%$) &   153 ($18.7\%$) &   819 \\
		Total          & 4,026 ($72.9\%$) & 449 ($8.1\%$) & 1,048 ($19.0\%$) & 5,523 \\
		\specialrule{.1em}{.05em}{.05em}
	\end{tabular}
	
	{\textit{1} -- denotes that data have frames with dust condition.}
\end{table}



\textit{Baselines and metrics for performance analysis}.
To measure the performance of our proposed solution for perception on the off-road environment in low visibility conditions, we have used the evaluation metrics most commonly found in the literature and competitions for performance evaluation of semantic segmentation algorithms. They are:
\begin{itemize}
	\item \textbf{Pixel accuracy} ($P_{acc}$). It is a simple accuracy metric that tells us the percentage of pixels in the image that are correctly classified. The \autoref{eq:pixel-acuracia} shows how this indicator is calculated, with $t_{i}$ representing the total number of pixels for the class $i$, and $\sum_{i}t_{i}$ representing the sum of all pixels belonging to all classes —the total amount of pixels in the image. Furthermore $n_{ii}$ represents the number of pixels of the class $i$ correctly inferred as belonging to the class $i$, and $\sum_{i}n_{ii}$ representing the total number of pixels correctly predicted in the whole image \citep{Long:2015:FCN:ieeecvpr,Liu:2018:AIR:Recent-progress-in-semantic-s}. This metric can also be expressed by class (Eq. \ref{eq:per-classe-pixel-acuracia}) instead of a global way.
	
	\begin{equation} \label{eq:pixel-acuracia}
	P_{acc}= \frac{\sum_{i}n_{ii}}{\sum_{i}t_{i}}
	\end{equation}
	
	\begin{equation} \label{eq:per-classe-pixel-acuracia}
	CP_{acc}= \frac{n_{ii}}{t_{i}}
	\end{equation}
	
	\item \textbf{Mean accuracy} ($mCP_{acc}$). The average accuracy among all classes can be calculated as shown by \autoref{eq:acuracia-media}. In this equation we have the sum of the accuracy calculated for each class  $\sum_{i}\frac{n_{ii}}{t_{i}}$ divided by the number of classes $n_{cl}$.
	
	\begin{equation} \label{eq:acuracia-media} 
	mCP_{acc}= \frac{1}{n_{cl}}\sum_{i}\frac{n_{ii}}{t_{i}}
	\end{equation}

	\item \textbf{Intersection over Union or Jaccard Index} ($IoU$). It is a statistic used to measure the diversity and similarity of sample sets. In the context of this research, it is a metric that quantifies the percentage of overlap between ground-truth and the segmentation mask inferred by the algorithm. The calculation is done by dividing the number of pixels in common between ground-truth and inferred mask — intersection $\cap$ — by the total number of pixels present, considering the ground-truth and inference — union $\cup$. The \autoref{eq:intersecao-uniao} presents the calculation process for this metric. It has $n_{ii}$ representing the number of pixels of the class $i$ correctly inferred to belong to the class $i$ (intersection $\cap$), the  $\sum_{j}n_{ji}$ representing the number of pixels in all classes $j$ that are inferred as belonging to the class $i$ (inference results), the $t_{i}$ representing the total number of pixels for the class $i$ (Ground-truth) and $t_{i}+\sum_{j}n_{ji}-n_{ii}$ representing the union.
	
	\begin{equation} \label{eq:intersecao-uniao}
	IoU= \frac{n_{ii}}{t_{i}+\sum_{j}n_{ji}-n_{ii}}
	\end{equation}
	
	\item \textbf{Mean Intersection over Union} ($mIoU$). The mean intersection over union between classes is very similar to the previous metric. However, it calculates the average of $IoU$ between classes. The \autoref{eq:media-da-intersecao-uniao} shows how this metric is calculated. Basically it is the sum of \autoref{eq:intersecao-uniao} divided by the number of classes $n_{cl}$ \citep{Long:2015:FCN:ieeecvpr,Liu:2018:AIR:Recent-progress-in-semantic-s}.
	
	\begin{equation} \label{eq:media-da-intersecao-uniao}
	mIoU = \frac{1}{n_{cl}}\sum_{i}IoU
	\end{equation}
	
	\item \textbf{Frequency Weighted Intersection over Union} (FWIoU). Refers to the average of the intersection over union between classes weighted by the frequency of occurrence as in \autoref{eq:frequencia-ponderada} \citep{Long:2015:FCN:ieeecvpr, Liu:2018:AIR:Recent-progress-in-semantic-s}.
	
	\begin{equation} \label{eq:frequencia-ponderada}
	FWIoU = \frac{1}{\sum_{k}t_{k}}\sum_{i}t_{i}IoU
	\end{equation}
\end{itemize}

\textit{Inference time evaluation}.
To estimate the inference time, we performed a sequence of 500 iterations to measure the mean, the standard deviation (SD), and calculate the boxplot parameters.

\textit{CMSNet arrangements}. In this work, we have presented the CMSNet framework. It is a configurable modular segmentation network framework that implements some state-of-the-art solutions for semantic segmentation. Their different modules can be configured to build several architectures solutions. It can use an output stride of 8 or 16, by choosing where the dilated convolution starts to be applied in the backbone pipeline. Furthermore, the architecture may be configured with either spatial pyramid pooling (SPP), atrous spatial pyramid pooling (ASPP), or global pyramid pooling (GPP). Besides, it may have a shortcut of high-resolution features. \autoref{tab:cmsnet-arrangements} shows the different arrangements and their configuration considered in the experiments. From here on, we use only the names defined in the \autoref{tab:cmsnet-arrangements} to refer to each of the arrangements.

\begin{table}[!t]
	\centering
	\scriptsize
	\caption{Different arrangements for CMSNet.}
	\label{tab:cmsnet-arrangements}
	\begin{tabular}{+c+c+c+c+c} 
		\specialrule{.1em}{.05em}{.05em}
		\bfseries Name &
		\bfseries Abbr. &
		\bfseries Output Stride & 
		\bfseries Pyramid  & 
		\bfseries Shortcut \\
		\specialrule{.1em}{.05em}{.05em}
		CMSNet-M0 & CM0 & 8    & GPP   &  No   \\
		CMSNet-M1 & CM1 & 8    & SPP      &  No   \\
		CMSNet-M2 & CM2 & 8    & ASPP     &  No   \\
		CMSNet-M3 & CM3 & 16   & GPP      & No    \\
		CMSNet-M4 & CM4 & 16   & SPP      & No    \\
		CMSNet-M5 & CM5 & 16   & ASPP     & No \\
		CMSNet-M6 & CM6 & 16   & GPP      & Yes \\
		CMSNet-M7 & CM7 & 16   & SPP      & Yes \\
		CMSNet-M8 & CM8 & 16   & ASPP     & Yes \\
		\specialrule{.1em}{.05em}{.05em}
	\end{tabular}
\end{table}

\textit{Training setup}. The training was performed in a computer with a GPU RTX 2060 with 6 GB and a 9th generation i7 processor, having six core and capable of run 12 threads. To accelerate the training process, we have split the proposed dataset and have used the subset, as shown in \autoref{tab:subset_composition}. After tunning the hyperparameter, we have included the validation set in the training processes to increase the training set diversity. Altogether, we have used 4.475 images for training, randomly distributed between all conditions and places. The strategy of using a subset of the data has allowed us to decrease training time for each CMSNet arrangement.
Each scenario takes 200 epochs to be trained using a batch of 4 images. We also have used artificial data augmentation techniques to help avoid over-fitting and increase training performance. We have used a learning rate of 0.007 with the first-order polynomial decaying until 0.

\subsection{Ablation study for CSMNet}
The \autoref{tab:summary-testes-com-configuracoes-para-backbone-mobilenetv2} shows the results of a investigation carry out with the architecture arrangements defined in \autoref{tab:cmsnet-arrangements}. In this ablation study was investigate how the different arrangements perform related with the metrics $mIoU$, $FWIoU$, $mCP_{acc}$, and $CP_{acc}$. 
For this experiment, we have used a test set having 1,048 images distributed by different places and conditions as specified in \autoref{tab:subset_composition}. Besides, the \autoref{tab:summary-testes-com-configuracoes-para-backbone-mobilenetv2} also shows the number of parameter demanded for each arrangement combination.

This study allows us to observe that the output stride smaller (8) makes a positive effect of 1\% in the ASPP $mIoU$ (CMSNet-M2 and CMSNet-M5). However, lower values for output stride harm the inference time, as can be seen in \autoref{subsec:results-on-kamino-dataset}. Despite increasing the processing time, the positive effect was not perceived on the $mIoU$ of all arrangements using this configuration, as is the case of CMSNet-M0 and CMSNet-M1. We also have noted that the metric $FWIoU$ suffers low variation independent of the CMSNet configuration.

In general, architectures with ASPP module (M2, M5, and M8) have performed better than ones with SPP (M1, M4, and M7), which in turn have achieved better results than GPP ones (M0, M3, and M6). However, the arrangements with ASPP demand more parameters than others. We also have noted that configurations with shortcuts (M6, M7, and M8) have not performed better, although they have more parameters than other arrangements.

We suppose that ASPP configurations have demanded more parameters because it has been implemented using standard 2D convolutions instead of the factored one used in the SPP module. The factored convolutions are composed of depthwise and pointwise convolution and are computationally less expensive.

\begin{table}[!t]
  \centering
  \scriptsize
	\caption{Tests with settings for \textit{backbone} MobileNetV2.}
	\label{tab:summary-testes-com-configuracoes-para-backbone-mobilenetv2}
  \begin{tabular}{+c+c+c+c+c+c} 
	\specialrule{.1em}{.05em}{.05em}
		\bfseries Name & 
		\bfseries mIoU\% & 
		\bfseries FWIoU\% & 
		\bfseries mCP\textsubscript{acc}\% & 
		\bfseries P\textsubscript{acc}\% & 
		\bfseries Param.\\
		\specialrule{.1em}{.05em}{.05em}
	 	CM0 & 84.66 & 95.72 & 94.33 & 97.78 & 2,144 k \\
	 	CM1 & 84.15 & 95.97 & 91.91 & 97.91 & 2,033 k \\
	 	CM2 & 86.98 & 96.51 & 92.11 & 98.21 & 4,408 k \\
	 	CM3 & 85.02 & 96.21 & 91.89 & 98.05 & 2,144 k \\
	 	CM4 & 85.25 & 96.30 & 91.88 & 98.09 & 2,033 k \\
	 	CM5 & 85.01 & 96.33 & 91.48 & 98.11 & 4,408 k \\
	 	CM6 & 80.67 & 96.08 & 85.99 & 97.97 & 2,150 k \\
	 	CM7 & 83.62 & 96.27 & 89.21 & 98.07 & 2,039 k \\
	 	CM8 & 84.02 & 96.31 & 89.72 & 98.09 & 4,414 k \\
    \specialrule{.1em}{.05em}{.05em}
	\end{tabular}
\end{table}

\subsection{Results on Kamino dataset}
\label{subsec:results-on-kamino-dataset}

\textit{Comparison with pre-trained networks}.
The \autoref{tab:methods_eval} shows the results for different arrangements presented in \autoref{tab:cmsnet-arrangements} compared with other architectures trained for fully urban environment. The architectures used for comparing were PSPNet \citep{Zhao:2017:PSPNet:ieeecvpr} and some variations of DeepLab -- MNV2, Xc65, and Xc71 \citep{Sandler:2018:MobileNetV2:IEEE-CVF,Chen:2018:deeplab:ieeeTPAMI,Chen:2018:EncoderDecoderWA:ECCV:deeplabv3Plus}. Cityscapes \citep{cordts:2016:cityscapes} was the urban dataset used in those networks. The link for pre-trained networks used in this experiment are: PSPNet \footnote{PSPNet url: \url{https://drive.google.com/file/d/1vZkk9nLvM9NNBCVCuEjnoXms30OMcZ8K}}, DeepLab+MNV2 \footnote{DeepLab+MNV2 url: \url{http://download.tensorflow.org/models/deeplabv3\_mnv2\_cityscapes\_train\_2018\_02\_05.tar.gz}}, DeepLab+Xc6 \footnote{DeepLab+Xc6 url: \url{http://download.tensorflow.org/models/deeplabv3\_cityscapes\_train\_2018\_02\_06.tar.gz}}, and DeepLab+Xc7 \footnote{DeepLab+Xc7 url: \url{http://download.tensorflow.org/models/deeplab\_cityscapes\_xception71\_trainfine\_2018\_09\_08.tar.gz}}.

For this experiment, we have included the main common classes between urban and off-road datasets. We have used the classes road, car, person, and background (everything else). ``All'' in \autoref{tab:methods_eval} mean $mIoU$ for all used classes. For each class, we have used the metric $IoU$. 

As observed previously in the \autoref{fig:test_pspnet_deeplab_cityscape_unpaved_roads}, we also can see quantitatively in the \autoref{tab:methods_eval} that pre-trained architectures with those urban datasets like Cityscapes do not perform so well in non-paved and off-road environments. Despite using more parameters, those architectures performed worst even in classes like car and person. 

We can see in the \autoref{tab:methods_eval} that the CMSNet-M0's most similar architectures (DeepLab+MNV2) have achieved worse results. It has reached 31.46\% of $mIoU$ (All) and has obtained only 3.57\% of $IoU$ for the class person. On the other hand, PSPNet and DeepLab+Xc65 have managed to reach 57.83\% and 55.68\% of $mIoU$, respectively. However, the results have been far from those achieved by our approach.

These results suggest that the perception subsystems being developed for autonomous vehicles focused in a well-paved urban environment may not be suitable for developing countries or will be restricted to a small set of roads in urban centers. As well, this restriction will limit the implementation of autonomous systems in cargo vehicles, such as buses and trucks.

\begin{table*}[!t]
	
  \centering
  \scriptsize
  \caption{Results of the semantic segmentation on the categories of the Kamino dataset.}
  \label{tab:methods_eval}
  \begin{tabular}{+l+c+c+c+c+c+c+c+c+c+c}
    \specialrule{.1em}{.05em}{.05em}
    \multirow{2}*{\bfseries Name}  & \multicolumn{6}{c}{\bfseries IoU (\%)}  &  \multicolumn{2}{c}{\bfseries Batch 1} &
    \multicolumn{2}{c}{\bfseries Batch 4} \\
     & 
    \bfseries Road &  
    \bfseries Car & 
    \bfseries Person &  
    \bfseries Bg & 
    \bfseries All & 
    \bfseries FPS & 
    \bfseries SD(\%) &
    \bfseries FPS & 
    \bfseries SD(\%) \\
    \specialrule{.1em}{.05em}{.05em}
	 CM0 & 95.72 & 75.92 & 71.06 & 95.96 & 84.66 & 19.16 & 4.92 & 20.59 & 8.77\\
	 CM1 & 95.96 & 74.16 & 70.22 & 96.27 & 84.15 & 19.37 & 4.08 & 20.52 & 5.88\\
	 CM2 & 96.51 & 78.74 & 75.89 & 96.78 & 86.98 & 16.46 & 3.24 & 17.03 & 8.37\\
	 CM3 & 96.23 & 76.71 & 70.63 & 96.49 & 85.02 & 28.87 & 5.38 & 32.65 & 8.95 \\
	 CM4 & 96.31 & 77.00 & 71.12 & 96.59 & 85.25 & 27.77 & 3.48 & 32.82 & 4.43 \\
	 CM5 & 96.35 & 75.43 & 71.63 & 96.62 & 85.01 & 27.14 & 3.78 & 30.24 & 3.88 \\
	 CM6 & 96.11 & 77.08 & 53.03 & 96.47 & 80.67 & 28.1 & 3.1 & 32.77 & 4.29 \\
	 CM7 & 96.27 & 75.57 & 66.05 & 96.60 & 83.62 & 27.1 & 4.79 & 32.81 & 3.44 \\
	 CM8 & 96.34 & 73.81 & 69.32 & 96.63 & 84.02 & 26.64 & 4.46 & 30.37 & 3.39 \\
	PSPNet       & 63.22 & 44.25 & 54.12 & 69.70 & 57.83 & 2.79 & 9.14 & -- & -- \\
	DMNV2 & 59.39 & 9.030 & 3.570 & 53.83 & 31.46 & 5.9 & 8.34 & -- & --\\
	DLX65 & 65.92 & 46.10 & 52.54 & 58.15 & 55.68 & 0.69 & 7.3 & -- & -- \\
	DLX71 & 63.09 & 55.30 & 8.660 & 64.93 & 47.99 & 2.32 & 8.0 & -- & -- \\
    \specialrule{.1em}{.05em}{.05em}
  \end{tabular}
\end{table*}

\textit{Inference time comparison}. We also have compared the frames per second (FPS) and have calculated the standard deviation (SD) achieved for each one of the arrangements in the \autoref{tab:cmsnet-arrangements}, PSPNet, DeepLab+MNV2 (DMNV2), DeepLab+Xc6 (DLX65), and DeepLab+Xc7 (DLX71). In the \autoref{tab:methods_eval}, we have the inference time in FPS and standard deviation in percentage for each one of these architectures. Such times were calculated using a GPU RTX2060 and a CPU core i7. 

As it can be seen, our architecture arrangements have achieved higher FPS and lower standard deviation than PSPNet and DeepLabs variations. The worst-case in FPS has been produced by DeepLab+Xc6, and the best case has been achieved by CMSNet-M4. Our solutions have achieved approximately 4\% of standard deviation while other architectures have obtained 8\%.

In those testes, our best performance in accuracy (CMSNet-M2) was also our worst-case in inference time. We also have calculated the inference for a batch of four images. We have achieved an average improvement of 1 FPS for output stride 8 (CM0, CM1, and CM2) and 4 FPS for output stride 16 (CM3, CM4, CM5, CM6, CM7, and CM8). 

\begin{figure*}[!t]
	\begin{center} 
	\includegraphics[width=\linewidth]{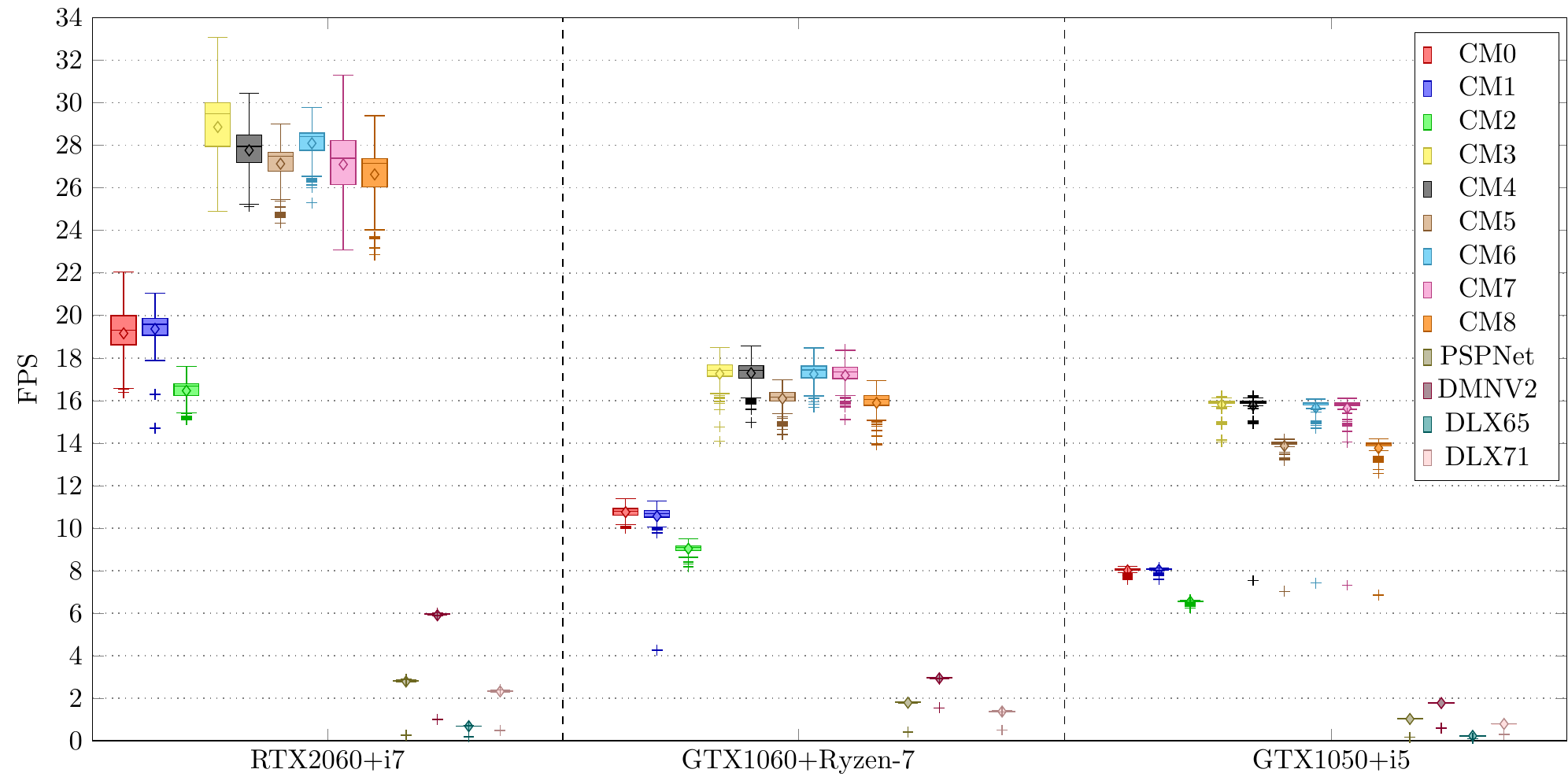}
	\end{center} 
	\caption{Inference time is shown in FPS (box-plot) as a function of the architecture model and hardware platform. The models were tested on three different platforms -- RTX2060+core-i7, GTX1060+Ryzen7, and GTX1050+core-i5.}
	\label{fig:inference_time}
\end{figure*}

\textit{Inference on different hardware.} We have also made tests in other hardware configurations. The \autoref{fig:inference_time} shows the results for a GPU GTX1050, a GPU GTX1060 with a CPU Ryzen7, and a GPU RTX2060 with a CPU core i7. For this comparison, we have used the box-plot graphic to permit us to observe data dispersion as the inferred FPS does not obey a normal distribution.

We have achieved the best result for inference with the CMSNet-M3 on the GPU RTX2060 and the worst case with the DeepLab+Xc65 on GPU GTX1050. Among the architectures composed with the framework CMSNet, those using output stride (OS) 16 performed better FPS than those one using output stride 8. The Global Pyramid Pooling (GPP) module (CM0, CM3, and CM6) achieved the best FPS regarding their output stride and shortcut strategies groups. On the other hand, the Atrous Spatial Pyramid Pooling (ASPP) module (CM2, CM5, and CM8) has achieved the worst FPS results considering the same groups.

In our observation, we have seen that the inference time performance for those different architecture has held the proportion regarding the computation power of each platform. Our proposed architectures have always performed better than the other ones used in the comparison. In all platforms, the CM2 has had the worst CMSNet's FPS, and DLX65 has had the worst inference time between all tested architectures.

\subsection{Results on DeepScene dataset}
\label{subsection:results-on-deepscene-dataset}
We have also compared some configuration generated by the framework CMSNet with architectures proposed in similar works published in the last years. The works presented in \cite{Semantic-Forested:Valada:2016} and \cite{Real-Time-Semantic-Off-Road:Maturana:2018} have been trained and compared in the DeepScene dataset \citep{Semantic-Forested:Valada:2016}. This dataset does not have the magnitude of our one proposed in this research, but it has been the possible way to compare the solutions' performance once those works have not published their source code.

\cite{Real-Time-Semantic-Off-Road:Maturana:2018} has presented two architectures: the FCN-based \citep{Long:2015:FCN:ieeecvpr} cnns-fcn with CNN-S backbone \citep{Chatfield:2014:BMVC} for feature extraction, and the dark-fcn with Darknet's backbone \citep{darknet13}. Those architectures have been trained and compared using the resolution $227\times227$ and $448\times448$. 

In the other hand, \cite{Semantic-Forested:Valada:2016} has proposed the UpNet built from a VGG backbone \citep{simonyan:2015:vgg}. The UpNet is an FCN similar architecture. However, there are some modifications in the last layer of VGG and on the number of upsampling steps. That architecture has been trained and compared using the resolution $300\times300$. 

Regarding CMSNet, we have trained CMSNet-M0 with a resolution of $300\times300$ (CM0-300) and with resolution $448\time448$ (CM0-448). Also, we have trained the arrangement with GPP (\autoref{fig:global-pooling-module}) for output stride (OS) 16 (CM3-300 and CM3-448).

The \autoref{tab:comparacao-avaliacao-metrica} shows the result $IoU$ per class and the $mIoU$. As it can be seen, the variations of architecture composed in CMSNet framework have reached better results than the networks proposed in \cite{Real-Time-Semantic-Off-Road:Maturana:2018} (cnns-fcn-227, dark-fcn-448). The CMNet's variation (CM0-300, CM0-448, CM3-300, and CM3-448) have reaches 78.89\%, 80.94\%, 77.68\%, and 79.37\% of $mIoU$ against 58.51\%, and 60.61\% of \cite{Real-Time-Semantic-Off-Road:Maturana:2018}.


\begin{table*}[!t]
  \centering
  \scriptsize
  \addtolength{\leftskip} {-2cm}
  \addtolength{\rightskip}{-2cm}
   \caption{Results of the semantic segmentation on the categories of the DeepScene dataset.}
  \label{tab:comparacao-avaliacao-metrica}
  \begin{tabular}{+l|+c +c +c +c +c|+c|+c|+c}
    \specialrule{.1em}{.05em}{.05em}
      & \multicolumn{6}{c|}{\bfseries IoU (\%)}  &  &  \\
    \cline{2-7}
    \bfseries Method & \bfseries Trail &  \bfseries Grass & \bfseries Vegetation & \bfseries Sky & \bfseries Obstacle & \bfseries mIoU & \bfseries FPS & \bfseries StdDev \\
    \specialrule{.1em}{.05em}{.05em}
	CM0-300      & 84.87 & 86.73 & 89.17 & 90.21 & 43.46 & \textbf{78.89} & \textbf{21.10} & 3.96\% \\
	CM0-448      & 86.70 & 87.72 & 89.78 & 91.06 & 49.42 & \textbf{80.94} &         16.07  & 2.96\% \\
	CM3-300      & 82.47 & 85.58 & 88.45 & 89.40 & 42.49 & \textbf{77.68} & \textbf{23.75} & 5.92\% \\
	CM3-448      & 84.69 & 87.06 & 89.46 & 90.30 & 45.35 & \textbf{79.37} & \textbf{21.33} & 4.66\% \\
	Upnet-300    & 85.03 & 86.78 & 90.90 & 90.39 & 45.31 & \textbf{79.68} &         20.09  & 9.47\% \\
	cnns-fcn-227 & 85.95 & 85.34 & 87.38 & 90.53 &  1.84 &         58.51  &          9.90  & 1.58\% \\
	dark-fcn-448 & 88.80 & 87.41 & 89.46 & 93.35 &  4.61 &         60.61  &         18.99  & 3.47\% \\
    \specialrule{.1em}{.05em}{.05em}
  \end{tabular}
\end{table*}

On the other hand, regarding the UpNet proposed in \cite{Semantic-Forested:Valada:2016}, our results of $mIoU$ were approximately equivalents. CM0-300 and CM0-448 were better, CM3-300 was equal, and CM3-448 was slightly inferior.

We have implemented the evaluation of inference time for these architectures. In the \autoref{tab:comparacao-avaliacao-metrica} are shown the inference time results in a GTX 1060. Except for the CM0-448, all our proposed solutions are faster than the others. 

\subsection{Analysis on adverse environmental conditions}

To evaluate the system behavior in the adverse conditions of visibility, we have separated the dataset in portions (Tab. \ref{tab:subset_composition}) and have calculated the metric with an incremental transition between subsets. We have measured the IoU results with conditions ranging from 100\% daytime images to 100\% images in poor visibility. The poor conditions included in these tests are rainy, dusty, night, and night with dust.

\begin{figure}[!t]
	\captionsetup[subfigure]{font=scriptsize,labelformat=empty}
	\begin{subfigure}[b]{0.245\linewidth}
		\caption{Daytime}
		\includegraphics[width=\linewidth]{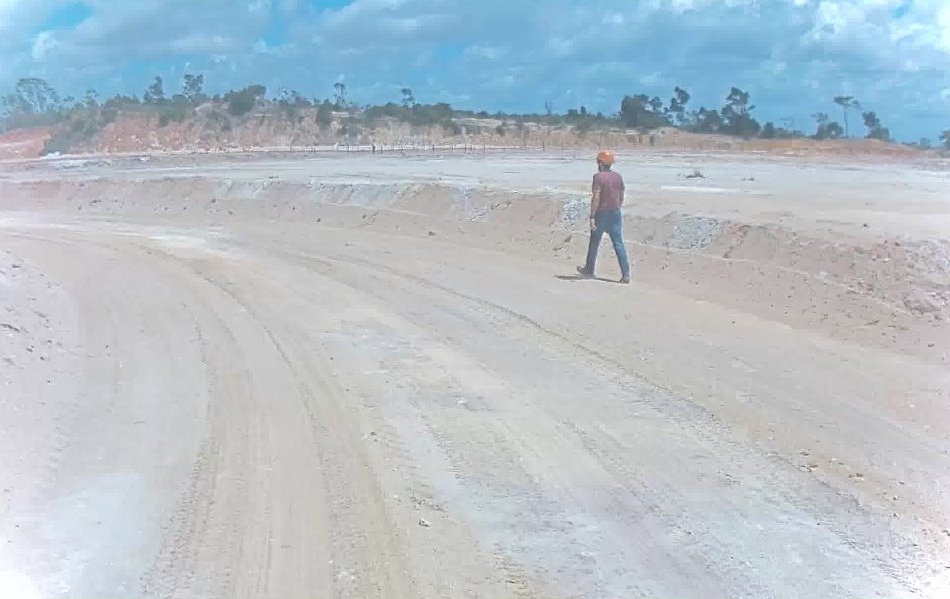}
	\end{subfigure}%
	\hfill
	\begin{subfigure}[b]{0.245\linewidth}
		\caption{Dusty}
		\includegraphics[width=\linewidth]{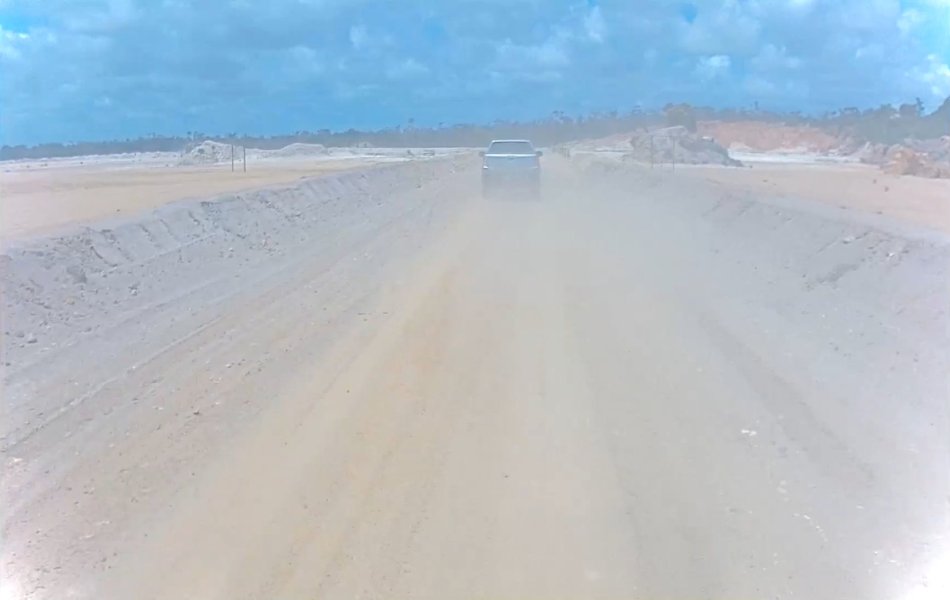}
	\end{subfigure}%
	\hfill
	\begin{subfigure}[b]{0.245\linewidth}
		\caption{Night}
		\includegraphics[width=\linewidth]{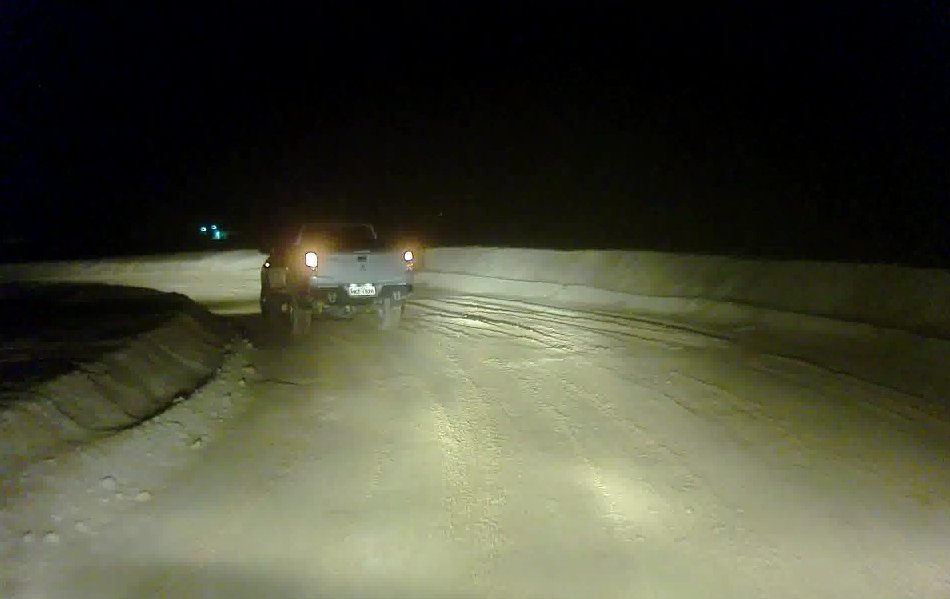}
	\end{subfigure}%
	\hfill
	\begin{subfigure}[b]{0.245\linewidth}
		\caption{Night-dusty}
		\includegraphics[width=\linewidth]{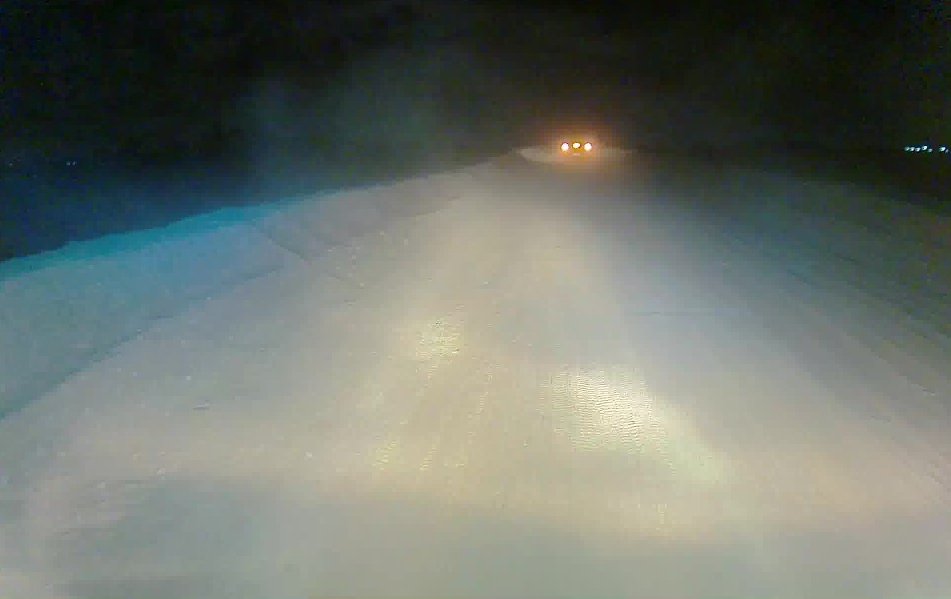}
	\end{subfigure}%
	\vspace{0.003\linewidth}
	
	\begin{subfigure}[b]{0.245\linewidth}
		\includegraphics[width=\linewidth]{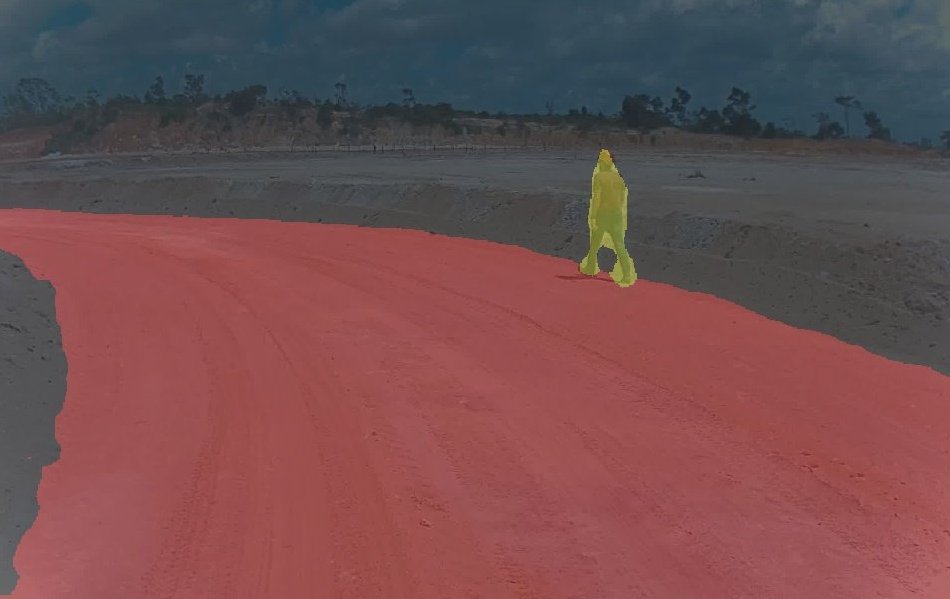}
	\end{subfigure}%
	\hfill
	\begin{subfigure}[b]{0.245\linewidth}
		\includegraphics[width=\linewidth]{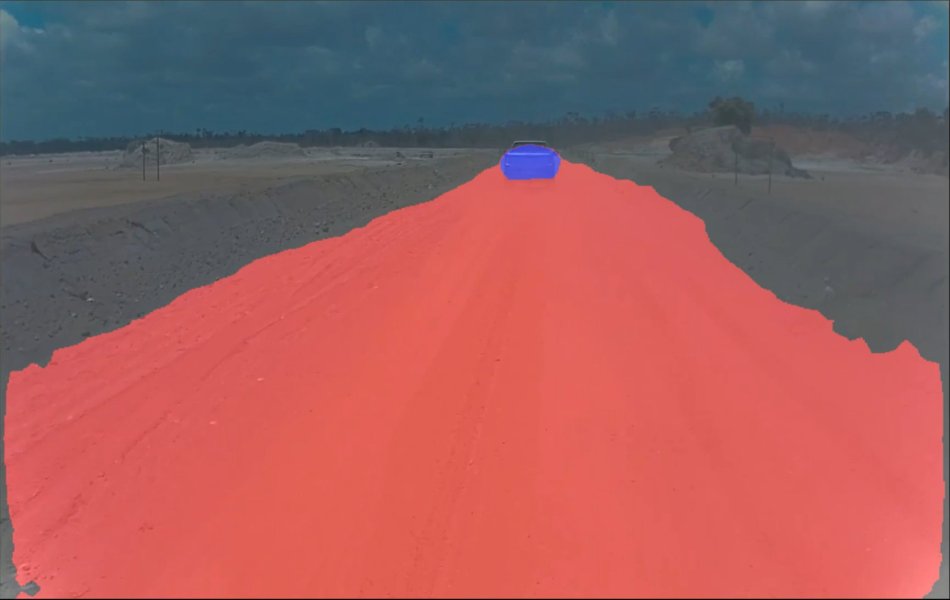}
	\end{subfigure}%
	\hfill
	\begin{subfigure}[b]{0.245\linewidth}
		\includegraphics[width=\linewidth]{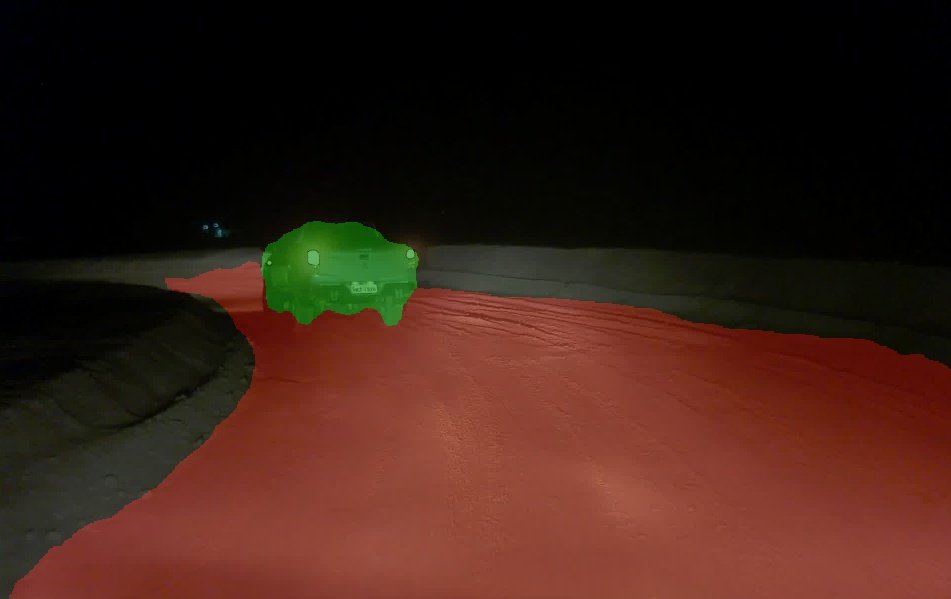}
	\end{subfigure}%
	\hfill
	\begin{subfigure}[b]{0.245\linewidth}
		\includegraphics[width=\linewidth]{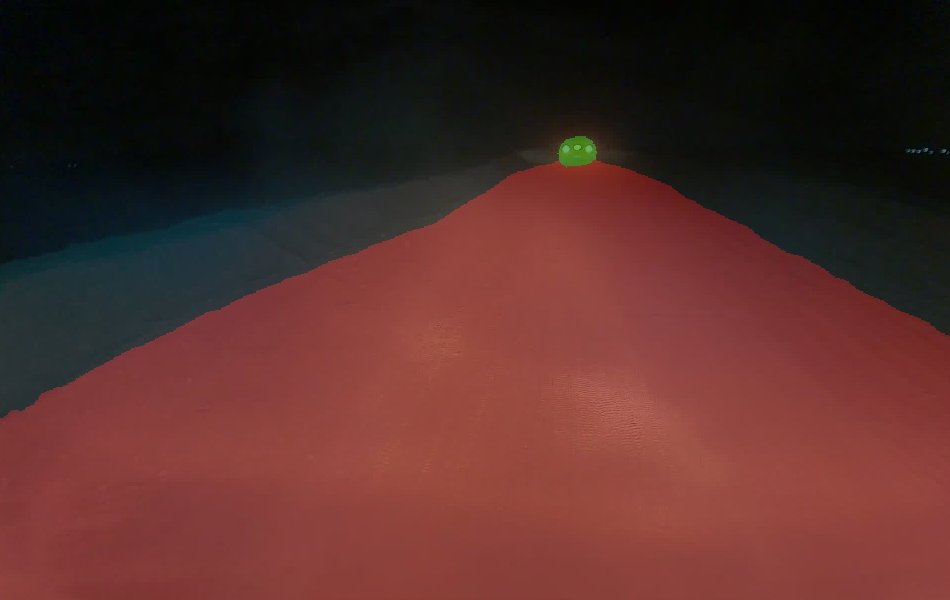}
	\end{subfigure}%
	\vspace{0.003\linewidth}
	
	\begin{subfigure}[b]{0.245\linewidth}
		\includegraphics[width=\linewidth]{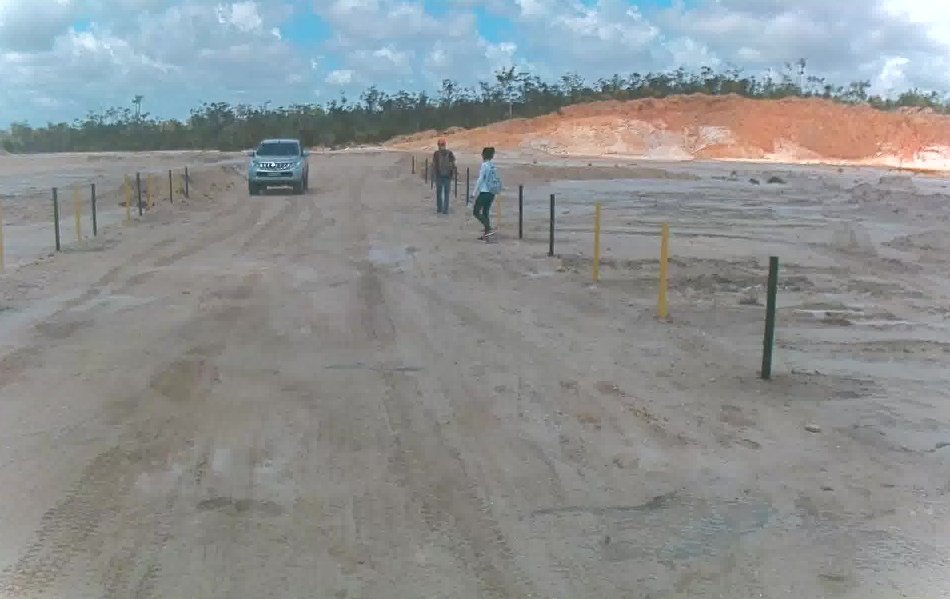}
	\end{subfigure}%
	\hfill
	\begin{subfigure}[b]{0.245\linewidth}
		\includegraphics[width=\linewidth]{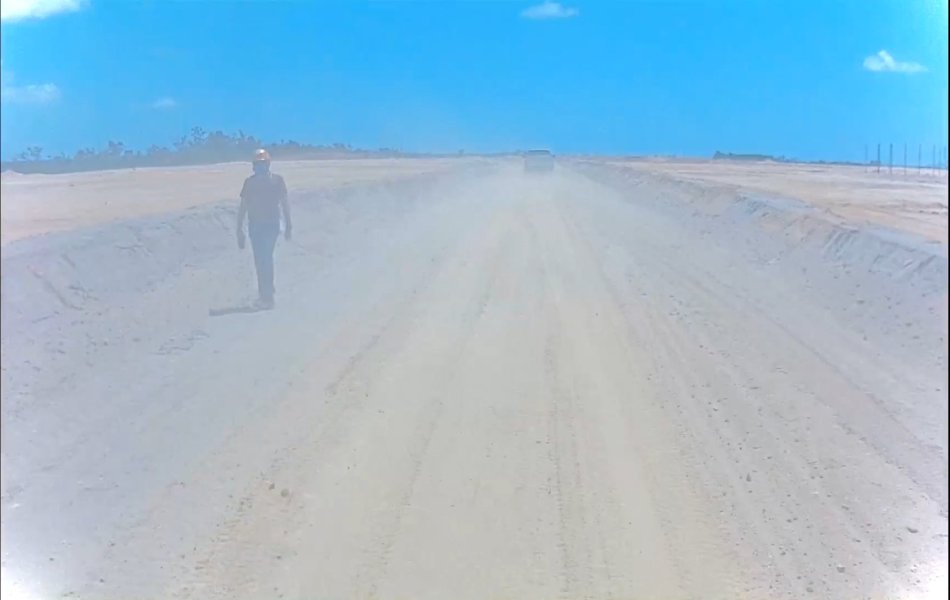}
	\end{subfigure}%
	\hfill
	\begin{subfigure}[b]{0.245\linewidth}
		\includegraphics[width=\linewidth]{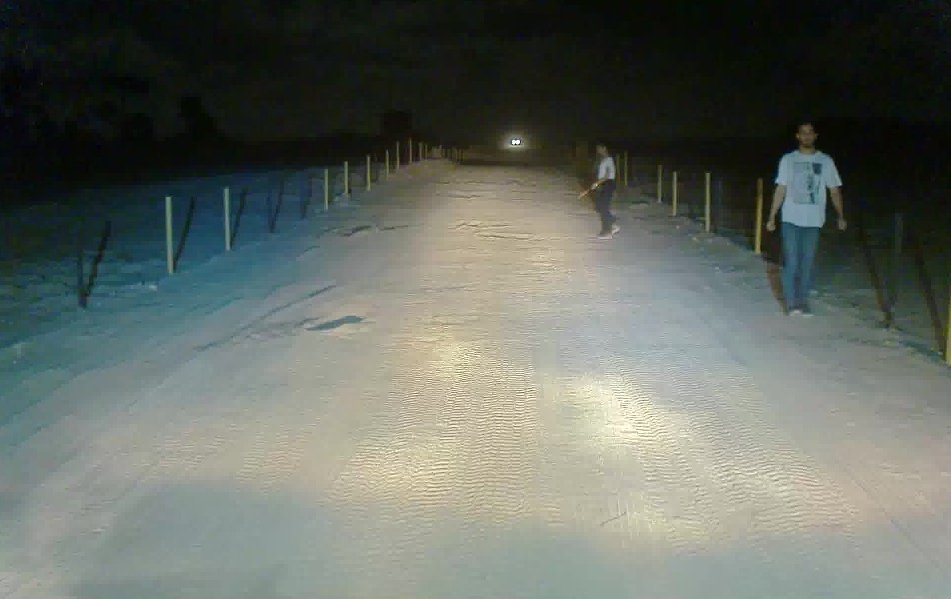}
	\end{subfigure}%
	\hfill
	\begin{subfigure}[b]{0.245\linewidth}
		\includegraphics[width=\linewidth]{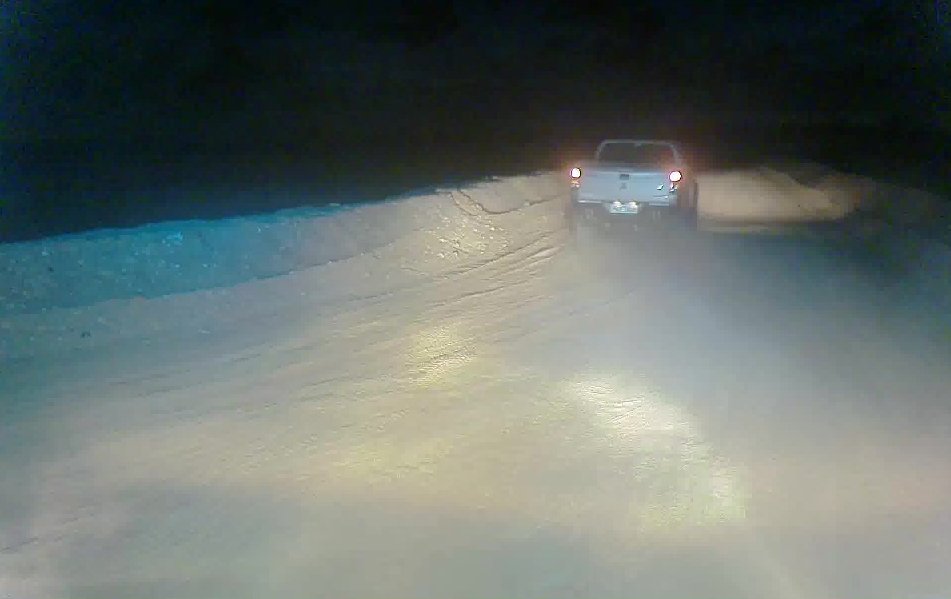}
	\end{subfigure}%
	\vspace{0.003\linewidth}
	
	\begin{subfigure}[b]{0.245\linewidth}
		\includegraphics[width=\linewidth]{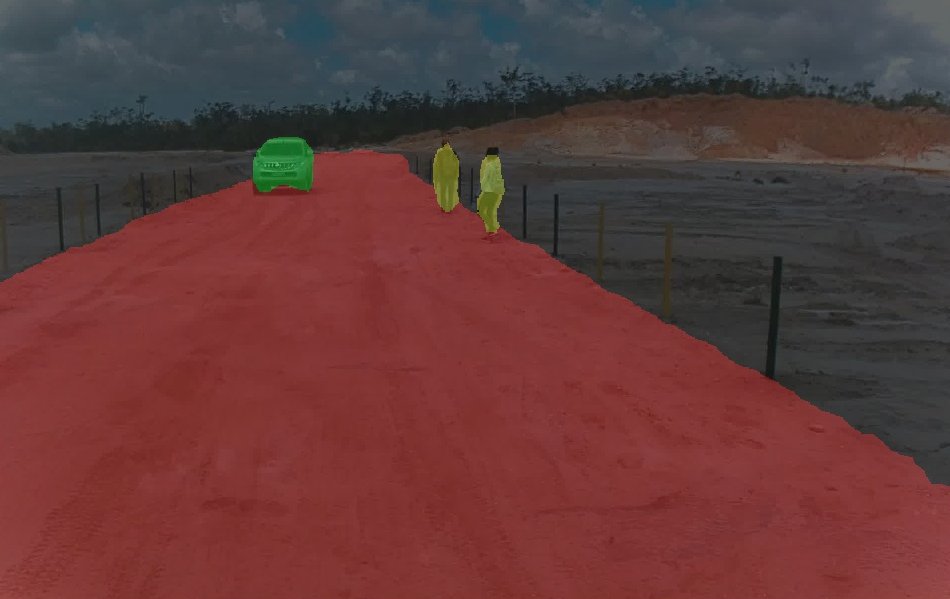}
	\end{subfigure}%
	\hfill
	\begin{subfigure}[b]{0.245\linewidth}
		\includegraphics[width=\linewidth]{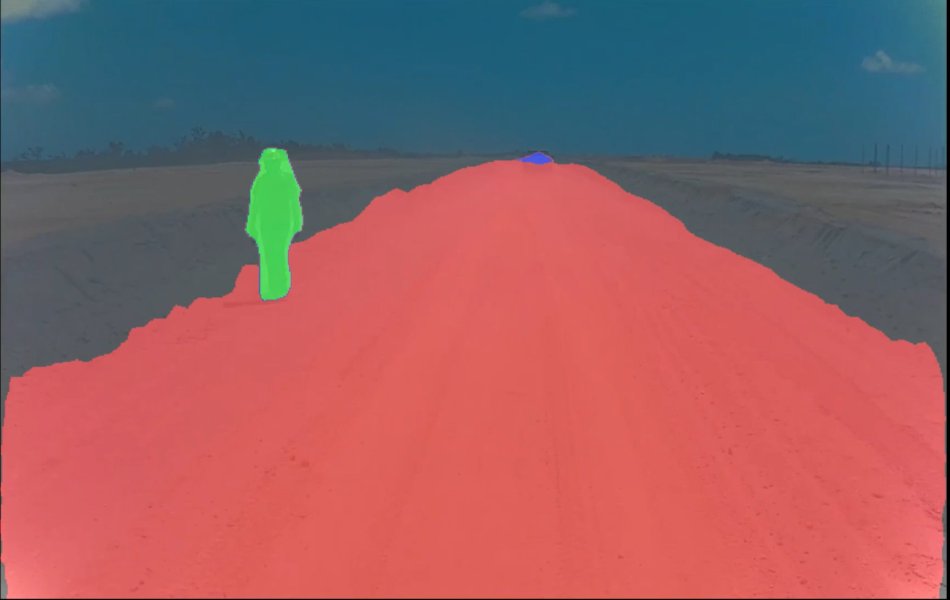}
	\end{subfigure}%
	\hfill
	\begin{subfigure}[b]{0.245\linewidth}
		\includegraphics[width=\linewidth]{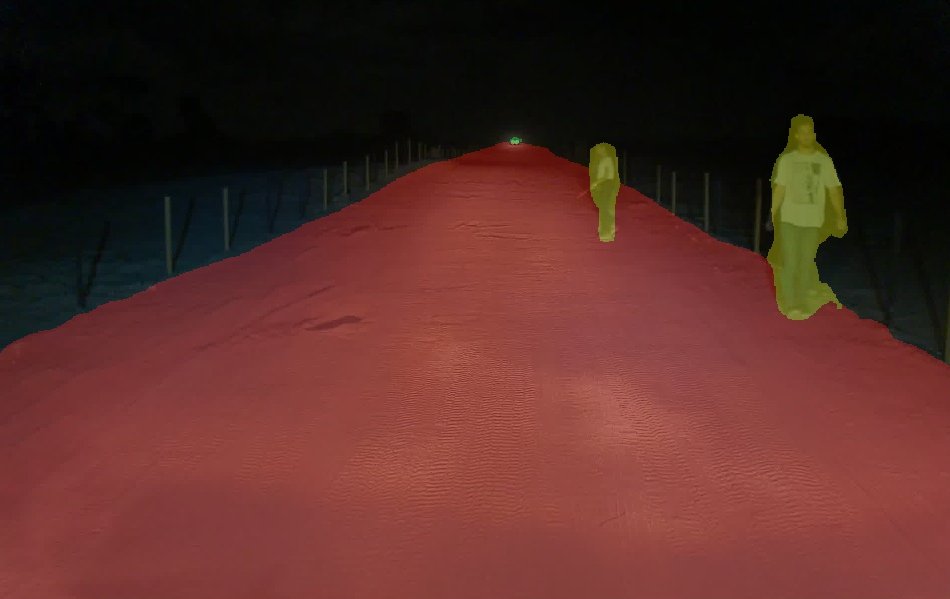}
	\end{subfigure}%
	\hfill
	\begin{subfigure}[b]{0.245\linewidth}
		\includegraphics[width=\linewidth]{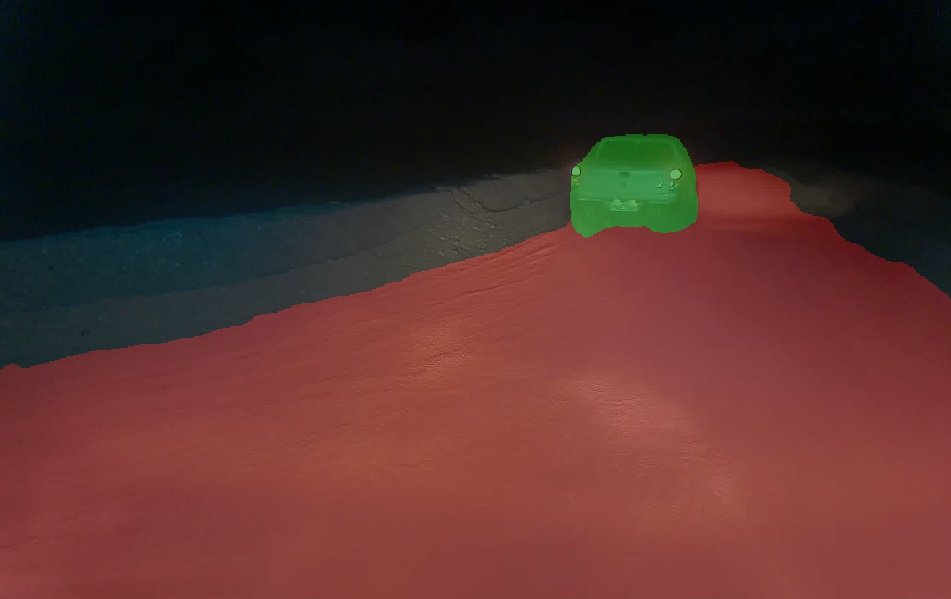}
	\end{subfigure}%

	\caption{Inference in different condition on off-road track.}
	\label{fig:daytime_dusty_night_and_night_dusty}
\end{figure}

\textit{Dusty condition.}
The subsets used in this evaluation were restricted to images collected in the off-road test track. Such an environment was the place where we have got dusty conditions, then we have used daytime images in the same locality as a good visibility counterpart. To get the dust condition, we used a pickup passing crossing in front of the cameras to raise dust on the test track. \autoref{fig:daytime_dusty_night_and_night_dusty} shows some pictures and their respective segmentation on the off-road track, including daytime, dusty, night, and night with dust.

\begin{figure}[!t]
	\begin{center} 
		\includegraphics[width=\linewidth]{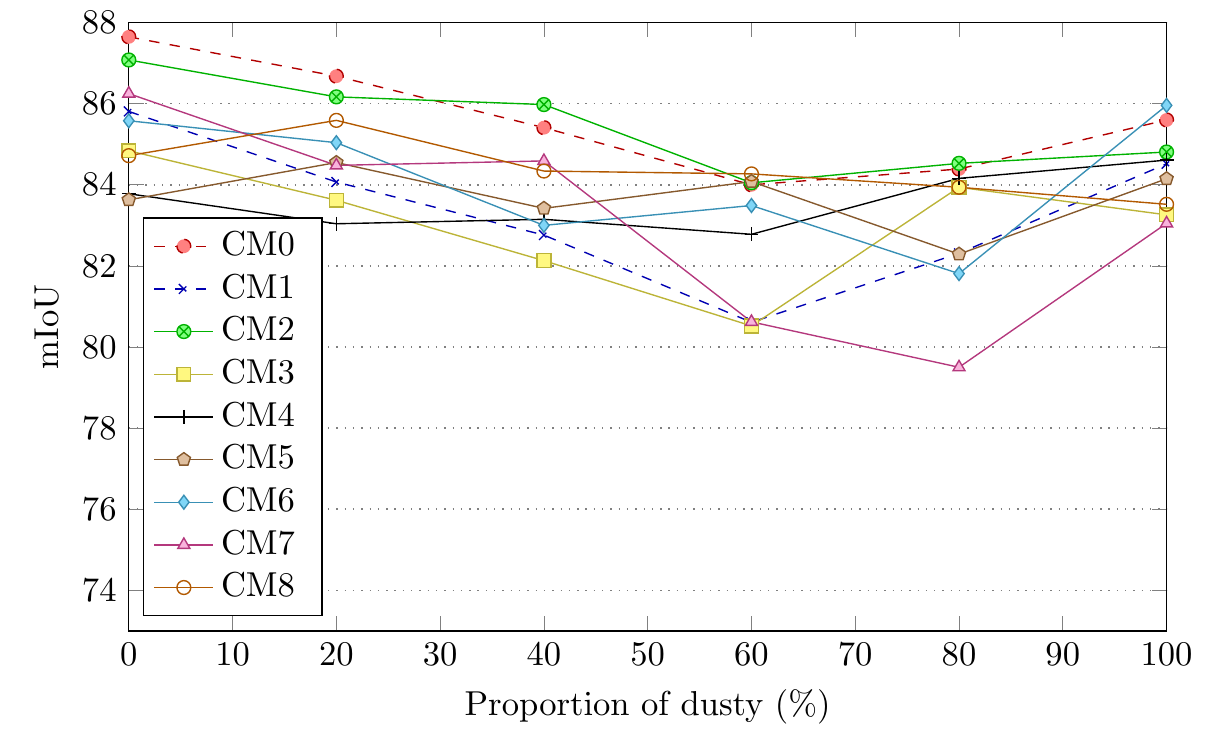}
	\end{center}
	\caption{Day vs. dusty condition evaluation.}
	\label{fig:day_dusty_condition}
\end{figure}

The graphic in \autoref{fig:day_dusty_condition} shows a downward trend in inference quality (IoU) as more images with dust are inserted into the test, and less good quality images are used. As can be seen, the configurations with output stride 8 (CM0, CM1, and CM2) are the ones that suffer less from increasing dusty condition images.  The best result, in a daytime condition, is 87.54\% of IoU, whereas it is 85.60\% in completely dusty circumstances.

\begin{figure}[!t]
	\begin{center} 
		\includegraphics[width=\linewidth]{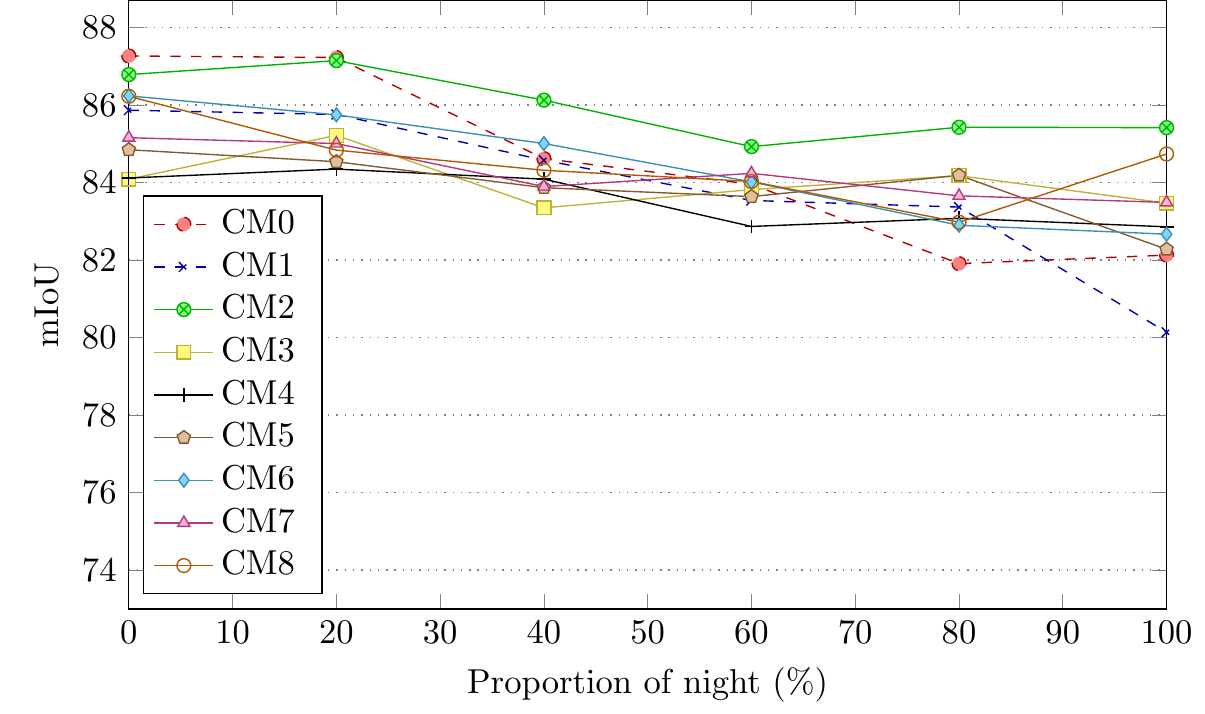}
	\end{center}
	\caption{Day vs. night condition evaluation.}
	\label{fig:day_night_condition}
\end{figure}

\textit{Night condition.}
The subsets used in this evaluation were also restricted to images collected in the off-road test track. However, we have replaced dust with the night in the bad visibility images. \autoref{fig:daytime_dusty_night_and_night_dusty} shows images and their segmentation for this situation. The \autoref{fig:day_night_condition} shows the graphic with the test results. The axis x (\%) represents the proportion of night images in the evaluation, and the axis y (mIoU) represents the inference performance archived by each configuration of CMSNet. The CM0 and CM1 (OS8 with GPP and SPP) were the architectures that most have decreased performance with the insertion of impairments. CM0 has decreased performance by over 5\%, and CM1 has lost almost 6\%. On the other hand, CM3 and CM4 (OS16 with GPP and SPP) have their performance decreased by the only 1\%. The best result in the night condition has been 85.42\% of IoU.

\begin{figure}[!t]
	\begin{center} 
		\includegraphics[width=\linewidth]{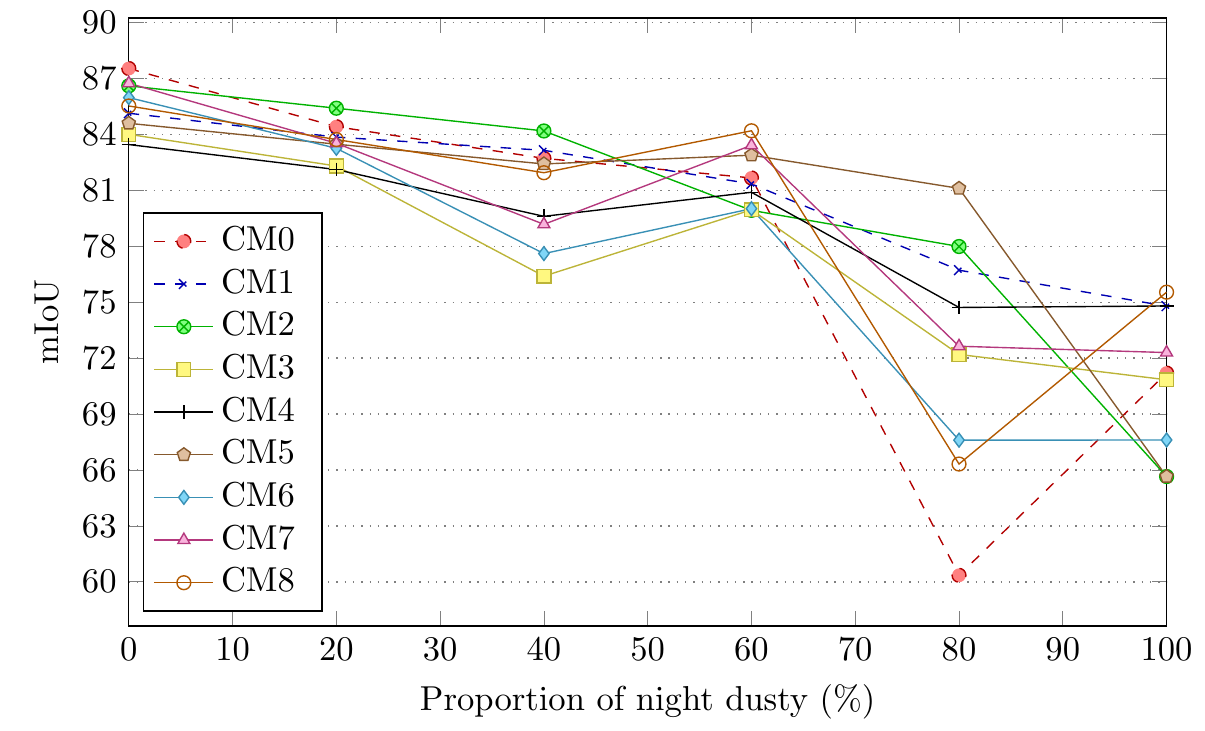}
	\end{center}
	\caption{Day vs. night dusty condition evaluation.}
	\label{fig:day_night_dusty_condition}
\end{figure}

\textit{Night with dust.}
In this test, we have mixed good quality images and images collected during the night and having dust (Fig. \ref{fig:daytime_dusty_night_and_night_dusty}). To generate the dust during the night, we have used the same strategy of having a pickup passing crossing in front of the vehicle. The degradation of inference quality in this scenario worse than the previous ones. The CM2 has lost about 21\% and CM6 near to 19\%, whereas the CM1 has lost 11\%, CM4 has degraded about 9\% and CM8 11,69\% (Fig. \ref{fig:day_night_dusty_condition}). In this scenario, we have achieved the best inference result of about 75\% of mIoU with the configurations CM1, CM4, and CM8.

\textit{Rainy condition.}
The subsets used in rainy tests were different from the previous ones. For this test, we have used good quality condition daytime images and bad condition images collected in unpaved roads in the metropolitan region of Salvador-BA (Fig. \ref{fig:daytime_dusty_night_and_night_dusty}). We have used this strategy to avoid getting the car stuck in the mud on the off-road test track.  

\begin{figure}[!t]
	\begin{center} 
		\includegraphics[width=\linewidth]{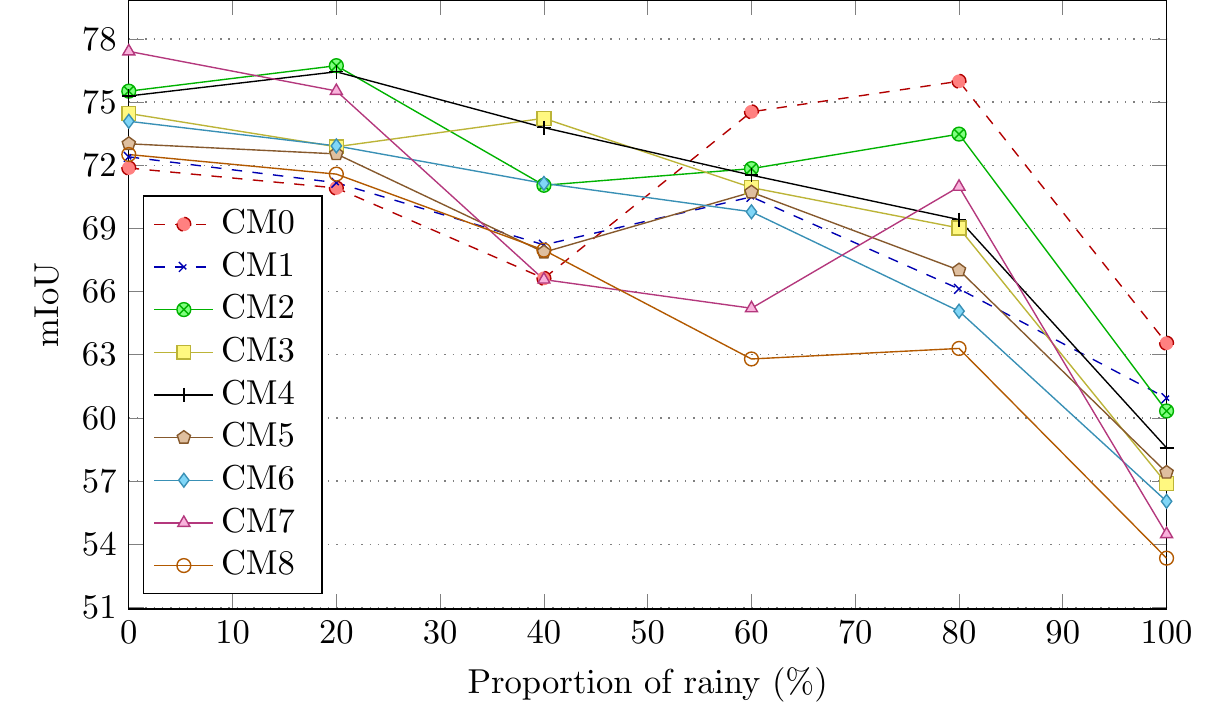}
	\end{center}
	\caption{Day vs. rainy condition evaluation.}
	\label{fig:day_rainy_condition}
\end{figure}

In the rainy condition scenario, inference degradation was even worse. We have the configuration CM7 with quality degradation of about 23\% and CM0 with mIoU degradation of 8\% (Fig. \ref{fig:day_rainy_condition}). The best inference result in this scenario, considering 100\% of daytime images, was near to 77\%, and with 100\% rainy condition, was 63.55\%.

\subsection{Synthetic Impairments}
In addition to collecting data in conditions of low visibility, we also have performed tests in adverse conditions with impairments generated synthetically. We have created fog and noise. In both situations, we have used the whole dataset shown in the \autoref{tab:subset_composition}.

\begin{figure}[!t]
	\begin{center} 
		\includegraphics[width=\linewidth]{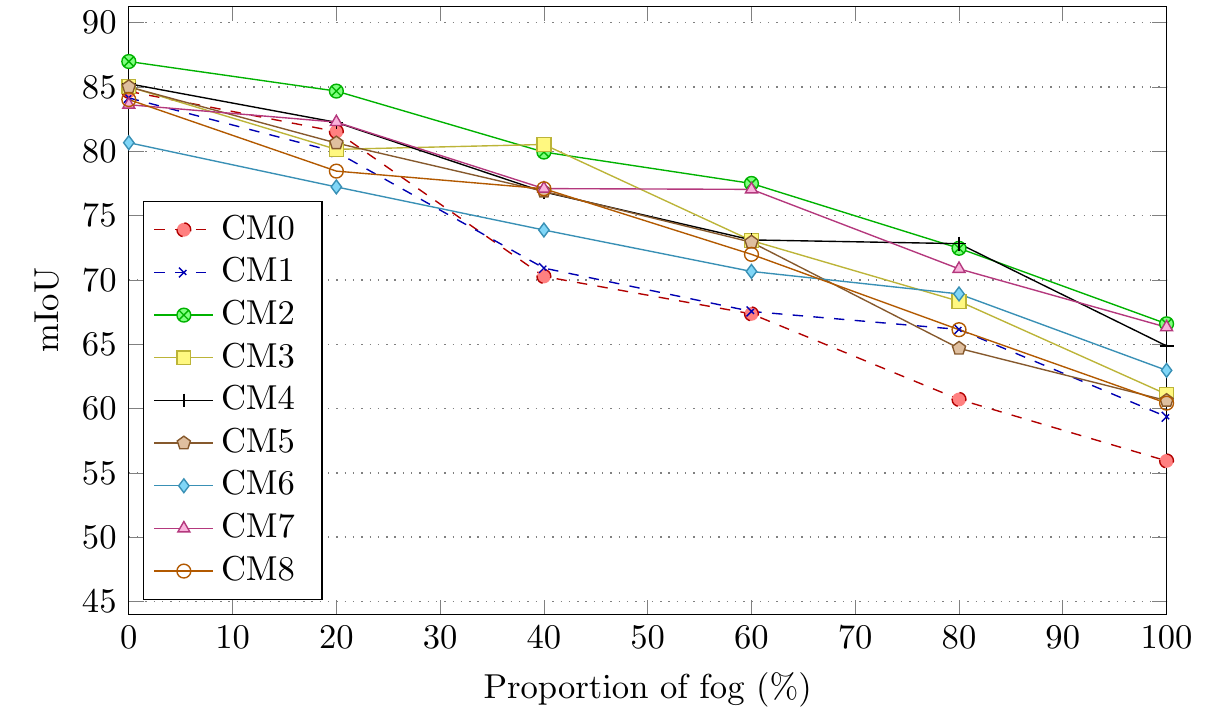}
	\end{center}
	\caption{Synthetic fog over the image.}
	\label{fig:fog_impairment}
\end{figure}

\textit{Synthetic fog.}
For the fogy condition, we have used a strategy similar to the previous testes where we have all the images without fog and start to do evaluations changing the dataset proportion by inserting fog images (going from 0\% until 100\%). For this teste (Fig. \ref{fig:daytime_dusty_night_and_night_dusty}), we have observed that the degradation of inference quality behaves like near a linear function. As it can be seen in \autoref{fig:fog_impairment}, the mIoU has been reduced by about  29\% for CM0 architecture (worst case) and has been decreased by 18\% for CM6 (best situation). In this test, the best inference result for daytime has been 86.98\% (CM2) of mIoU while considering 100\% of fog has been 66.59\% (CM2) of mIoU.

\begin{figure}[!t]
	\begin{center} 
		\includegraphics[width=\linewidth]{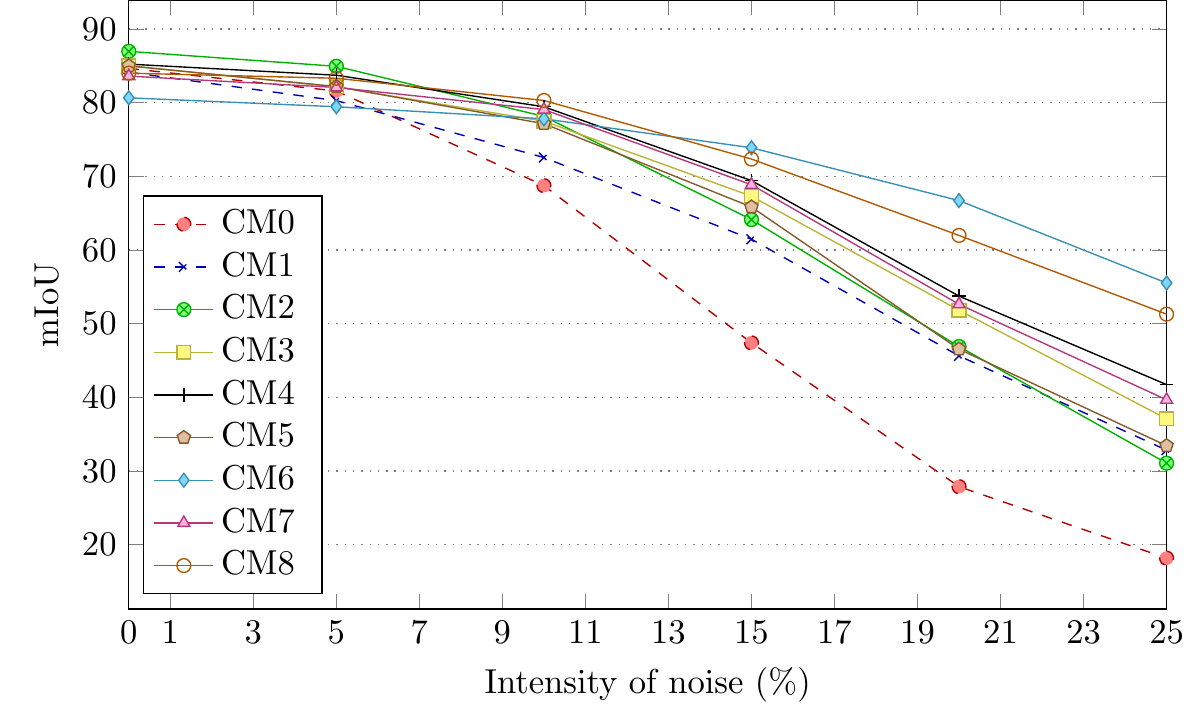}
	\end{center}
	\caption{Additive Gaussian noise over the image.}
	\label{fig:noise_impairment}
\end{figure}

\textit{Synthetic noise.}
To compose the dataset with artificial noise, we have used a different strategy from the previous tests. Instead of gradually replacing images without impairments with ones having the condition, we have increased the severity of the noise over images signal for all samples simultaneously. We have started the inference with 0\% of noise and have evaluated until 25\% of noise. Figures \ref{fig:daytime_dusty_night_and_night_dusty} and \ref{fig:noise_impairment} show the result. The worst degradation has been produced by CM0 with mIoU 67\% smaller, and the less intense degradation has been achieved by CM6 with a mIoU decrease of 25\%. The best inference result, considering 25\% of noise over image signal, has been 55.53\% (CM6) of mIoU.

\subsection{Comparing situations}
\autoref{table_conditions} shows the mIoU achieved by different configurations of CMSNet in diverse conditions of visibility, and \autoref{tab:degradation} shows the level of mIoU degradation achieved by each architecture on each scenario. Regarding the tests carried out on the off-road test track, the situation with night and dust has had the worst mIoU degradation in comparison with daytime images (Tab. \ref{tab:degradation}), and the worst absolute mIoU for all architectures (Tab. \ref{table_conditions}). On the other hand, the day dusty condition has had the best results related to inference quality degradation and absolute mIoU. Concerning the tests carried out with synthetic impairments, the fog has been less harmful than the noise. We also have noticed that rain has been more damaging to the inference quality than dust and night.

\begin{table*}[!t]
	\centering
	\scriptsize
	\addtolength{\leftskip} {-2cm}
	\addtolength{\rightskip}{-2cm}
	\caption{Comparison of mIoU for the evaluated methods on the different environmental conditions of our Kamino dataset during at daytime and nigth according to \autoref{tab:subset_composition}. ``All'' column is the averaged mIoU to a fully balanced set from the all the other subsets.}
	\label{table_conditions}
	\begin{tabular}{+l|+c +c +c +c| +c +c|+c +c +c}
		\specialrule{.1em}{.05em}{.05em}
		& \multicolumn{9}{c}{\bfseries mIoU (\%)}    \\
		\cline{2-10}
		\bfseries Method & \bfseries D. Off-R. & \bfseries Dust & \bfseries Night & \bfseries N.Dust & \bfseries D. Unpaved & \bfseries Rain & \bfseries All & \bfseries Fog & \bfseries Noise \\
		\specialrule{.1em}{.05em}{.05em}
		CM0  & \bfseries 87.65 & \bfseries 85.60 & 82.13 & 71.20 & 71.87 & \bfseries 63.55 & 84.66 & 55.93 & 18.17 \\ 
		CM1  & 85.81 & 84.52 & 80.14 & \bfseries 74.79 & 72.40 & 60.94 & 84.15 & 59.36 & 32.84 \\ 
		CM2  & \bfseries 87.08 & 84.81 & \bfseries 85.42 & 65.65 & \bfseries 75.52 & 60.33 & \bfseries 86.98 & \bfseries 66.59 & 31.07 \\ 
		CM3  & 84.84 & 83.26 & \bfseries 83.47 & 70.84 & 74.46 & 56.88 & \bfseries 85.02 & 61.11 & 37.09 \\ 
		CM4  & 83.78 & 84.61 & 82.86 & \bfseries 74.80 & \bfseries 75.29 & 58.58 & \bfseries 85.25 & 64.87 & 41.73 \\ 
		CM5  & 83.63 & 84.15 & 82.28 & 65.65 & 73.02 & 57.41 & \bfseries 85.01 & 60.62 & 33.44 \\ 
		CM6  & 85.58 & \bfseries 85.96 & 82.67 & 67.61 & 74.09 & 56.04 & 80.67 & 62.97 & \bfseries 55.53 \\ 
		CM7  & 86.25 & 83.05 & \bfseries 83.49 & 72.30 & \bfseries 77.41 & 54.48 & 83.62 & \bfseries 66.32 & 39.67 \\ 
		CM8  & 84.72 & 83.52 & \bfseries 84.74 & \bfseries 75.54 & 72.51 & 53.34 & 84.02 & 60.42 & \bfseries 51.30 \\ 
		
		\specialrule{.1em}{.05em}{.05em}
	\end{tabular}
\end{table*}

\begin{table}[!t]
	\centering
	\scriptsize
	\addtolength{\leftskip} {-2cm}
	\addtolength{\rightskip}{-2cm}
	\caption{Comparison of mIoU degradation for the evaluated methods on the different environmental conditions of our Kamino dataset during at daytime and nigth according to \autoref{tab:subset_composition}.}
	\label{tab:degradation}
	\begin{tabular}{+l|+c +c +c +c +c +c}
		\specialrule{.1em}{.05em}{.05em}
		& \multicolumn{6}{c}{\bfseries mIoU (\%)}   \\
		\cline{2-7}
		\bfseries Method & \bfseries Dust & \bfseries Night & \bfseries N.Dust & \bfseries Rain & \bfseries Fog & \bfseries Noise \\
		\specialrule{.1em}{.05em}{.05em}
		CM0	& 1.67 & 5.14 & 16.07 &  \bfseries 8.32 & 28.73 & 66.49 \\
		CM1	& 1.35 & 5.73 & \bfseries 11.08 & \bfseries 11.46 & 24.79 & 51.31 \\
		CM2	& 1.98 & \bfseries 1.37 & 21.14 &  15.19 & \bfseries 20.39 & 55.91 \\
		CM3	& \bfseries 0.83 & \bfseries 0.62 & \bfseries 13.25 &  17.58 & 23.91 & 47.93 \\
		CM4	& \bfseries 0.00 & \bfseries 1.26 & \bfseries 9.32 &  16.71 & \bfseries 20.38 & 43.52 \\
		CM5	& \bfseries 0.70 & 2.57 & 19.20 &  15.61 & 24.39 & 51.57 \\ 
		CM6	& \bfseries 0.28 & 3.57 & 18.63 & 18.05 & \bfseries 17.70 & \bfseries 25.14 \\ 
		CM7	& 2.11 & 1.67 & \bfseries 12.86 & 22.93 & \bfseries 17.30 & 43.95 \\
		CM8	& 2.71 & 1.49 & \bfseries 10.69 & 19.17 & 23.60 & \bfseries 32.72 \\
		\specialrule{.1em}{.05em}{.05em}
	\end{tabular}
\end{table}

The configurations CM3 and CM4 have had less mIoU degradation on the off-road tests, the architectures CM0 and CM1 have had the best results on raining testes, and CM6 and CM7 have performed better on synthetic impairments. However, CM3 and CM4 use output stride 16 and demand less parameter and MAC operations to carry out inference. 

\subsection{Field Experiments and real-time embedded inference}

Although there has been a growth in the CNN application for vision algorithms, enabling increasingly accurate semantic segmentation, there is still a challenge of equalizing the demand for computational power since visual perception for autonomous vehicles needs to run in real-time. To perform the field tests, we have ported two configurations (CM0-TRT and CM3-TRT) to achieve real-time inference and to embed them in a car. We have used the Drive PX2 hardware composed of ARM64 CPUs and CUDA cores. To carry out the reimplementation of our network, we have used the framework TensorRT and C++/CUDA to remove, fuze, and customize some layers.  

\autoref{tab:inference_comparison} shows the results achieved in the embedded hardware Drive PX2. We also have tested the optimized architectures in a GPU 1080 TI and have achieved a significant increase in FPS compared with our simulation using Tensorflow. With the CM0-TRT that demands more MAC operations, we have reached about 8 FPS in Drive PX2 and 40 FPS in the GTX 1080TI. On the other hand, with the CM3-TRT, we have achieved 21 FPS in Drive PX2 and almost 100 FPS for GTX 1080TI.

\begin{table}[!t]
	\centering
	\scriptsize
	\addtolength{\leftskip} {-2cm}
	\addtolength{\rightskip}{-2cm}
	\caption{Inference time for optimized networks.}
	\label{tab:inference_comparison}
	\begin{tabular}{+c+l| +c +c}
		\specialrule{.1em}{.05em}{.05em}
		\bfseries Method & \bfseries Arquitecture & \bfseries FPS & \bfseries Std. \\
		\specialrule{.1em}{.05em}{.05em}
		\multirow{2}{*}{CM0-TRT}	& Drive PX2  &  7.92 & 0.06\% \\
		& GTX 1080TI & 40.47 & 1.42\% \\
		\hline
		\multirow{2}{*}{CM3-TRT}	& Drive PX2  & 21.19 & 0.17\%\\
		& GTX 1080TI & 99.09 & 5.74\% \\
		\hline
		CM0	& \multirow{2}{*}{GTX 1080TI}  & 24,37 & 2.87\%\\
		CM3	& 							  & 35,42 & 8.99\% \\
		\specialrule{.1em}{.05em}{.05em}
	\end{tabular}

\end{table}

The optimized networks (CM0-TRT and CM3-TRT) have been capable of delivering better performance than their standard implementation and simulation on Tensorflow (CM0 and CM3). As can be seen in \autoref{tab:inference_comparison} about the comparison for GTX 1080TI, the optimized version of CM0 almost has doubled the FPS and decreased the standard deviation (Std.) by the heaf. For the CM3, the inference speed has been more than double.

We have noted that the standard deviation for the embedded ARM64 platform has been much less than for the x86\_64 hardware (GTX 1080TI). This indicates that, despite not having an FPS as high as the x86, the ARM platform delivers better predictability and stability for the system.

\begin{figure}[!t]
	\captionsetup[subfigure]{font=scriptsize,labelformat=empty}
	\begin{subfigure}[b]{0.325\linewidth}
		\includegraphics[width=\linewidth]{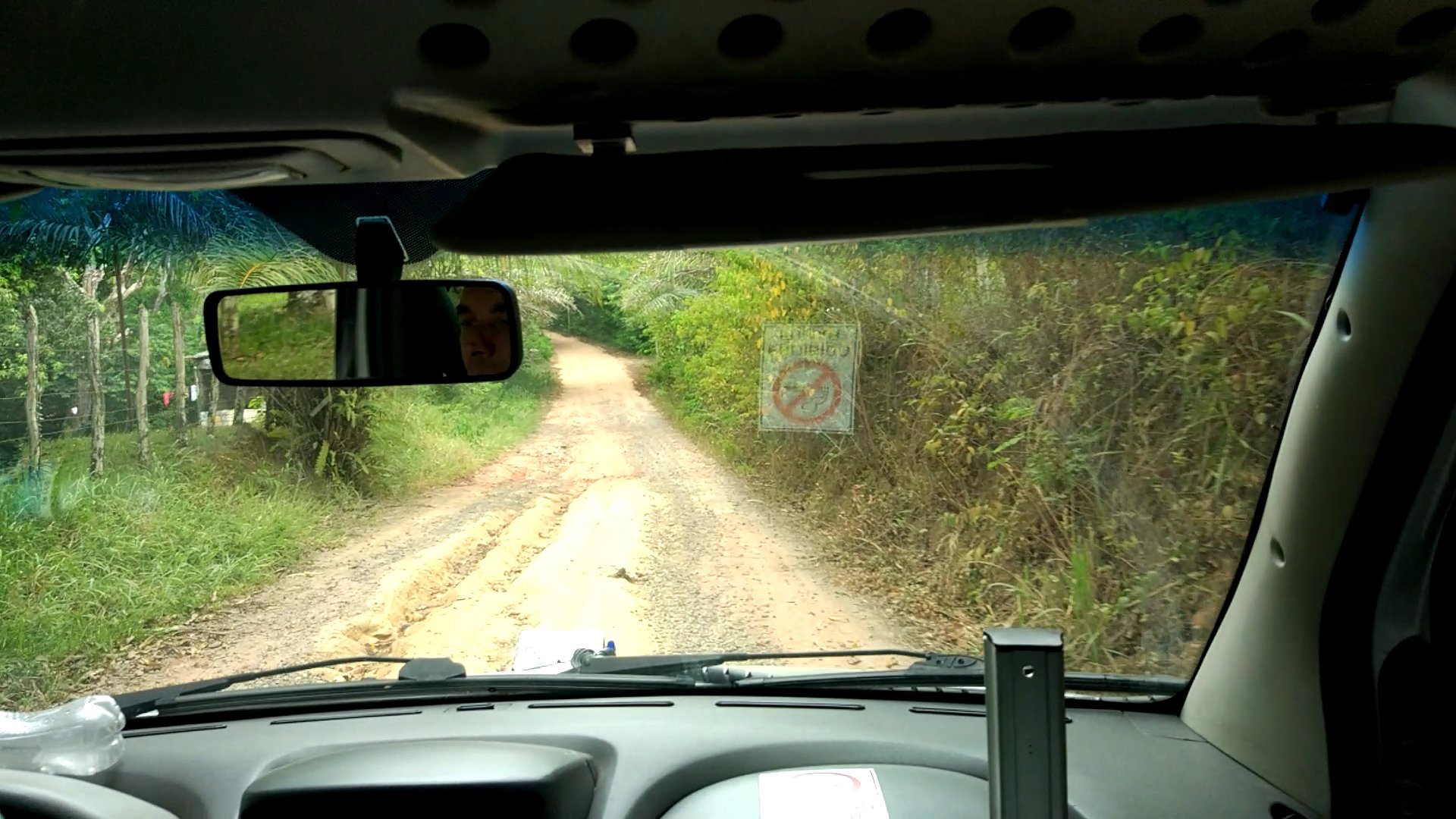}
	\end{subfigure}%
	\hfill
	\begin{subfigure}[b]{0.325\linewidth}
		\includegraphics[width=\linewidth]{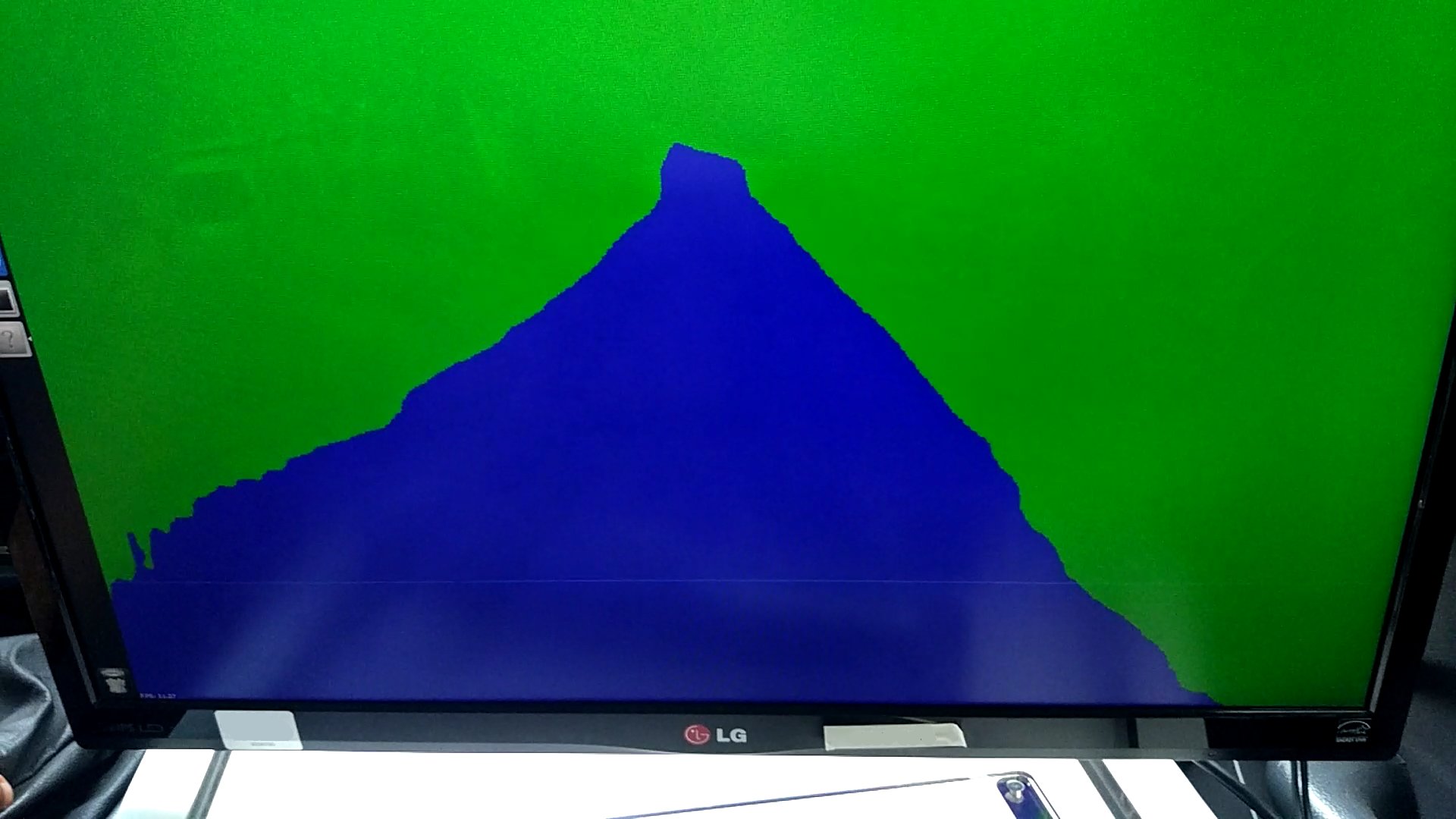}
	\end{subfigure}%
	\hfill
	\begin{subfigure}[b]{0.325\linewidth}
		\includegraphics[width=\linewidth]{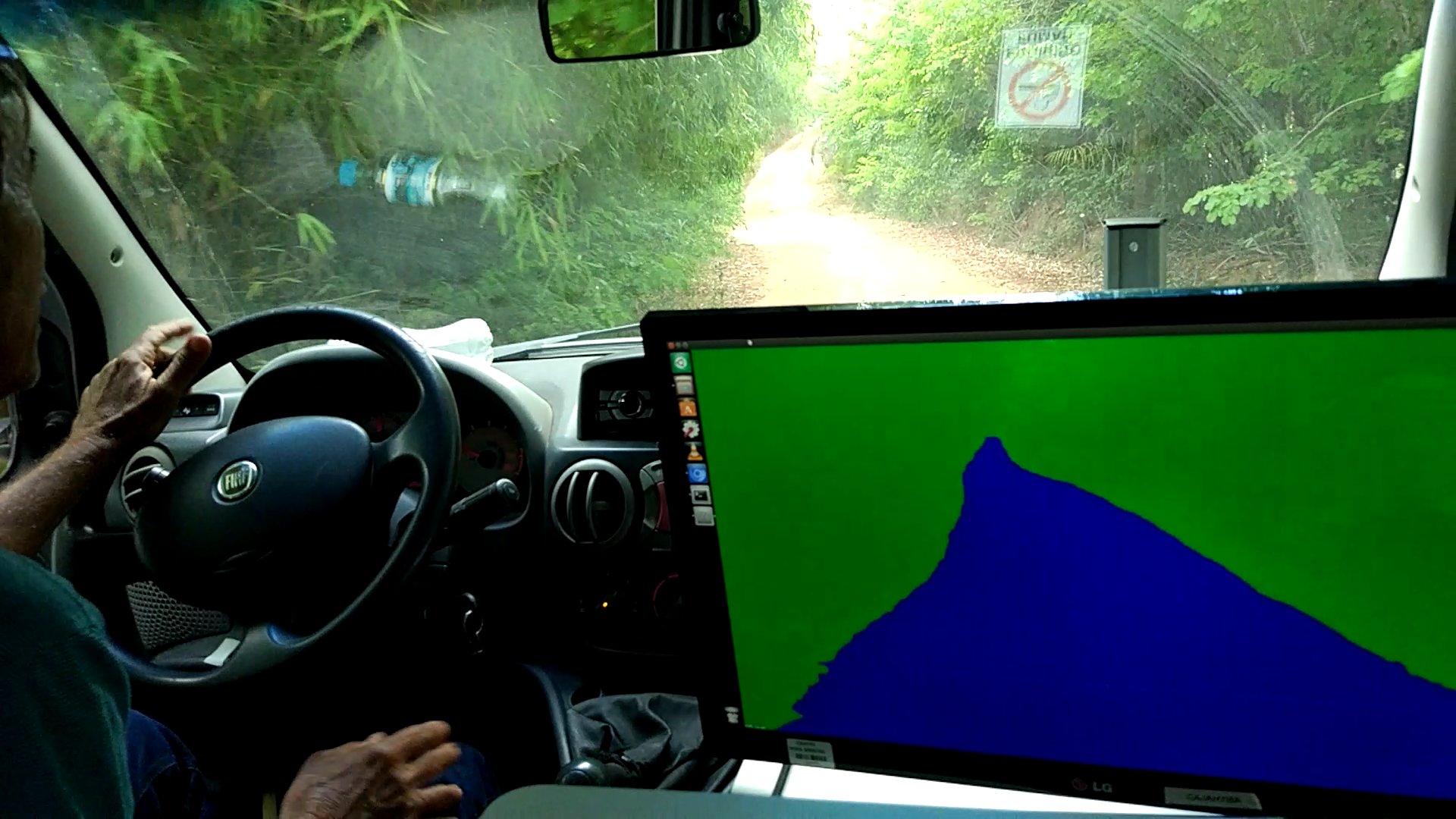}
	\end{subfigure}%
	\vspace{0.005\linewidth}
	
	\begin{subfigure}[b]{0.325\linewidth}
		\includegraphics[width=\linewidth]{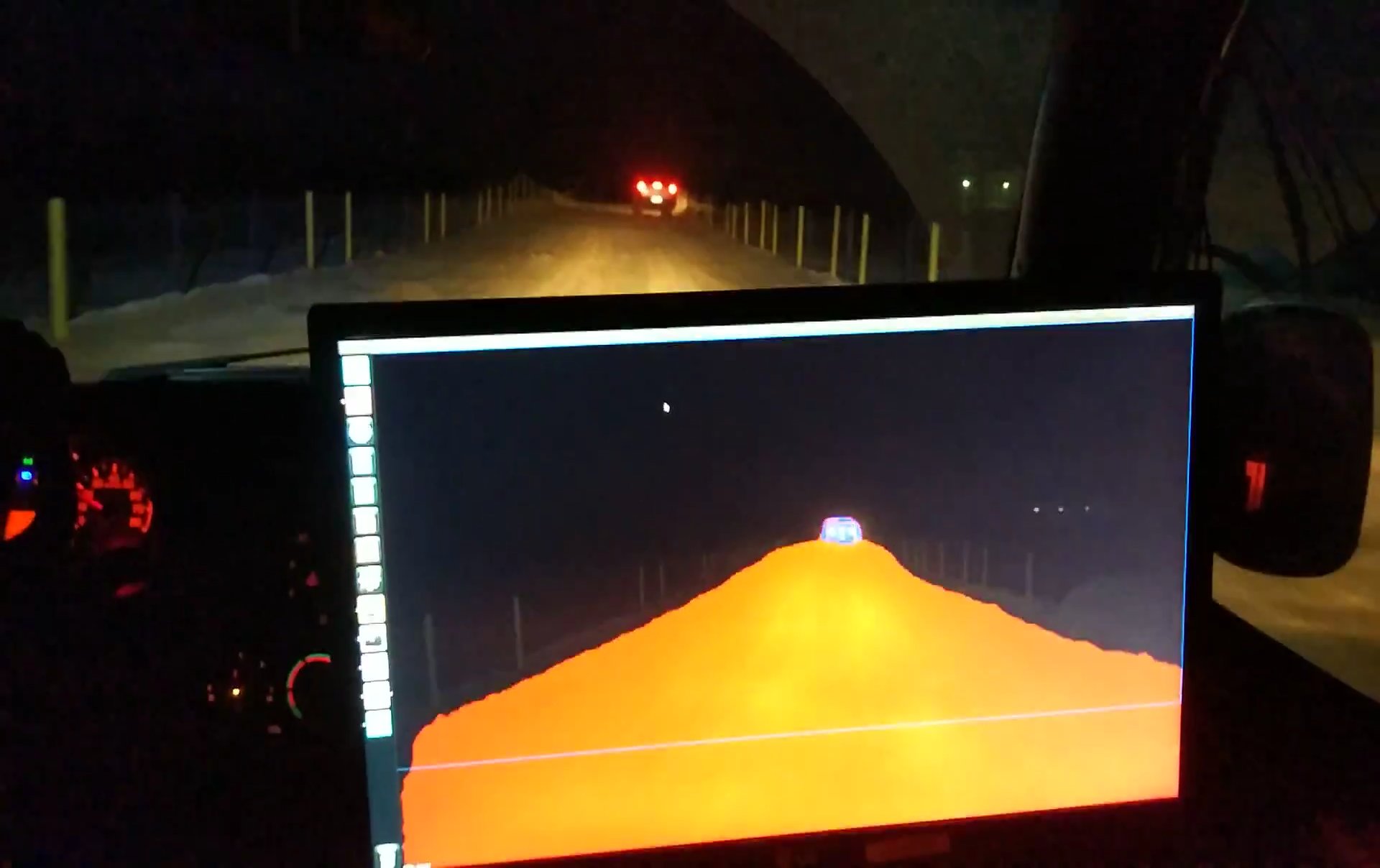}
	\end{subfigure}%
	\hfill
	\begin{subfigure}[b]{0.325\linewidth}
		\includegraphics[width=\linewidth]{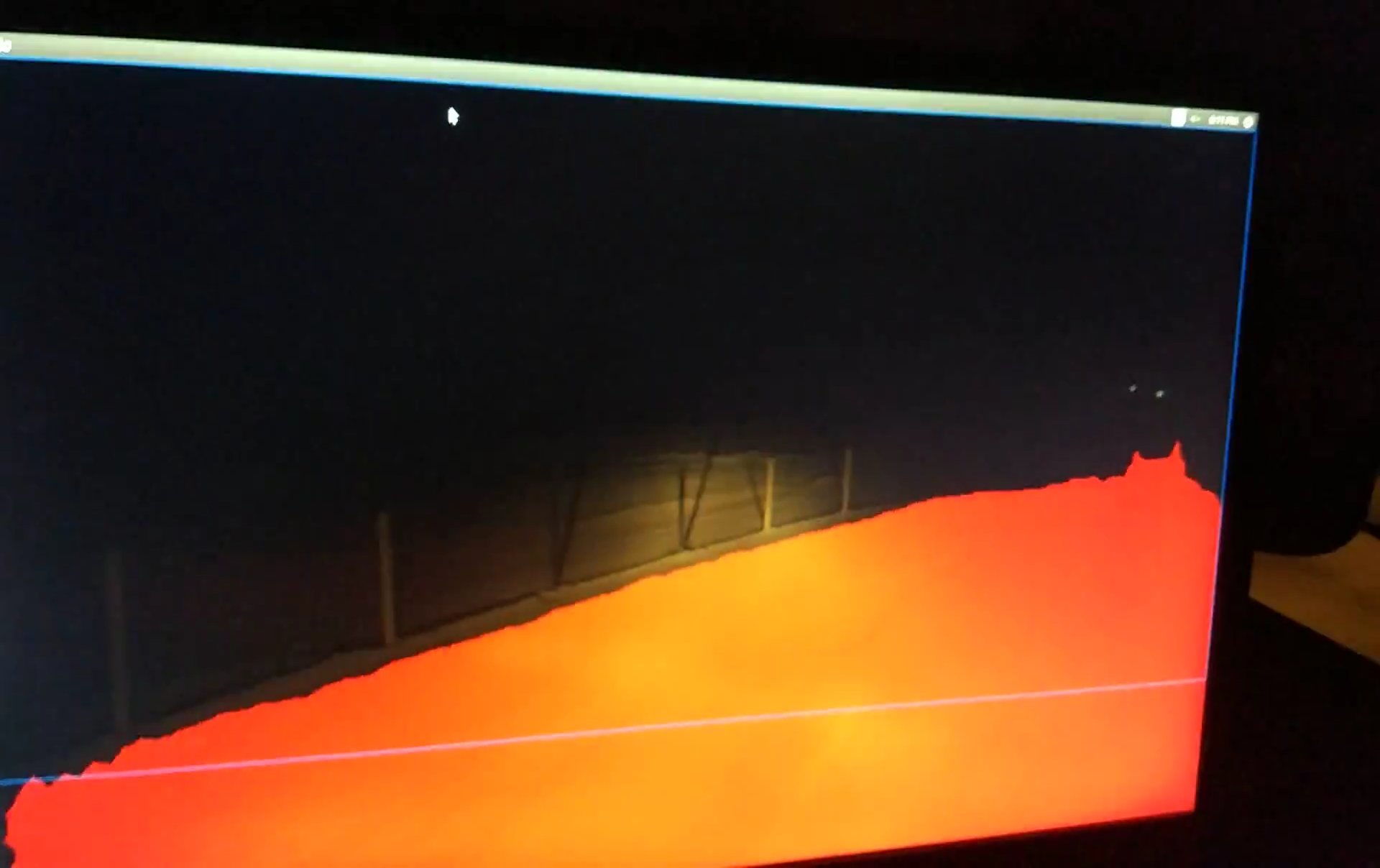}
	\end{subfigure}%
	\hfill
	\begin{subfigure}[b]{0.325\linewidth}
		\includegraphics[width=\linewidth]{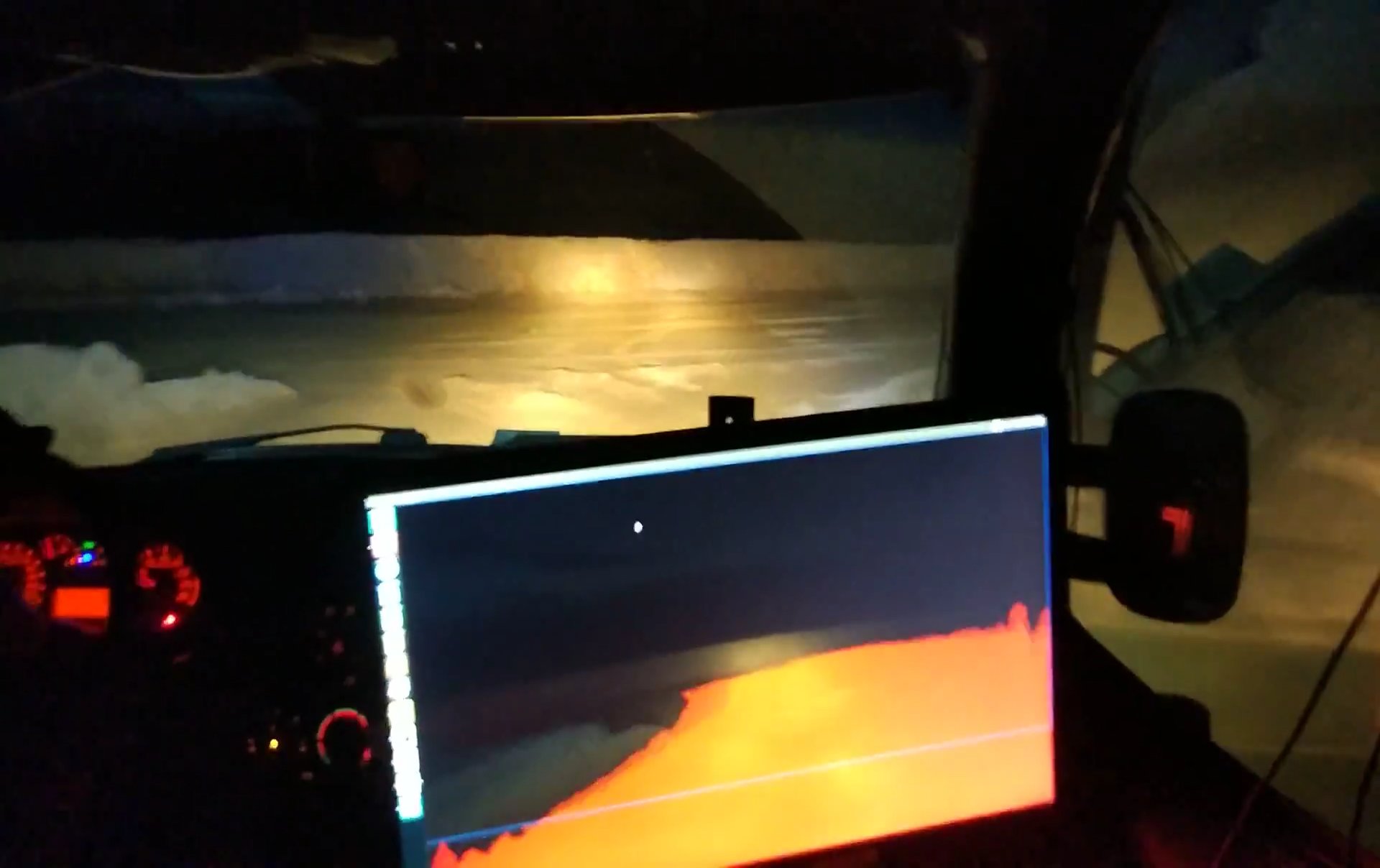}
	\end{subfigure}%

	\caption{Field tests carried out under different visibility conditions.}
	\label{fig:field_tests}
\end{figure}

\autoref{eq:distance-response-21-fps} shows a relationship between the velocity of the vehicle $V_{km/h}$ and distance $D_{m}$ traveled from the moment of the image capture and the processed information delivered. Considering the speed of 30 km/h with inference at 21 FPS on DRIVE PX 2, it is possible to have the information for decision making still 47 ms after the capture or only 39 cm from the event point. With this approach, we have obtained an acceptable response between what is perceived directly on the road and through the test monitor (Fig.	\ref{fig:field_tests}).


\begin{equation} \label{eq:distance-response-21-fps} 
D_{m}= \frac{V_{km/h}}{3.6*FPS}
\end{equation}

\section{Conclusion}
\label{sec:conclusion}
In this work, the researchers have proposed a perception system for autonomous vehicles and Advanced Driver Assistance Systems (ADAS) specialized in unpaved roads and off-road environments. The proposal focused on the use of deep learning with convolutional neural networks to perform the semantic segmentation of obstacles and areas of traffic on roads where there is no clear distinction between what is or not the track. Besides, the researchers have designed and built an off-road test track and assembled a hardware platform, including cameras and LiDARs, to enable the appropriate conditions for creating a dataset and conducting tests and validation of the proposed system. 

This work also evaluated the main components and tricks found in state-of-the-art architectures intended for semantic segmentation allowing researchers to propose a configurable modular segmentation network (CMSNet) framework. The CMSNet enabled testing of multiple configurations to find the most efficient arrangements to solve the segmentation problem for obstacles, unpaved roads, and off-road environments.

During the development of this research, it was built a dataset comprising almost 12,000 images, exploring various aspects of off-road environments and unpaved roads commonly found in developing countries. The proposed dataset was designed to include several conditions, such as rainy, night, and dusty, allowing the exploration of these aspects in the development of the research. Such a dataset was the most complete among those compared in this work.

In addition to the night, dust, and rain, the researchers also carry out tests with artificially generated impairments, such as fog and noise. Considering the tests performed, it was possible to conclude that the strategy of using convolutional networks and deep learning have been proved adequate. With the proposed CMSNet framework, it was possible to generate architectures capable of segment areas of traffic and obstacles with a high degree of accuracy.

Besides properly segmenting, the proposed networks needed to be embedded in a vehicle to carry out field tests. Such a goal was achieved through the appropriate selection of a backbone for features extraction focused on computational efficiency and by porting the trained network to hardware capable of guaranteeing the appropriated processing and stability, respecting the restrictions of the application.

With the embedded platform Drive PX2 was possible to deliver 21 FPS, and with a GTX 1080TI GPU was possible to achieve almost 100 FPS. Despite the lower computational power available on the embedded platform, the system's stability has proved to be satisfactory. The standard deviation over the average time to perform each inference cycle was only 0.16\%  Drive PX2, while, in the  GTX 1080TI, that value was about 5\%.

\FloatBarrier



\bibliography{references}

\end{document}